%% file: main.tex
\documentclass{article}
\pdftrailerid{}
\ifdefined\pdfsuppressptexinfo\pdfsuppressptexinfo=-1\fi  %
\usepackage{iclr2027_conference,times}
\usepackage{amsmath,amssymb,amsthm}
\usepackage{mathtools}
\usepackage{enumitem}
\usepackage[dvipsnames,table]{xcolor}
\usepackage{algorithm}
\usepackage{algorithmic}
\usepackage{booktabs}
\usepackage{multirow}
\usepackage{wrapfig}
\usepackage{adjustbox}
\usepackage{caption}
\usepackage{subcaption}
\usepackage{graphicx}
\usepackage{pifont}
\usepackage{float}
\usepackage{pgfplots}
\pgfplotsset{compat=1.17}
\usetikzlibrary{arrows.meta,calc,shapes.geometric,shapes.misc}
\usepackage{hyperref}
\hypersetup{colorlinks=true,linkcolor=MidnightBlue,citecolor=MidnightBlue,urlcolor=MidnightBlue}

\newcommand{\fmt}[1]{{\scriptsize(#1)}}

\AtBeginDocument{%
\setlength{\abovedisplayskip}{4pt plus 2pt minus 2pt}
\setlength{\belowdisplayskip}{4pt plus 2pt minus 2pt}
\setlength{\abovedisplayshortskip}{4pt plus 2pt minus 2pt}
\setlength{\belowdisplayshortskip}{4pt plus 2pt minus 2pt}
}
\renewcommand\arraystretch{1.4}
\newtheorem{theorem}{Theorem}
\newtheorem{proposition}[theorem]{Proposition}
\newtheorem{lemma}[theorem]{Lemma}
\newtheorem{corollary}[theorem]{Corollary}

\iclrfinalcopy
\makeatletter
\let\iclrlhead\lhead
\def\preprintmark#1{\iclrlhead{Preprint}}
\let\iclrmaketitle\maketitle
\def\maketitle{\let\lhead\preprintmark\iclrmaketitle\let\lhead\iclrlhead}
\makeatother

\title{Amortizing Exact Orthogonality:\\
       Drift-Triggered Restoration for Orthogonal Fine-Tuning}
\author{Lanxin Zhao$^{1}$ \quad Xiaoyu Li$^{2}$ \quad Dai Shi$^{3}$ \quad Junbin Gao$^{1}$ \quad Andi Han$^{1}$ \\
$^{1}$The University of Sydney \quad
$^{2}$University of New South Wales \quad
$^{3}$University of Cambridge \\
\texttt{lzha6608@uni.sydney.edu.au}, \texttt{andi.han@sydney.edu.au}}
\date{}

\begin{document}
\maketitle

\begin{abstract}
Orthogonal finetuning (OFT) adapts a pretrained weight by an orthogonal transform and preserves its singular values and row angles, but existing methods construct an exact or approximate orthogonal transform at every optimization step. We measure how far a transform $A$ is from orthogonal by its orthogonality defect $\|A^\top A-I\|_2$. For OFTv2, which truncates a Cayley--Neumann series, we show that the truncation order does not determine the defect. At its default order, block defects in checkpoints from the reference implementation span five orders of magnitude. We propose \emph{Restarted Tangent Orthogonal Finetuning} (RTOFT), which uses the first-order update $W(I+S)$ of a weight $W$ with a skew-symmetric $S$ and restores exact orthogonality by a polar decomposition only when the drift $\|S\|_2$ reaches a tolerance $\tau$. The defect of $I+S$ is exactly $\|S\|_2^2$. Therefore, reading the drift after every update and restoring before the next forward pass guarantees a defect of at most $\tau^2$ in every forward pass, independently of the loss, data, model, and optimizer. A final restoration makes the returned transform exactly orthogonal. On GLUE, RTOFT reduces the training time of block-diagonal, subspace, and butterfly parameterizations from $1.10$--$1.58\times$ to $0.97$--$1.13\times$ that of LoRA, and its block-diagonal version exceeds LoRA by $1.98\pm0.39$ points on the six-task average over five seeds. At $\tau=0.10$, the largest drift read during training stays between $0.10$ and $0.12$ across DeBERTaV3, Qwen, and diffusion models, whereas it ranges from $0.17$ to $1.65$ when restoring once per epoch. In all seven settings of our ablation, the tangent update is worse without restoration than with it, and restoring after every step yields no measurable accuracy gain over restoring at the tolerance and costs $4.5$--$27.5\%$ more per step.
\end{abstract}

\begin{figure}[H]
\centering
\vspace{-5pt}
\includegraphics[width=\linewidth]{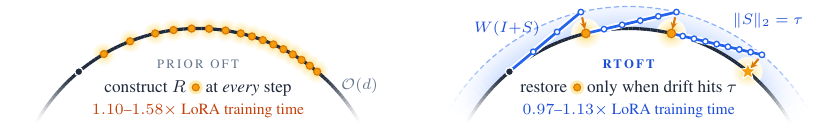}
\vspace{-16pt}
\caption{\textbf{Training with prior orthogonal finetuning (left) and with RTOFT (right).} Prior methods construct an exact or approximate orthogonal matrix $R$ at every step. RTOFT uses the tangent update $W(I{+}S)$ with a skew-symmetric matrix $S$, for which the orthogonality defect of $I{+}S$ is exactly $\|S\|_2^2$, and restores exact orthogonality when the drift $\|S\|_2$ reaches the tolerance $\tau$. The arc is the set $\mathcal{O}(d)$ of orthogonal matrices, shown for a $2{\times}2$ rotation with $\tau$ enlarged for visibility, and the training times are those on GLUE relative to LoRA.}
\label{fig:teaser}
\vspace{-9pt}
\end{figure}

\section{Introduction}
\vspace{-5pt}

Parameter-efficient finetuning (PEFT) adapts pretrained models with a small number of trainable parameters. Low-rank methods such as LoRA \citep{hu2022lora} use additive updates, whereas orthogonal finetuning (OFT) adapts a pretrained weight matrix $W$ to $W^{+}=WR$ with an orthogonal matrix $R$, which preserves the singular values of the weight and the angles between its rows \citep{qiu2023oft}. OFT can retain more of the general capability of the pretrained model while adapting it to the target task \citep{huang2026peftarena}, and recent variants that use block-diagonal, butterfly, or low-dimensional rotations have largely closed the gap in trainable parameters to low-rank methods \citep{qiu2025orthogonal,liu2024parameter,wu2025efficient}. Despite these benefits, OFT generally takes longer to train than low-rank methods due to the orthogonal transform, exact or approximate, that current methods construct and differentiate through at every optimization step.

OFTv2 replaces the exact Cayley transform by a truncated Cayley--Neumann series and keeps competitive downstream performance \citep{qiu2025orthogonal}, so the transform need not be exactly orthogonal at every step. This raises the question of \emph{whether an orthogonal transform, exact or approximate, needs to be constructed at every step at all}.
Moreover, the truncation order of OFTv2 does not determine how far its transform is from orthogonal. We measure this by the \emph{orthogonality defect} $\delta(A):=\|A^\top A-I\|_2$ of a transform $A$. At a fixed truncation order, the defect of the blocks of a block-diagonal transform varies by orders of magnitude across blocks and tasks (Section~\ref{sec:audit}).

\begin{figure}[t]
    \centering
    \input{figures/fig2_evidence}
    \vspace{-5pt}
    \caption{
    \textbf{Accuracy, training time, general capability, and numbers of restorations of RTOFT.}
    \textbf{(a)} Average score on six GLUE tasks with DeBERTaV3-base against training time relative to LoRA. At the same number of trainable parameters, RTOFT takes $12\%$ less time than OFTv2 and scores $1.09$ points higher.
    \textbf{(b)} One step of RTOFT and one restoration, where $W_n$ is the weight after the $n$-th restoration. Between two restorations, a forward pass needs one multiplication by the structured matrix $S$, and the drift $\|S\|_2$ is read after each update without gradient. Prior methods construct a transform at every step and differentiate through it. The transform of OFTv2, denoted by $\widetilde R_k(Q)$, is the order-$k$ truncation of the Cayley transform of a skew-symmetric matrix $Q$.
    \textbf{(c)} Change in target performance and in general capability on the PEFT-Arena benchmark relative to the pretrained model. RTOFT with the drift trigger ($\tau$) improves on OFTv2 at the same block size on both. Without restoration, RTOFT loses $10.82$ points more general capability than with restoration once per epoch (period), and a projection onto the orthogonal group after training recovers only $0.24$ points.
    \textbf{(d)} With $\tau=0.10$ in three settings, where S2I denotes segmentation-to-image, the numbers of restorations differ by up to $187\times$, and the largest drift, read after an update and before the restoration that it triggers, is $0.10$--$0.12$.
    }
    \label{fig:hero}
    \vspace{-10pt}
\end{figure}
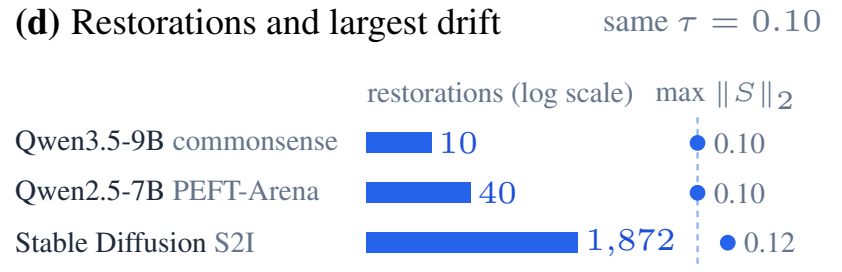

We propose \emph{Restarted Tangent Orthogonal Finetuning} (RTOFT; Figure~\ref{fig:teaser}), which does not construct an orthogonal transform at every step. It instead uses the tangent update
\[
    W^{+}=W(I+S),\qquad S^\top=-S,
\]
which needs only one multiplication by the structured matrix $S$. The orthogonality defect of $I+S$ is known exactly and is given by
\[
    \delta(I+S)
    =\|(I+S)^\top(I+S)-I\|_2
    =\|S\|_2^2.
\]
We call $\|S\|_2$ the \emph{drift}. When the drift reaches a tolerance $\tau$, RTOFT restores exact orthogonality, and we refer to this rule as the drift trigger. A restoration multiplies $W$ by the orthogonal polar factor of $I+S$ and resets $S$ to zero, and $W$ stays fixed until the next restoration. Reading the drift after each update and restoring before the next forward pass guarantees a defect of at most $\tau^2$ in every forward pass, with no assumption on the task or the optimizer, and a final restoration makes the returned transform exactly orthogonal.

The contributions of this work can be summarized as follows.
\begin{itemize}
    \item \textbf{Orthogonality of truncated maps.}
    We show that the truncation order $k$ of OFTv2 does not determine its orthogonality defect. At the default $k=5$, block defects in checkpoints from the reference implementation span five orders of magnitude (Table~\ref{tab:audit-full}). Compared with the default, $k=3$ needs one block multiplication fewer, has a smaller defect, and reaches the same downstream accuracy within $0.03\pm0.05$ points (Section~\ref{sec:audit}).

    \item \textbf{Tangent updates with a bounded defect.}
    We propose RTOFT and show that its defect is exactly $\|S\|_2^2$ between restorations and at most $\tau^2$ in every forward pass under the drift trigger (Section~\ref{sec:analysis}). At $\tau=0.10$, the largest drift read during training stays between $0.10$ and $0.12$ across DeBERTaV3, Qwen, and diffusion models, although the numbers of restorations differ by orders of magnitude.

    \item \textbf{Training time and accuracy.}
    On GLUE, RTOFT reduces the training time from $1.10$--$1.58\times$ to $0.97$--$1.13\times$ that of LoRA for three orthogonal parameterizations, namely the truncated block-diagonal one of OFTv2 and the exact subspace and butterfly ones of PSOFT and BOFT. Its block-diagonal version exceeds OFTv2 by $1.09\pm0.25$ and LoRA by $1.98\pm0.39$ points on the six-task average (Section~\ref{sec:glue}).

    \item \textbf{When to restore.}
    We show that the tangent update is worse without restoration than with it in all seven settings of our ablation study. Compared with the drift trigger, restoring after every step yields no measurable accuracy gain and increases the time of a step by $4.5$--$27.5\%$ on three GLUE tasks (Section~\ref{sec:ablation} and Table~\ref{tab:step-time}).
\end{itemize}

\vspace{-5pt}
\section{Preliminaries}
\label{sec:related}
\vspace{-5pt}

Let $W\in\mathbb{R}^{d_{\rm out}\times d}$ denote a pretrained weight matrix.
Low-rank PEFT adapts $W$ additively, and LoRA \citep{hu2022lora} uses
$W^{+}=W+\Delta W$, where $\Delta W$ has rank at most
$r\ll\min(d_{\rm out},d)$.

Orthogonal finetuning (OFT) instead uses
$W^{+}=WR$ with
$R\in\mathcal{O}(d):=\{R\in\mathbb{R}^{d\times d}:R^\top R=I\}$,
constraining the adapted transform to the orthogonal group
\citep{qiu2023oft}.
Except for HRA \citep{yuan2024hra}, which uses Householder reflections, the orthogonal PEFT methods considered here construct $R$ from
structured skew-symmetric matrices. We write
${\rm Skew}(d):=\{S\in\mathbb{R}^{d\times d}:S^\top=-S\}$,
${\rm skew}(A):=(A-A^\top)/2$, and denote by
${\rm polar}(A)$ the orthogonal polar factor of $A$.
Throughout, $\|\cdot\|_2$ denotes the spectral norm. The orthogonality defect
$\delta(A)=\|A^\top A-I\|_2$ of $A\in\mathbb{R}^{d\times d}$ is zero if and only if
$A\in\mathcal{O}(d)$.

We consider three structured parameterizations of a skew-symmetric matrix $S$. The block-diagonal
parameterization of OFT and OFTv2 uses
$S={\rm blockdiag}(S_1,\dots,S_{d/b})$ with $S_i\in{\rm Skew}(b)$ for a block size $b$
dividing $d$. The subspace parameterization of PSOFT uses $S=P_r^\top S_rP_r$ with $S_r\in{\rm Skew}(r)$ and a fixed matrix $P_r\in\mathbb{R}^{r\times d}$ with orthonormal rows. The butterfly parameterization of BOFT is a product of sparse factors, which reduces to one skew-symmetric matrix to first order (Appendix~\ref{app:generality})
\citep{liu2024parameter,qiu2025orthogonal,wu2025efficient}. Appendix~\ref{app:variants} also considers a low-rank parameterization, and Appendix~\ref{app:related} discusses further related work.

\vspace{-5pt}
\section{Orthogonality Defect of Truncated Cayley--Neumann Maps}
\label{sec:audit}
\vspace{-5pt}

At every training step, OFTv2 \citep{qiu2025orthogonal} evaluates an
order-$k$ truncation $\widetilde R_k$ of the Cayley transform
$R=(I+Q)(I-Q)^{-1}$ with
$Q\in{\rm Skew}(d)$, which equals the Cayley--Neumann series $I+2\sum_{i\geq1}Q^i$ when $\|Q\|_2<1$. Here, $k$ is the truncation order set in the reference implementation, and Lemma~\ref{lem:cnp} gives the matrix that the implementation returns for each $k$. For $Q\neq0$, $\widetilde R_k$ is generally not
orthogonal, and its defect depends on both $k$ and
$\|Q\|_2$. Since $\|Q\|_2$ changes during training, the defect also changes at a fixed $k$.

\begin{table}[t]
\centering
\small
\renewcommand{\arraystretch}{1.05}
\begin{tabular}{lcccc}
\toprule
$k$ & CoLA & RTE & STS-B & spread \\
\midrule
$2$ & $5.4\times10^{-3}$ & $2.7\times10^{-2}$ & $6.4\times10^{-2}$ & $12$ \\
$3$ & $7.9\times10^{-9}$ & $9.3\times10^{-7}$ & $1.2\times10^{-5}$ & $1460$ \\
$4$ & $7.9\times10^{-9}$ & $9.3\times10^{-7}$ & $1.2\times10^{-5}$ & $1460$ \\
$\mathbf{5}$ & $\mathbf{6.2\times10^{-6}}$ &
$\mathbf{1.5\times10^{-4}}$ &
$\mathbf{8.1\times10^{-4}}$ & $\mathbf{132}$ \\
$6$ & $5.6\times10^{-14}$ & $1.5\times10^{-10}$ &
$9.8\times10^{-9}$ & $1.7\times10^{5}$ \\
\bottomrule
\end{tabular}
\caption{Median orthogonality defect over the adapted blocks of checkpoints from the
reference OFTv2 implementation on three GLUE tasks, where the same checkpoints are evaluated at each truncation order $k$. Spread is the ratio of the largest to the smallest median across
the tasks, and $k=5$ (bold) is the default. The smallest and largest block defects are given in
Table~\ref{tab:audit-full}.}
\label{tab:audit}
\end{table}
Lemma~\ref{lem:cnp} shows that the leading term of $\delta(\widetilde R_k)$ is proportional to a power of $\|Q\|_2$ that depends on $k$.
A transform with defect $\delta$ lengthens any input direction by a factor of at most
$\sqrt{1+\delta}$ (Corollary~\ref{cor:stretch}), so $\delta=10^{-2}$ corresponds to a stretch of about half a percent.
Table~\ref{tab:audit} shows that the defects differ greatly across tasks at the same order.
The defect at a higher order on one task can even exceed that at a lower order on another. For instance, the median defect at $k=6$ on STS-B is larger than that at $k=3$ on CoLA, although the latter
needs two block multiplications fewer. Over the $5{,}184$ adapted blocks that we measure, the defects at the default $k=5$ range from $1.3\times10^{-7}$ to $1.4\times10^{-2}$ (Table~\ref{tab:audit-full}).

If the norm $\|Q\|_2$ is $\rho$ times as large on one task as on another, a defect with leading term proportional to $\|Q\|_2^p$ is about $\rho^p$ times as large. With the exponents $p=2$, $6$, $6$, $4$, and $10$ for $k=2,\ldots,6$ (Lemma~\ref{lem:cnp}), all five rows of
Table~\ref{tab:audit} are consistent with one ratio $\rho\approx3.4$ across the tasks.

Moreover, the defect is not monotone in the order. By
Lemma~\ref{lem:cnp}, the matrix that the reference implementation returns at order $k$ is the exact Cayley
transform multiplied by $I-Q^\mu$, where $\mu=3$ for $k\in\{3,4\}$ and
$\mu=k-1$ for $k\geq5$, and $k=2$ is a separate case. The defect is of order $\|Q\|_2^{2\mu}$ when $\mu$ is odd, because the leading terms
cancel (Corollary~\ref{cor:parity}), and of order $\|Q\|_2^\mu$ when $\mu$ is even. The exponents measured on the implementation agree with Lemma~\ref{lem:cnp} to two decimals (Appendix~\ref{app:ablation-detail}). In particular, $k=4$
returns the same matrix as $k=3$ at the same cost.

The default $k=5$ does not have the smallest defect on any of the three tasks. At $k=3$, the defect is smaller on all of them and one block multiplication fewer is needed. When used for finetuning, $k=3$ matches the default on the six-task GLUE average, with a paired difference of $+0.03\pm0.05$ points
(mean and standard error over five seeds), and trains $3.6\%$ faster
(Appendix~\ref{app:ablation-detail}).

\vspace{-5pt}
\section{Tangent Updates with Drift-Triggered Restoration}
\label{sec:method}

\vspace{-5pt}

\vspace{-3pt}
\subsection{Tangent Update}
\label{sec:tangent}

\vspace{-3pt}
When parameterized by the same tangent vector, the exponential, Cayley, and
polar retractions have the same first-order expansion around the identity
\citep{absil2008optimization,boumal2023intromanifolds}.
Between two restorations, RTOFT keeps only this common first-order term and drops the higher-order terms:
\begin{equation}
    W^{+}=W(I+S),
    \qquad S\in{\rm Skew}(d).
    \label{eq:tangent-update}
\end{equation}
Here, $W$ is the pretrained weight before the first restoration and the restored weight afterward. Since $W(I+S)x=Wx+W(Sx)$, the forward pass needs one multiplication by the structured matrix $S$ in addition to the pretrained layer, and no orthogonal matrix, matrix inverse, or
iterative orthogonalization is computed (Figure~\ref{fig:hero}b).

\vspace{-3pt}
\subsection{Drift-Triggered Restoration}
\label{sec:restart}

\vspace{-3pt}
For a skew-symmetric $S$, $(I+S)^\top(I+S)=I-S^2$ and hence $\delta(I+S)=\|S\|_2^2$.
At a \emph{restoration}, RTOFT replaces $I+S$ by the
nearest orthogonal matrix in the Frobenius norm, which is the polar factor
\citep{fan1955metric,higham1986polar} ${\rm polar}(I+S)=(I+S)(I-S^2)^{-1/2}$. It multiplies the weight by this factor and resets $S$ to zero. Because the tangent update then starts again from an orthogonal transform, we refer to the method as restarted. We denote by $R_n$ the product of the polar factors of the first $n$ restorations. Equivalently, $R_n$ may be stored in an
adapter while the pretrained weight remains frozen
(Appendix~\ref{app:variants}).

Under the \emph{drift trigger}, RTOFT restores when the drift $\|S\|_2$, the square root of the defect, reaches a tolerance $\tau$. The number of steps between two restorations is then determined by how fast the drift grows. For comparison, we consider a \emph{fixed period}, under which exact orthogonality is restored after a fixed number of steps. We refer to a period of one epoch as the \emph{one-epoch period}.
Under the block parameterization,
$\|S\|_2=\max_i\|S_i\|_2$, so reading the drift needs one spectral norm of a $b\times b$ matrix
per block. This costs at most $O(db^2)$ per optimizer step and is computed without gradient outside the forward pass.
The training procedure is summarized in Algorithm~\ref{alg:rtoft}
(Appendix~\ref{app:proof}).

\vspace{-3pt}
\subsection{Relation to the Cayley--Neumann Family}
\label{sec:equivalence}

\vspace{-3pt}
The tangent map $I+S$ appears in prior work. The truncation of
OFTv2 \citep{qiu2025orthogonal} at $k=2$ is
$I+2Q$, which is the tangent map in Equation~\ref{eq:tangent-update} with
$S=2Q$, and Lemma~\ref{lem:cnp} gives it the same defect
$4\|Q\|_2^2=\|S\|_2^2$. RTOFT differs from OFTv2 in
restoration. OFTv2 does not restore orthogonality during training, whereas RTOFT keeps the inexpensive
tangent map and restores exact orthogonality whenever the drift reaches the tolerance.

Restoration can also be combined with a truncation of higher order. When both maps are restored once per epoch, the $k=2$ map scores $0.86$ points above the default $k=5$ on the six-task GLUE average (Table~\ref{tab:order-restart}).

\vspace{-5pt}
\section{Bounds on the Orthogonality Defect}
\label{sec:analysis}

\vspace{-5pt}

Let $W^{\rm pre}\in\mathbb{R}^{d_{\rm out}\times d}$ be the pretrained
weight and $W_n=W^{\rm pre}R_n$ the weight after the $n$-th restoration. The start of training is counted as $n=0$, with $R_0=I$, and a final restoration at the end of training gives the returned model. We call the steps between two consecutive restorations a \emph{cycle}.
Within a cycle, let $S\in{\rm Skew}(d)$ denote the current tangent parameter, which is zero at the start of the cycle. All proofs are provided in Appendix~\ref{app:proof}. Let $\mathcal{SO}(d)$ be the group of rotations, that is, the orthogonal $d\times d$ matrices with determinant $+1$.

\begin{theorem}[Orthogonality at and between restorations]
\label{thm:main}
In exact arithmetic, at every restoration, including the final one,
\[
    W_n=W^{\rm pre}R_n,\quad R_n\in\mathcal{SO}(d),
\]
and within each cycle, the transform $A=R_n(I+S)$ satisfies
\[
    \|A^\top A-I\|_2=\|S\|_2^2.
\]
Hence restorations exactly preserve the row Gram
matrix and the singular values of $W^{\rm pre}$.
\end{theorem}

Theorem~\ref{thm:main} indicates that the defect between two restorations can be computed exactly from the drift at every step.

\begin{corollary}[Defect bound under the drift trigger]
\label{cor:tau}
Suppose that, for a fixed $\tau>0$, $\|S\|_2$ is evaluated after every optimizer step and exact orthogonality is restored
before the next forward pass whenever $\|S\|_2\geq\tau$. Then, in exact arithmetic and independently
of the loss, data, model, or optimizer, every forward pass satisfies
\[
    \|A^\top A-I\|_2\leq\tau^2,
    \qquad
    \|(I+S)-{\rm polar}(I+S)\|_2
    =\sqrt{1+\|S\|_2^2}-1
    \leq\tfrac12\tau^2.
\]
\end{corollary}

Corollary~\ref{cor:tau} indicates that the tolerance, which is fixed before training, bounds the defect of every forward pass, as long as $\|S\|_2$ is read after every update and each triggered restoration happens before the next forward pass. When many matrices are adapted, the trigger reads the largest $\|S\|_2$ among them and restores all of them together. The number of restorations refers to the number of these events.

The tolerance can be chosen from a \emph{stretch budget} $\varepsilon$, the largest fraction by which $I+S$ is allowed to lengthen any direction. By Corollary~\ref{cor:budget}, the corresponding tolerance is
$\tau(\varepsilon)=\sqrt{(1+\varepsilon)^2-1}$, which equals $0.10012$ for
$\varepsilon=0.5\%$. We use $\tau=0.10$. It
guarantees that $I+S$ stretches any direction
by at most $0.499\%$ in every forward pass, or equivalently that $\delta\leq10^{-2}$.

We next compare the drift trigger with a fixed period. A period that is chosen before training to keep the drift below the tolerance has to be set for the fastest growth of the drift in a run, whereas the trigger restores only when the drift reaches the tolerance. To state this formally, we define the \emph{update length} of a set of steps as the sum of the spectral norms of the updates of $S$ in these steps. The update length since the last restoration is an upper bound on $\|S\|_2$. For a fixed sequence of updates, the trigger restores no more often than any schedule that keeps this update length below the tolerance, such as a period chosen from a bound on the step size (Proposition~\ref{prop:fewest}). Proposition~\ref{cor:howmany} further shows that, when every update of $S$ is small compared with the tolerance, such a period restores more often than the trigger, up to one restoration, by at least a factor close to the ratio of the largest mean update norm over one period to the mean over the run. This ratio is about two when the update norm follows a learning rate with warmup and linear decay and the period is short compared with the run. In our runs, a fixed period set to the shortest interval between two restorations of the trigger restores three to twelve times as often as the trigger (Appendix~\ref{app:pathlength}).

The bound $\delta\leq\tau^2$ requires the polar factor
and its product with the weight to be computed in single precision or better, even when the rest of the
model is trained in half precision. In bfloat16, the restorations of a schedule with $\tau=0.05$ leave a deviation from orthogonality of twenty times the bound $\tau^2=2.5\times10^{-3}$ (Appendix~\ref{app:precision}).

Finally, we compare RTOFT with restoring after every step. For gradient descent and under conditions on the gradient of the loss, the step size, the tolerance, and the cycle length, the transform at the end of each cycle is within $O(\tau^2)$ of that obtained from the same start by restoring exact orthogonality after every step (Lemma~\ref{lem:tracking}).

\vspace{-5pt}
\section{Experiments}
\label{sec:experiments}
\vspace{-5pt}

This section examines whether RTOFT keeps the accuracy of orthogonal
finetuning and the general capability that it retains, also referred to as retention, at a lower training cost. We consider
DeBERTaV3-base \citep{he2021debertav3} on GLUE, Qwen3.5-9B and Llama-3.1-8B on commonsense
reasoning, Stable Diffusion \citep{rombach2022ldm} on controllable generation,
and Qwen2.5-7B under PEFT-Arena \citep{huang2026peftarena} for retention.
Unless stated otherwise, RTOFT uses the block parameterization, denoted RTOFT-B, and the drift
trigger at $\tau=0.10$. We also report RTOFT with the one-epoch period, which differs only in the
restoration schedule. Within each benchmark, the methods that we run share one training
setup and learning-rate search, and RTOFT is compared with OFTv2 at the same block size
and number of trainable parameters. On GLUE, we use two protocols. Under the test-split protocol, the validation set of each task is split into two halves, one for selecting the checkpoint and one for the reported score, and we run every method in a common pipeline over five seeds (Table~\ref{tab:glue-test}). Under the validation-score protocol of prior work, the reported score is the best one on the validation set (Table~\ref{tab:glue-val}). The experiments consist of more than $880$ finetuning runs, and the runs in Table~\ref{tab:glue-runtime-full} alone take
$1{,}221$ V100-hours. Details on training,
evaluation, and timing are given in Appendix~\ref{app:impl}.

\vspace{-3pt}
\subsection{Accuracy and Efficiency on GLUE}
\label{sec:glue}

\vspace{-3pt}
\begin{table}[t]
\centering
\scriptsize
\renewcommand{\arraystretch}{1.06}
\setlength{\tabcolsep}{3.0pt}
\renewcommand{\arraystretch}{0.93}
\caption{Comparison of finetuning methods on GLUE with DeBERTaV3-base under the test-split protocol of \citet{wu2025efficient}. Every method is re-run in one pipeline, and RTOFT-B is the block variant of RTOFT. ``\#Param'' and ``Adapter'' represent the number of trainable parameters and the total stored adapter state, including frozen factors (Appendix~\ref{app:variants}), respectively. Cells are means over five seeds with standard deviations as subscripts. Time is the mean wall-clock time per task on one V100 in \emph{mm:ss}, and Rel.\ Time is normalized to LoRA (the protocol is given in Appendix~\ref{app:impl}). Bold and underlined values indicate the best and second-best results, respectively, and tied second-best results are unmarked.}
\label{tab:glue-test}
\begin{adjustbox}{max width=\linewidth}
\begin{tabular}{lcccc c cccccc c}
\toprule
& \multicolumn{5}{c}{\textbf{Efficiency}}
& \multicolumn{7}{c}{\textbf{GLUE performance}} \\
\cmidrule(lr){2-6}
\cmidrule(lr){7-13}
Method
& \#Param
& Adapter
& Mem.\,(GB)
& Time $\downarrow$
& Rel.\ Time $\downarrow$
& CoLA
& STS-B
& RTE
& MRPC
& SST-2
& QNLI
& Avg. \\
\midrule

Full finetuning
& 184M
& 184M
& 5.9
& 149:03
& 1.08$\times$
& 68.20$_{3.56}$
& 91.34$_{0.40}$
& 83.35$_{2.10}$
& 90.48$_{1.28}$
& 95.18$_{0.71}$
& 92.94$_{0.35}$
& 86.92 \\

\midrule
\multicolumn{13}{l}{\emph{Low-rank adaptation}} \\

LoRA$_{r=8}$
& 1.33M
& 1.33M
& 4.5
& 137:54
& 1.00$\times$
& 71.98$_{3.60}$
& 91.42$_{0.57}$
& 84.53$_{1.85}$
& 90.63$_{0.51}$
& 95.01$_{1.62}$
& 93.92$_{0.62}$
& 87.91 \\

PiSSA$_{r=8}$
& 1.33M
& 1.33M
& 4.5
& \textbf{127:10}
& \textbf{0.92$\times$}
& 67.52$_{2.48}$
& 91.43$_{1.16}$
& 86.17$_{1.45}$
& 90.25$_{2.42}$
& 94.69$_{1.47}$
& 92.90$_{0.86}$
& 87.16 \\

DoRA$_{r=8}$
& 1.41M
& 1.41M
& 5.8
& 138:28
& 1.00$\times$
& 67.99$_{1.15}$
& 91.92$_{0.94}$
& 87.68$_{3.26}$
& 89.24$_{1.75}$
& 95.25$_{2.31}$
& 93.99$_{0.37}$
& 87.68 \\

\midrule
\multicolumn{13}{l}{\emph{Orthogonal finetuning}} \\

PSOFT$_{r=46}$
& \textbf{0.08M}
& 7.71M
& \textbf{4.1}
& 169:52
& 1.23$\times$
& 71.60$_{1.98}$
& 91.42$_{0.39}$
& 85.75$_{3.98}$
& 90.53$_{1.40}$
& 95.18$_{0.70}$
& 93.60$_{1.24}$
& 88.01 \\

HRA$_{r=16}$
& 1.33M
& 1.33M
& 15.2
& 283:35
& 2.06$\times$
& 70.05$_{1.69}$
& 90.47$_{1.01}$
& 86.46$_{0.82}$
& \textbf{92.82}$_{1.32}$
& 94.71$_{0.81}$
& 93.15$_{0.75}$
& 87.94 \\

BOFT$_{b=8,m=2}$
& 1.41M
& 1.41M
& 6.3
& 218:01
& 1.58$\times$
& 68.85$_{2.09}$
& 91.85$_{0.72}$
& 86.78$_{2.82}$
& 89.91$_{1.17}$
& 95.33$_{1.56}$
& 93.85$_{0.41}$
& 87.76 \\

OFTv2$_{b=32}$ ($k{=}5$)
& 1.29M
& \textbf{1.29M}
& 4.6
& 151:03
& 1.10$\times$
& 71.64$_{1.53}$
& 91.60$_{0.46}$
& 88.48$_{2.01}$
& 92.15$_{3.90}$
& 94.90$_{0.75}$
& 94.06$_{1.00}$
& 88.81 \\

\midrule
\rowcolor{gray!10}
RTOFT-B$_{b=32}$ (one-epoch period)
& 1.29M
& \textbf{1.29M}
& 4.6
& 133:11
& 0.97$\times$
& \underline{72.52}$_{1.08}$
& \underline{92.30}$_{0.36}$
& \underline{88.55}$_{1.62}$
& \underline{92.78}$_{0.84}$
& \underline{95.88}$_{0.39}$
& \underline{94.65}$_{0.37}$
& \underline{89.45} \\
\rowcolor{gray!10}
\textbf{RTOFT-B$_{b=32}$ (ours)}
& 1.29M
& \textbf{1.29M}
& 4.6
& \underline{133:07}
& 0.97$\times$
& \textbf{73.55}$_{0.84}$
& \textbf{92.44}$_{0.14}$
& \textbf{89.20}$_{0.92}$
& 92.65$_{0.31}$
& \textbf{96.75}$_{0.16}$
& \textbf{94.80}$_{0.14}$
& \textbf{89.90} \\

\bottomrule
\end{tabular}
\end{adjustbox}
\vspace{-10pt}
\end{table}

\paragraph{Comparison with OFTv2 and LoRA.}
Table~\ref{tab:glue-test} shows that RTOFT-B improves on
OFTv2 on all six GLUE tasks at the same block size and number of trainable parameters, and that it reduces the mean training time from
$1.10\times$ to $0.97\times$ that of LoRA. The improvement on the
six-task average is $1.09\pm0.25$ points over the five shared seeds (mean and standard error of the paired difference), and it is positive on every seed. Subscripts in the tables are standard deviations over seeds.
OFTv2 is run at its default $k{=}5$, which $k{=}3$ matches on this average (Section~\ref{sec:audit}).
Compared with LoRA, RTOFT-B takes about the same training time
and improves the six-task average by $1.98\pm0.39$ points, again positive
on every seed.

The one-epoch period takes about the same training time as the drift trigger
($133{:}11$ versus $133{:}07$) and reaches $89.45$, compared with $89.90$ for the trigger.
The paired improvement of the trigger is $0.45\pm0.13$ points and is
positive on every seed. The two schedules share the learning rates, which are selected for the period (Table~\ref{tab:hp-glue-test}), and
differ only in when they restore.

\vspace{-6pt}
\paragraph{Validation-score protocol.}
The same ordering holds under the eight-task validation-score
protocol of prior PEFT comparisons (Table~\ref{tab:glue-val}). Over the same
five seeds, the drift trigger reaches an average of $90.48$, compared with
$89.95$ for the one-epoch period and $89.89$ for OFTv2. The paired improvement of the trigger over the period is
$0.53\pm0.02$ points and is positive on every seed and every task.

\begin{table}[t]
\centering
\scriptsize
\renewcommand{\arraystretch}{0.96}
\setlength{\tabcolsep}{3.5pt}
\caption{
Comparison of finetuning methods on GLUE with DeBERTaV3-base under the validation-score protocol.
The baselines are taken from \citet{liu2024parameter}, and OFTv2 and both
RTOFT schedules are re-run over the same five seeds. Subscripts denote
standard deviations over seeds, averaged over the eight tasks in the All column. Among the three re-run rows, the smallest
standard deviation for each task is boldfaced in the subscript.
Bold and underlined values indicate the best and second-best results, respectively.
}
\label{tab:glue-val}
\begin{adjustbox}{max width=\linewidth}
\begin{tabular}{lcc cccccccc c}
\toprule
Method & \#Param\,(M) & Mem.\,(GB)
       & MNLI & SST-2 & CoLA & QQP & QNLI & RTE & MRPC & STS-B & All \\
\midrule
Full finetuning & 184  & N/A & 89.90 & 95.63 & 69.19 & \textbf{92.40} & 94.03 & 83.75 & 89.46 & 91.60 & 88.25 \\
BitFit          & 0.10 & N/A & 89.37 & 94.84 & 66.96 & 88.41 & 92.24 & 78.70 & 87.75 & 91.35 & 86.20 \\
H-Adapter       & 1.22 & N/A & 90.13 & 95.53 & 68.64 & 91.91 & 94.11 & 84.48 & 89.95 & 91.48 & 88.28 \\
P-Adapter       & 1.18 & N/A & 90.33 & 95.61 & 68.77 & 92.04 & 94.29 & 85.20 & 89.46 & 91.54 & 88.41 \\
LoRA$_{r=8}$    & 1.33 & N/A & \underline{90.65} & 94.95 & 69.82 & 91.99 & 93.87 & 85.20 & 89.95 & 91.60 & 88.50 \\
AdaLoRA         & 1.27 & N/A & \textbf{90.76} & 96.10 & 71.45 & \underline{92.23} & \textbf{94.55} & 88.09 & 90.69 & 91.84 & 89.46 \\
BOFT$^{b=8}_{m=2}$ & 0.75 & N/A & 90.25 & \textbf{96.44} & 72.95 & 92.10 & 94.23 & 88.81 & 92.40 & \underline{91.92} & 89.89 \\
\midrule
OFTv2$_{b=16}$  & 0.62 & 3.28
 & 90.38$_{0.29}$ & 95.95$_{0.11}$ & 73.47$_{1.82}$ & 92.06$_{0.37}$
 & 94.11$_{0.68}$ & 88.70$_{2.01}$ & \underline{92.75}$_{1.21}$ & 91.69$_{0.28}$ & 89.89$_{0.85}$ \\
RTOFT$_{b=16}$ (one-epoch period) & 0.62 & 3.28
 & 90.40$_{0.15}$ & 96.10$_{\mathbf{0.07}}$ & \underline{73.55}$_{1.59}$ & 91.96$_{0.12}$
 & \underline{94.30}$_{0.52}$ & \underline{89.19}$_{1.36}$ & 92.21$_{0.71}$ & 91.89$_{\mathbf{0.13}}$ & \underline{89.95}$_{0.58}$ \\
\rowcolor{gray!10}
\textbf{RTOFT}$_{b=16}$ \textbf{(ours)} & 0.62 & 3.28
 & 90.54$_{\mathbf{0.13}}$ & \underline{96.30}$_{\mathbf{0.07}}$ & \textbf{74.65}$_{\mathbf{1.43}}$ & 92.05$_{\mathbf{0.11}}$
 & \textbf{94.55}$_{\mathbf{0.46}}$ & \textbf{90.69}$_{\mathbf{1.23}}$ & \textbf{92.95}$_{\mathbf{0.64}}$ & \textbf{92.12}$_{\mathbf{0.13}}$ & \textbf{90.48}$_{\mathbf{0.52}}$ \\
\bottomrule
\end{tabular}
\end{adjustbox}
\vspace{-10pt}
\end{table}

\vspace{-10pt}
\begin{table}[t]
\centering
\scriptsize
\renewcommand{\arraystretch}{1.06}
\setlength{\tabcolsep}{3.2pt}
\caption{
Comparison of finetuning methods on commonsense reasoning with Qwen3.5-9B under the Commonsense-15K protocol
(Appendix~\ref{app:impl}). ``Adapter'' is the total stored adapter state,
including frozen factors. Time is the end-to-end wall-clock training time in
\emph{mm:ss}, and accuracy cells are means over five seeds with standard
deviations as subscripts. RTOFT-B uses $\tau=0.10$, and the one-epoch row differs
only in its restoration schedule. Bold and underlined values indicate the best and second-best results among the parameter-efficient methods, respectively. Full finetuning, which skips the evaluation and checkpointing that the other rows run every $233$ steps,
and RTOFT-B without restoration are shown for reference.
}
\label{tab:commonsense}
\begin{adjustbox}{max width=\linewidth}
\begin{tabular}{lcccc cccccccc c}
\toprule
& \multicolumn{4}{c}{\textbf{Efficiency}}
& \multicolumn{9}{c}{\textbf{Commonsense reasoning}} \\
\cmidrule(lr){2-5}
\cmidrule(lr){6-14}
Method
& \#Param
& Adapter
& Mem.\,(GB)
& Time $\downarrow$
& BoolQ
& PIQA
& SIQA
& HellaSwag
& WinoG.
& ARC-e
& ARC-c
& OBQA
& Avg. \\
\midrule

Full finetuning
& 8.95B
& 8.95B
& 52.09
& 118:02
& 71.84$_{0.73}$
& 90.19$_{0.39}$
& 77.54$_{0.47}$
& 93.89$_{0.21}$
& 81.09$_{0.42}$
& 96.10$_{0.13}$
& 92.51$_{0.16}$
& 92.90$_{0.42}$
& 87.01$_{0.09}$ \\

\midrule
\multicolumn{14}{l}{\emph{Low-rank adaptation}} \\

LoRA$_{r=8}$
& 12.19M
& 12.19M
& 26.69
& 140:00
& \underline{73.09}$_{0.26}$
& 90.53$_{0.25}$
& 78.40$_{0.29}$
& 94.34$_{0.14}$
& 76.48$_{0.33}$
& \underline{98.32}$_{0.06}$
& 93.77$_{0.22}$
& 93.02$_{0.24}$
& 87.24$_{0.11}$ \\

PiSSA$_{r=8}$
& 12.19M
& 12.19M
& 26.69
& \textbf{133:58}
& 72.11$_{0.23}$
& \underline{91.41}$_{0.16}$
& \textbf{79.38}$_{0.25}$
& \underline{94.98}$_{0.10}$
& 78.92$_{0.61}$
& 97.94$_{0.18}$
& \underline{93.79}$_{0.27}$
& \underline{93.41}$_{0.43}$
& \underline{87.74}$_{0.18}$ \\

DoRA$_{r=8}$
& 12.99M
& 12.99M
& 29.78
& 153:53
& 72.75$_{0.39}$
& 90.59$_{0.39}$
& 78.20$_{0.29}$
& 94.44$_{0.17}$
& 76.56$_{0.55}$
& 98.27$_{0.16}$
& 93.77$_{0.33}$
& 93.30$_{0.24}$
& 87.24$_{0.15}$ \\

LoRA-XS$_{r=316}$
& 11.18M
& 492.68M
& \underline{26.66}
& 134:09
& 72.97$_{0.35}$
& 90.32$_{0.22}$
& 79.17$_{0.37}$
& 94.55$_{0.11}$
& \textbf{79.95}$_{0.29}$
& 97.47$_{0.17}$
& 93.17$_{0.11}$
& 92.97$_{0.24}$
& 87.57$_{0.09}$ \\

\midrule
\multicolumn{14}{l}{\emph{Orthogonal finetuning}} \\

HRA$_{r=16}$
& 11.53M
& 11.53M
& OOM
& N/A
& \multicolumn{8}{c}{N/A}
& N/A \\

OFTv2$_{b=32}$
& \textbf{11.17M}
& \textbf{11.17M}
& 26.95
& 174:50
& 71.59$_{0.48}$
& 91.02$_{0.27}$
& 77.89$_{0.33}$
& 94.23$_{0.09}$
& 76.09$_{0.38}$
& 98.23$_{0.15}$
& 93.69$_{0.42}$
& 91.80$_{0.55}$
& 86.82$_{0.17}$ \\

\midrule
\rowcolor{gray!10}
RTOFT-B$_{b=32}$ (one-epoch period)
& \textbf{11.17M}
& \textbf{11.17M}
& \textbf{26.58}
& \underline{134:05}
& 68.38$_{0.33}$
& 89.77$_{0.26}$
& 78.71$_{0.25}$
& 94.92$_{0.13}$
& \underline{78.93}$_{0.53}$
& 98.06$_{0.15}$
& 93.69$_{0.39}$
& \textbf{95.40}$_{0.49}$
& 87.23$_{0.21}$ \\

\rowcolor{gray!10}
\textbf{RTOFT-B$_{b=32}$ (ours)}
& \textbf{11.17M}
& \textbf{11.17M}
& \textbf{26.58}
& 134:18
& \textbf{73.61}$_{0.21}$
& \textbf{91.57}$_{0.25}$
& \underline{79.27}$_{0.38}$
& \textbf{95.03}$_{0.16}$
& \textbf{79.95}$_{0.69}$
& \textbf{98.34}$_{0.22}$
& \textbf{93.86}$_{0.22}$
& \textbf{95.40}$_{0.49}$
& \textbf{88.38}$_{0.24}$ \\

RTOFT-B$_{b=32}$ (no restoration)
& 11.17M & 11.17M & 26.58 & 134:04
& 71.65$_{0.39}$ & 88.84$_{0.15}$ & 77.85$_{0.29}$ & 93.94$_{0.15}$ & 77.47$_{0.86}$ & 97.52$_{0.12}$ & 92.74$_{0.06}$ & 94.74$_{0.27}$
& 86.84$_{0.15}$ \\

\bottomrule
\end{tabular}
\end{adjustbox}
\end{table}

\paragraph{Training time across parameterizations.}
With the one-epoch period, the training time relative to LoRA decreases from $1.10\times$ to $0.97\times$ for the block
parameterization, from $1.23\times$ to $1.03\times$ for the subspace
parameterization, and from $1.58\times$ to $1.12\times$ for the butterfly
parameterization (Tables~\ref{tab:glue-test}, \ref{tab:psoft},
and~\ref{tab:tboft}). With the drift trigger, the subspace and butterfly variants stay within $1\%$ of these times and gain $0.21$ and $0.28$ points over the period on the six-task average (Appendix~\ref{app:generality}). HRA is a chain of Householder reflections, to which the tangent update does not
apply, and it takes more than twice the training time of LoRA.

The reduction comes from not constructing an orthogonal transform in the forward pass. The reference implementation of OFTv2 constructs the transform in every forward pass and differentiates through it, whereas RTOFT applies the tangent map and computes a polar factor only at a restoration. Table~\ref{tab:step-time} reports the time of a training step of the main methods on the same GPU without evaluation. For the block parameterization at $b=32$, for which the construction is the cheapest of the three, a step of OFTv2 at its default $k=5$ takes longer than a step of RTOFT at every sequence length, up to $1.46$ times as long. Caching the transform of OFTv2 during evaluation changes its training time by at most $2.0\%$ (Appendix~\ref{app:launch}), so evaluation does not explain the reduction. The training times of RTOFT include the time for reading the drift.
Operation counts and further timing results are provided in
Appendices~\ref{app:glue_runtime}, \ref{app:launch}, and~\ref{app:cost}.

\vspace{-3pt}
\subsection{Retention on PEFT-Arena}
\label{sec:arena}
\vspace{-3pt}

PEFT-Arena \citep{huang2026peftarena} evaluates both the performance on a target task
and the general capability that the finetuned model retains. At the same number of trainable parameters, RTOFT loses
$2.35$ points of general capability, compared with $2.60$ for OFTv2 at
block size $32$, labeled OFT in PEFT-Arena, and achieves a larger gain on the target task
(Figure~\ref{fig:hero}c and Table~\ref{tab:arena-full}). Over three
runs of each method, the lowest general capability of RTOFT is higher than the highest of OFTv2.

Among the other methods in Table~\ref{tab:arena-full}, the two that retain more general capability than RTOFT,
VeRA and IA$^3$, have a much smaller gain or no gain on the target task. The two that have a larger gain on the target task, full finetuning and OFTv2 at block size $128$, lose considerably
more general capability and use at least four times as many trainable parameters.

\vspace{-3pt}
\subsection{Commonsense Reasoning and Image Generation}
\label{sec:transfer}
\vspace{-3pt}

\paragraph{Qwen3.5-9B and Llama-3.1-8B.}
\label{sec:commonsense}
On Qwen3.5-9B under the Commonsense-15K protocol, OFTv2 takes
$174{:}50$ of training time and reaches an eight-task average of $86.82$,
whereas RTOFT at the same number of trainable parameters takes $134{:}18$ and reaches $88.38$
(Table~\ref{tab:commonsense}). The one-epoch period takes about the same time as the trigger ($134{:}05$) and reaches $87.23$. The mean of the trigger over five seeds is higher than that of the period on seven of the eight tasks and equal on OBQA, and the difference is the largest on BoolQ ($73.61$ versus $68.38$). On BoolQ, the model trained with the period answers ``true'' on up to $90.9\%$ of the items, compared with $66.3\%$ for the trigger, while $62.2\%$ of the labels are ``true''
(Appendix~\ref{app:commonsense}). In a separate batch of five seeds, the trigger again exceeds the period, by $0.49$ on the eight-task average ($95\%$ confidence interval (CI) $[0.23,0.76]$, positive on every seed) and by $2.24$ on BoolQ ($[0.06,4.43]$). We also compare the two schedules on Llama-3.1-8B with RTOFT-B at $b=32$ over five paired seeds. The trigger exceeds the period by $1.02$ on the eight-task average ($[0.13,1.91]$, positive on every seed), with a standard deviation over seeds of $0.09$ compared with $0.69$. The largest difference is on HellaSwag, where the trigger exceeds the period by $10.81$ ($[3.92,17.70]$), with standard deviations of $0.38$ and $5.44$ (Appendix~\ref{app:commonsense-llama}).

\vspace{-10pt}
\paragraph{Stable Diffusion.}
\label{sec:diffusion}
\begin{table}[t]
\centering
\caption{
Comparison of finetuning methods on controllable generation with Stable Diffusion
(S2I: segmentation-to-image on ADE20K; L2F: landmark-to-face on CelebA-HQ). Each method is trained and timed at its best
trainable-parameter budget over a $1$--$26$M sweep, except for HRA, which uses the configuration of \citet{yuan2024hra}. COFT is the constrained variant of OFT, $c$ and $bn$ are numbers of blocks, $m$ is the number of factors of BOFT, and mIoU, mAcc, and aAcc are the mean intersection over union, the mean accuracy, and the average accuracy. Time is the wall-clock time on S2I
relative to LoRA. RTOFT-B uses $\tau=0.10$, and the one-epoch row differs
only in its restoration schedule.
}
\label{tab:diffusion}
\vspace{-6pt}
\fontsize{5.7}{6.7}\selectfont
\renewcommand{\arraystretch}{1.02}
\setlength{\tabcolsep}{1.0pt}
\resizebox{0.62\linewidth}{!}{%
\begin{tabular}{@{}l r c ccc c@{}}
\toprule
Method & \#P.\,(M) & Time & \multicolumn{3}{c}{S2I $\uparrow$} & L2F $\downarrow$ \\
\cmidrule(lr){4-6}\cmidrule(l){7-7}
 & & $\times$LoRA & mIoU & mAcc & aAcc & Error \\
\midrule
LoRA$_{r=16}$ & 2.50 & 1.00 & 24.58 & 37.74 & 60.31 & 8.01 \\
\midrule
OFT$_{4\,\mathrm{blocks}}$ & 26.40 & 1.11 & 27.18 & 40.23 & 62.58 & 7.03 \\
COFT$_{4\,\mathrm{blocks}}$ & 26.40 & 1.11 & 26.81 & 39.95 & 62.82 & 6.97 \\
OFTv2$_{c=2}$ & 26.34 & 1.09 & 27.79 & 40.78 & 64.36 & 6.61 \\
BOFT$^{m=4}_{bn=16}$ & 26.40 & 1.44 & 29.01 & 41.43 & 67.94 & 5.60 \\
HRA$_{r=8,\lambda=10^{-3}}$ & 0.89 & 1.53 & 29.19 & 40.80 & 67.73 & 5.51 \\
\midrule
RTOFT-B$_{2\,\mathrm{blocks}}$ (no restoration)
& 26.35 & 1.01 & 26.32 & 40.91 & 64.90 & 6.86 \\
\rowcolor{gray!6}
RTOFT-B$_{2\,\mathrm{blocks}}$ (one-epoch period)
& 26.35 & \underline{1.01}
& \underline{29.45} & \underline{42.26}
& \underline{67.97} & \underline{5.36} \\
\rowcolor{gray!12}
\textbf{RTOFT-B$_{2\,\mathrm{blocks}}$ (ours)}
& 26.35 & 1.02
& \textbf{30.18} & \textbf{42.78}
& \textbf{69.19} & \textbf{5.31} \\
\bottomrule
\end{tabular}%
}
\end{table}

Table~\ref{tab:diffusion} compares the methods on controllable generation with Stable Diffusion. With the one-epoch period, RTOFT takes $1.01\times$ the training time of LoRA, less
than every other orthogonal method, and improves on OFTv2 at about the same number of trainable parameters by $1.66$ mIoU on segmentation-to-image. With
the drift trigger at the same $\tau=0.10$ as in the language experiments, mIoU increases further from
$29.45$ to $30.18$, mAcc from $42.26$ to $42.78$, and aAcc from $67.97$
to $69.19$, and the training time increases only from $1.01\times$ to
$1.02\times$ that of LoRA. The trigger also has the lowest landmark-to-face
error of $5.31$, compared with $5.36$ for the period. Training times, additional baselines, and generated examples
are given in Tables~\ref{tab:diffusion-runtime} and
\ref{tab:diffusion-full} and Appendix~\ref{app:qualitative}.

\vspace{-5pt}
\section{Ablation Studies}
\label{sec:ablation}
\vspace{-5pt}

\vspace{-3pt}
\subsection{Training with and without Restoration}
\label{sec:restart-necessary}
\vspace{-3pt}

Table~\ref{tab:restart-summary} compares the tangent update with and without restoration in seven settings, where the reference of each column is the method in its header. With restoration, RTOFT has the best result in all seven. Without restoration, it falls below the reference in five of them, and the two exceptions are
landmark-to-face and commonsense reasoning. Restoration improves the six-task GLUE average, the commonsense average, and the general capability on PEFT-Arena (Table~\ref{tab:arena-full}) by $1.68\pm0.37$, $0.39$, and
$10.82$ points, respectively, and the improvement on GLUE is positive on all five shared seeds.

\begin{table}[H]
\centering
\small
\renewcommand{\arraystretch}{1.12}
\setlength{\tabcolsep}{5.2pt}

\caption{Comparison of the tangent update with and without restoration in seven settings.
The row with restoration uses the one-epoch period throughout, and the drift
trigger is evaluated in Section~\ref{sec:experiments}.}
\label{tab:restart-summary}

\resizebox{\linewidth}{!}{%
\begin{tabular}{lccccccc}
\toprule
& \multicolumn{3}{c}{OFTv2}
& \multicolumn{2}{c}{OFT}
& PSOFT
& BOFT \\
\cmidrule(lr){2-4}
\cmidrule(lr){5-6}
\cmidrule(lr){7-7}
\cmidrule(lr){8-8}

Variant
& \begin{tabular}[c]{@{}c@{}}GLUE\\6-task avg.\ $\uparrow$\end{tabular}
& \begin{tabular}[c]{@{}c@{}}Common.\\8-task avg.\ $\uparrow$\end{tabular}
& \begin{tabular}[c]{@{}c@{}}Arena\\retention $\uparrow$\end{tabular}
& \begin{tabular}[c]{@{}c@{}}S2I\\mIoU $\uparrow$\end{tabular}
& \begin{tabular}[c]{@{}c@{}}L2F\\error $\downarrow$\end{tabular}
& \begin{tabular}[c]{@{}c@{}}Subspace\\6-task avg.\ $\uparrow$\end{tabular}
& \begin{tabular}[c]{@{}c@{}}Butterfly\\6-task avg.\ $\uparrow$\end{tabular} \\
\midrule

Reference
& 88.81
& 86.82
& $-2.60$
& 27.18
& 7.03
& 88.01
& 87.76 \\

Tangent update, no restoration
& 87.77
& 86.84
& $-13.35$
& 26.32
& 6.86
& 87.03
& 87.07 \\

\textbf{Tangent update, restoration} (one-epoch period)
& \textbf{89.45}
& \textbf{87.23}
& $\mathbf{-2.53}$
& \textbf{29.45}
& \textbf{5.36}
& \textbf{88.19}
& \textbf{88.31} \\

\bottomrule
\end{tabular}%
}
\vspace{-10pt}
\end{table}

Making only the returned transform orthogonal does not give this gain. On PEFT-Arena, a polar projection after training without restoration recovers only $0.24$
of the $10.82$ points. On CoLA, projecting the checkpoints of a run without restoration at five points of training does not improve the score either
(Appendix~\ref{app:ablation-detail}). Restoration thus works by keeping the iterates close to the orthogonal group during training, beyond making the returned transform orthogonal.

The block, subspace, and butterfly parameterizations have a fixed \emph{support}, that is, $S$ ranges over a fixed linear space of skew-symmetric matrices, and after a restoration the gradient with respect to the parameters of $S$ is zero only if the loss has no first-order change in this space (Proposition~\ref{prop:reset-signal}). We refer to the variant that builds $S=UV^\top-VU^\top$ from two trainable factors $U$ and $V$ as RTOFT-L, whose support changes during training. Its restoration resets $V$ to zero, the gradient with respect to $U$ is proportional to $V$, and restoration reduces its six-task
average by $0.35\pm0.22$ points (Table~\ref{tab:variants}).
The rest of this section considers parameterizations with a fixed support. Appendix~\ref{app:restoration-roles} provides further evidence on how restoration works, including its interaction with the moments of Adam.
\vspace{-3pt}
\subsection{Truncation Order with and without Restoration}
\label{sec:order-restart}
\vspace{-3pt}
Table~\ref{tab:order-restart} compares the tangent map ($k=2$) with the default truncation of OFTv2 ($k=5$), each trained with and without restoration. Without restoration, $k=5$ is better by $1.04$
points. With restoration,
$k=2$ is better by $0.86$ points and needs three block multiplications fewer
per step.

\begin{table}[H]
\vspace{-10pt}
\centering
\footnotesize
\setlength{\tabcolsep}{5pt}
\renewcommand{\arraystretch}{1.08}

\caption{Six-task GLUE averages over five seeds of the tangent map ($k{=}2$) and the default Cayley--Neumann truncation of OFTv2 ($k{=}5$), each trained without restoration and with the one-epoch period.}
\label{tab:order-restart}

\begin{tabular}{lccc}
\toprule
Map & No restoration & With restoration & Restoration gain \\
\midrule
Tangent, $k{=}2$
& 87.77
& \textbf{89.45}
& $+1.68$ \\

Cayley--Neumann, $k{=}5$
& \textbf{88.81}
& 88.59
& $-0.22$ \\

\midrule
$k{=}5$ minus $k{=}2$
& $+1.04$
& $\mathbf{-0.86}$
& -- \\
\bottomrule
\end{tabular}

\vspace{-10pt}
\end{table}

Compared with $k=2$, the defect that restoration corrects is smaller at $k=5$. At $\|Q\|_2=0.16$, the value reached on CoLA, the defect of the $k=5$ map is
$78$ times smaller than that of $k=2$ by Lemma~\ref{lem:cnp}. Restoration improves $k=5$ by $0.16$ to $0.52$ points on five of the six tasks and reduces it by $3.03$ points on RTE, so its average change is negative (Appendix~\ref{app:ablation-detail}). Table~\ref{tab:order-restart} also shows that the improvement of RTOFT over OFTv2 requires both the tangent map and restoration, because neither the tangent map without restoration nor the map of OFTv2 with restoration achieves it. Restoration makes the cheaper tangent map usable, because each restoration multiplies the accumulated rotation into $W$ and resets $S$, so every cycle starts from an exactly orthogonal transform.

\vspace{-3pt}
\subsection{Drift under a Fixed Period and under the Drift Trigger}
\label{sec:drift-transfer}
\vspace{-3pt}
\begin{figure}[t]
\centering
\includegraphics[width=\linewidth]{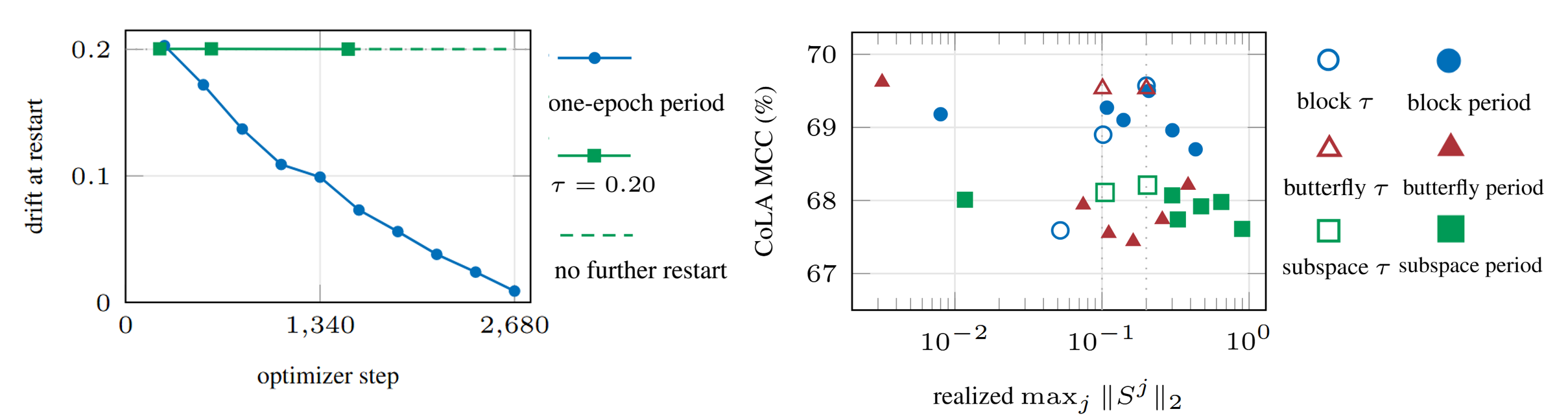}
\vspace{1pt}

\begin{minipage}[t]{0.49\linewidth}
\centering
{\scriptsize
(a) drift at each restoration of one run
}
\end{minipage}
\hfill
\begin{minipage}[t]{0.49\linewidth}
\centering
{\scriptsize
(b) score against the largest drift
}
\end{minipage}

\caption{Drift under a fixed period and under the drift trigger.
\textbf{(a)} Drift at each restoration (``restart'' in the figure) of one CoLA run. Under the one-epoch period, it
decreases from $0.203$ to $0.009$. Under the trigger with $\tau=0.20$, it stays near $\tau$, and the intervals between restorations become longer.
\textbf{(b)} Score on CoLA against the largest drift $\max_j\|S^j\|_2$, where $S^j$ is $S$ after step $j$, for the block, subspace, and butterfly parameterizations under fixed periods (filled markers) and drift triggers (open markers). At the same tolerance, the three parameterizations reach about the same largest drift.
The results on RTE are given in Figure~\ref{fig:law-full}.}
\label{fig:saw}
\vspace{-10pt}
\end{figure}

A fixed period does not determine the drift between restorations. Figure~\ref{fig:saw}a shows that under the one-epoch period, the drift reached before each restoration decreases more than tenfold during a run, whereas under the trigger it stays near $\tau$ and the
interval between restorations becomes longer. Figure~\ref{fig:saw}b shows that the drift also differs greatly across the three parameterizations under a fixed period, and that it stays near $\tau$ under the trigger.

The same is observed across models. With $\tau=0.10$,
the largest drift read during training stays between $0.10$ and $0.12$ across
DeBERTaV3, Qwen, and diffusion models (Figure~\ref{fig:hero}d, and Appendix~\ref{app:ablation-detail} for DeBERTaV3). Under the one-epoch period, it ranges from $0.17$ to $0.80$ across four settings with
block size $b=32$, so that the defects differ by a factor of twenty-two, and it is $1.65$ on segmentation-to-image (Appendix~\ref{app:ablation-detail}). The numbers of restorations under the trigger differ by orders of magnitude. By Proposition~\ref{prop:pathlength}, this number is at most the total update length of the run divided by $\tau$, whereas the number under a period is fixed by the number of training steps.

The drift read after an update can exceed $\tau$ by the growth of one step, and it triggers a restoration
before the next forward pass. For example, on
segmentation-to-image, the trigger at $\tau=0.10$ restores $1{,}872$
times in $27{,}820$ steps, and every forward pass stays within the defect bound $\tau^2$
(Appendix~\ref{app:impl}).

We further examine whether the number of restorations or their timing matters. In a separate set of SST-2 runs under the test-split protocol, we vary the learning rate from half to twice its tuned value (Table~\ref{tab:lr-stress}). The improvement of the trigger over the one-epoch period is the largest at twice the tuned rate,
where it is $1.69$ points and positive on every paired
seed. At this rate, the best seed of the period is below the worst seed of the trigger. Restoring after every step
performs as well as the trigger, and a fixed period with the same number of restorations as the trigger
recovers less than half of the $1.69$ points. The remaining difference comes from the timing of the restorations.

\paragraph{Choice of the tolerance.}
The tolerance can be chosen from the stretch budget (Corollary~\ref{cor:budget}), and the results are not very sensitive to it. None of the scores on the GLUE tasks decreases by more than one standard error when $\tau$ increases from $0.10$ to $0.20$, and, relative to $\tau=0.10$, the eight-task average on Qwen3.5-9B changes by at most $0.07$ points at $95\%$ confidence at $\tau=0.05$ and at $\tau=0.20$. Restoring after every step performs as well as the trigger and costs $4.5$--$27.5\%$ more per step (Appendix~\ref{app:launch}). Removing restoration reduces the general capability on PEFT-Arena (Section~\ref{sec:restart-necessary}). Replacing restoration by an $L_2$ penalty on $S$ either leaves the drift above $\tau$ or reduces the score on CoLA by up to $13.9$ points (Appendix~\ref{app:ablation-detail}). Because Corollary~\ref{cor:tau} holds for each matrix, the trigger can also be applied to each adapted matrix separately. With the same $\tau$, such a trigger needs $60\%$ fewer restorations of single matrices, and its accuracy is within the variation over seeds (Appendix~\ref{app:ablation-detail}). The tolerance could also be chosen for each layer.

\section{Related Work}

Orthogonal parameterizations keep the constraint in one of three ways. The first evaluates an exact map, such as the Cayley transform or the matrix exponential, at every update, and dynamic trivialization does so around the current iterate \citep{lezcanocasado2019trivializations}. The second lets the iterates leave the group with a bounded deviation, as in Landing \citep{ablin2022landing,ablin2024infeasible}, and POGO addresses the lack of restoration in such methods \citep{javaloy2026simple}. The third projects the iterates back onto the group, after every step in geometric numerical integration \citep{hairer2006geometric} and on a preset schedule for near-orthogonal weights \citep{higham1997orthonormality,jia2017svb} and for training that merges the orthogonal transform into the weight and reinitializes it \citep{qiu2025poet}. In orthogonal finetuning, OFTv2 fixes the truncation order of a Cayley--Neumann series \citep{qiu2025orthogonal}, and LoCO composes rotations through a parallel approximation and bounds its deviation from the group a priori \citep{nguyen2026loco}. In the Lanczos algorithm, it is classical to monitor the loss of orthogonality, which comes from finite-precision arithmetic, and to reorthogonalize only when it crosses a threshold \citep{parlett1979lanczos,simon1984lanczos}. RTOFT applies this rule to an adapter that by design is not orthogonal between restorations. Appendix~\ref{app:related} discusses further related work.

\section{Conclusion}
\label{sec:conclusion}

\vspace{-5pt}

This paper proposes RTOFT, which does not construct an orthogonal transform at every step of orthogonal finetuning. Its training time is close to that of LoRA, and it preserves or improves the target performance and the retained general capability of orthogonal finetuning. To use RTOFT, a practitioner chooses the stretch budget $\varepsilon$, the largest fraction by which $I+S$ may lengthen any direction, and sets $\tau=\sqrt{(1+\varepsilon)^2-1}$. During training, $\|S\|_2$ is read after every update, exact orthogonality is restored before the next forward pass whenever it reaches $\tau$, and it is restored once more at the end of training. Every forward pass then stays within the budget, and the returned transform is exactly orthogonal. The rule applies to the block-diagonal, subspace, and butterfly parameterizations and to any other adapter whose matrices $S$ form a fixed linear space of skew-symmetric matrices. More broadly, when the violation of a constraint can be computed exactly from the current iterate, whether to enforce the constraint at intermediate iterates and how to satisfy it at the returned solution can be decided separately.

\raggedbottom

\bibliographystyle{iclr2027_conference}
\bibliography{reference}

\section*{AI use statement}

Two tools were used. Claude (Anthropic) helped with \LaTeX{} and checked and fixed the
mathematics, reading the derivations and proofs for mistakes and correcting them; ChatGPT (OpenAI) helped with
formatting and commented on wording. Every claim, proof, experimental result and conclusion was
then verified by the authors, who are responsible for the entire paper.

\clearpage
\appendix

\section{Additional Related Work}
\label{app:related}

\paragraph{Orthogonal finetuning.}
Existing parameterizations differ in how the orthogonal transform is constructed, and all of them construct it and differentiate through it at every training step. Table~\ref{tab:param-comparison} compares four of them with RTOFT. OFT \citep{qiu2023oft} uses the Cayley transform of block-diagonal skew-symmetric
matrices. OFTv2 \citep{qiu2025orthogonal} replaces
the matrix inverse by a truncated Neumann series, which is cheaper but leaves a nonzero
orthogonality defect. HRA \citep{yuan2024hra} uses a chain of $r$ Householder reflections, which is exactly
orthogonal but has to be evaluated sequentially. Its regularization strength $\lambda$ strikes a balance between adaptation
capacity and regularity. BOFT \citep{liu2024parameter} composes sparse butterfly factors into
a dense transform with few trainable parameters. PSOFT \citep{wu2025efficient} restricts the
adaptation to an $r$-dimensional subspace with $R=I+P_r^\top(R_r-I)P_r$, where $P_r$ has orthonormal rows and $R_r\in\mathcal{O}(r)$. LOFT
\citep{zhao2026loft} selects this subspace based on the task and keeps it fixed. LoCO \citep{nguyen2026loco}
composes rotations through a parallel approximation whose deviation from the group is bounded a priori.
None of these methods uses a monitored orthogonality defect of the adapter to decide when to restore orthogonality.

\paragraph{Other baselines.}
Other baselines in our tables include BitFit \citep{benzaken2022bitfit}, the H-Adapter and P-Adapter of \citet{houlsby2019parameter} and \citet{pfeiffer2021adapterfusion}, LoRA \citep{hu2022lora}, LoRA-GA \citep{wang2024lorga}, AdaLoRA \citep{zhang2023adalora}, LoRA-XS \citep{wang2024loraxs}, PiSSA \citep{meng2024pissa}, DoRA \citep{liu2024dora}, MiLoRA \citep{wang2025milora}, MiSS \citep{kang2026miss}, KeepLoRA \citep{luo2026keeplora}, VeRA \citep{kopiczko2024vera}, IA$^3$ \citep{liu2022ia3}, and GiVA \citep{giva}.

\paragraph{Orthogonality constraints in other settings.}
Dynamic trivialization moves the center of a local chart to the current iterate, but every update within the chart still evaluates an exact
map onto the group \citep{lezcanocasado2019trivializations}. Cayley SGD retracts with an iterative approximation of the Cayley transform, whose accuracy is set by the number of iterations \citep{li2020cayley}.
Neumann-series approximations to Cayley maps also appear in orthogonal recurrent networks \citep{mucllari2024ncgru}. MuonBP amortizes the orthogonalization of optimizer updates
\citep{khaled2025muonbp}. Riemannian adaptive methods extend Adam-type updates to products of manifolds \citep{becigneul2019riemannian}. RTOFT keeps the moments of the optimizer unchanged across a restoration, and the gradient with respect to $S$ then changes by an amount that is bounded by the drift (Proposition~\ref{cor:frame}).

In Landing, the iterates leave the manifold with a bounded deviation, and a potential that attracts them back toward it is added
\citep{ablin2022landing,ablin2024infeasible}. POGO \citep{javaloy2026simple} addresses the lack of restoration in such methods. Periodic restoration of
near-orthogonal weights predates PEFT \citep{higham1997orthonormality,jia2017svb}. In orthogonal model training, \citet{qiu2025poet} measure the orthogonality error but merge the orthogonal transform into the weight and reinitialize it after a preset number of steps. Projection methods for differential equations on manifolds take an unconstrained step and project the result onto the manifold after every step \citep[Sec.~IV.4]{hairer2006geometric}. Unlike these methods, RTOFT measures the defect and restores exactly when it reaches the defect bound.

\section{Algorithm and Proofs}
\label{app:proof}

Algorithm~\ref{alg:rtoft} summarizes the training procedure of RTOFT.

\begin{algorithm}[!htb]
\caption{RTOFT with the drift trigger. Under a fixed period $M$, the test $\|S_t\|_2\ge\tau$ is replaced by ``$M$ divides $t$''. Training without restoration omits the test and the polar factor in the last line. The update of $S_t$ is written for gradient descent, and any optimizer can be used.}
\label{alg:rtoft}
\begin{algorithmic}[1]
\REQUIRE pretrained weights $W^{\rm pre}$, step size $\eta>0$,
tolerance $\tau>0$, iterations $T$.
\STATE Initialize $W_0\leftarrow W^{\rm pre}$, \; $S_0\leftarrow0$.
\FOR{$t=1,\dots,T$}
\STATE
$S_t\leftarrow
S_{t-1}
-
\eta\,
{\rm skew}\!\left(
\nabla_S
\mathcal{L}\bigl(W_{t-1}(I+S_{t-1})\bigr)
\right)$
\IF{$\|S_t\|_2\ge\tau$}
\STATE
$W_t\leftarrow
W_{t-1}{\rm polar}(I+S_t)$,
\quad
$S_t\leftarrow0$
\COMMENT{restoration}
\ELSE
\STATE $W_t\leftarrow W_{t-1}$
\ENDIF
\ENDFOR
\STATE \textbf{return}
$W_T\,{\rm polar}(I+S_T)$
\end{algorithmic}
\end{algorithm}

\subsection{Spectral Properties of $I+S$}

\begin{lemma}
\label{lem:skew}
Let $S\in\mathrm{Skew}(d)$ have nonzero eigenvalues $\pm i\theta_1,\dots,\pm i\theta_\nu$ with
$\theta_j>0$ and a zero eigenvalue of multiplicity $d-2\nu$. Then
\begin{enumerate}\itemsep1pt
\item[(a)] $(I+S)^\top(I+S)=(I+S)(I+S)^\top=I-S^2=I+S^\top S\succeq I$;
\item[(b)] the singular values of $I+S$ are $\sqrt{1+\theta_j^2}$, each twice, and $1$ with
multiplicity $d-2\nu$. In particular, $\|I+S\|_2=\sqrt{1+\|S\|_2^2}$, $\sigma_{\min}(I+S)\ge1$, and
$I+S$ is invertible with $\|(I+S)^{-1}\|_2\le1$;
\item[(c)] $\det(I+S)=\prod_{j=1}^{\nu}(1+\theta_j^2)>0$, and $I-S$ is invertible;
\item[(d)] $\mathrm{polar}(I+S)=(I+S)(I-S^2)^{-1/2}$ is a function of $S$, and its eigenvalue on the eigenvector of $S$
for $i\theta$ is $e^{i\arctan\theta}$.
\end{enumerate}
\end{lemma}

\begin{proof}
Since $S$ is real and skew-symmetric, it is normal and hence unitarily diagonalizable with purely imaginary eigenvalues. Since
$S^\top=-S$, routine algebra yields (a). The matrix $I-S^2$ has
eigenvalue $1+\theta^2$ on the eigenvector of $S$ for $i\theta$, so the singular values of $I+S$ are
$\sqrt{1+\theta^2}$. The largest is $\sqrt{1+\max_j\theta_j^2}=\sqrt{1+\|S\|_2^2}$
because $\|S\|_2=\max_j\theta_j$, which shows (b). For (c), $\det(I+S)=\prod_j(1+i\theta_j)(1-i\theta_j)
=\prod_j(1+\theta_j^2)>0$, and $I-S$ is invertible because its eigenvalues $1\mp i\theta_j$ and $1$ are
nonzero. For (d), $(I+S)^\top(I+S)=I-S^2\succ0$ by (a), so the polar factor is
$(I+S)\big((I+S)^\top(I+S)\big)^{-1/2}=(I+S)(I-S^2)^{-1/2}$. Both factors are functions of $S$, and
on the eigenvector of $S$ for $i\theta$ the product has eigenvalue
$(1+i\theta)/\sqrt{1+\theta^2}=e^{i\arctan\theta}$.
\end{proof}

\subsection{Proof of Theorem~\ref{thm:main}}

We prove the claim at restorations by induction on $n$. For $n=0$, $W_0=W^{\mathrm{pre}}$ and $R_0=I\in\mathcal{SO}(d)$. Assume that
$W_n=W^{\mathrm{pre}}R_n$ with $R_n\in\mathcal{SO}(d)$, and let $\Phi_n=\mathrm{polar}(I+S)$ be the polar factor at the next restoration. The matrix
$\Phi_n$ is orthogonal, and by Lemma~\ref{lem:skew}(c),
$\det\Phi_n=\det(I+S)/|\det(I+S)|=1$, so $\Phi_n\in\mathcal{SO}(d)$. Hence
$W_{n+1}=W_n\Phi_n=W^{\mathrm{pre}}(R_n\Phi_n)$ with $R_{n+1}=R_n\Phi_n\in\mathcal{SO}(d)$. This completes the induction.

Within a cycle, since $R_n$ is orthogonal, Lemma~\ref{lem:skew}(a) gives
$A^\top A-I=(I+S)^\top R_n^\top R_n(I+S)-I=-S^2$, so
$\|A^\top A-I\|_2=\|S^2\|_2=\|S\|_2^2$, where the last equality holds because $S$ is normal. Moreover,
$W_nW_n^\top=W^{\mathrm{pre}}R_nR_n^\top(W^{\mathrm{pre}})^\top=W^{\mathrm{pre}}(W^{\mathrm{pre}})^\top$, and since
the row norms, the row inner products, and the singular values of a matrix $W$ are determined by $WW^\top$, they are preserved at restorations. \hfill$\square$

The weight acts as $y=Wx$, so its neurons are its rows and the adapter multiplies on the right. This is the transposed form of OFT, which writes $z=W^\top x$ with the neurons as columns. Since the angles between the rows are preserved at restorations, so is the hyperspherical energy of the rows \citep{liu2018mhe}, as in OFT \citep{qiu2023oft}. The following proposition gives further properties between restorations.

\begin{proposition}[Geometry between restorations]
\label{prop:inflight}
Let $S\in{\rm Skew}(d)$ and
$W_n=W^{\rm pre}R_n$ with $R_n\in\mathcal{SO}(d)$.
\begin{enumerate}
\item[(i)]
The distance from $I+S$ to the orthogonal matrices is
\[
    \min_{R\in\mathcal{O}(d)}\|I+S-R\|_2
    =
    \sqrt{1+\|S\|_2^2}-1,
\]
with the minimum attained by ${\rm polar}(I+S)$, and
\[
\begin{aligned}
&\big\|
W_n(I+S)(I+S)^\top W_n^\top
-
W^{\rm pre}(W^{\rm pre})^\top
\big\|_2 \\
&\hspace{3cm}\le
\sigma_{\max}(W^{\rm pre})^2\|S\|_2^2 .
\end{aligned}
\]

\item[(ii)]
For every singular-value index $i$,
\[
    \sigma_i(W^{\rm pre})
    \le
    \sigma_i\!\bigl(W_n(I+S)\bigr)
    \le
    \sqrt{1+\|S\|_2^2}\,
    \sigma_i(W^{\rm pre}).
\]
For every row vector $w_i$ of $W_n$,
\[
    \|w_i(I+S)\|_2^2
    =
    \|w_i\|_2^2+\|w_iS\|_2^2.
\]
If $W^{\rm pre}$ has full column rank, then
\[
    \kappa_2\!\bigl(W_n(I+S)\bigr)
    \le
    \sqrt{1+\|S\|_2^2}\,
    \kappa_2(W^{\rm pre}),
\]
where $\kappa_2$ denotes the ratio of the largest to the smallest singular value.
\end{enumerate}
\end{proposition}

\begin{proof}
\emph{(i)} For a nonsingular $X$ and any unitarily invariant norm, the minimum
of $\|X-R\|$ over orthogonal $R$ is attained at the orthogonal polar factor of $X$ and equals
$\|\Sigma-I\|$, where $\Sigma$ is the diagonal matrix of the singular values of $X$ \citep{fan1955metric,higham1986polar}.
By Lemma~\ref{lem:skew}(b), all singular values of $I+S$ are at least $1$, so the minimum in the spectral norm
is $\max_i|\sigma_i(I+S)-1|=\sqrt{1+\|S\|_2^2}-1$. For the weights,
\begin{align*}
W_n(I+S)(I+S)^\top W_n^\top-W^{\mathrm{pre}}(W^{\mathrm{pre}})^\top
&=W^{\mathrm{pre}}R_n\big[(I+S)(I+S)^\top-I\big]R_n^\top(W^{\mathrm{pre}})^\top\\
&=-\,W^{\mathrm{pre}}R_nS^2R_n^\top(W^{\mathrm{pre}})^\top ,
\end{align*}
where the first equality follows from $W_n=W^{\mathrm{pre}}R_n$ and $R_nR_n^\top=I$ and the second from Lemma~\ref{lem:skew}(a). The spectral norm of the right-hand side is at most $\|W^{\mathrm{pre}}\|_2^2\|S^2\|_2=\sigma_{\max}(W^{\mathrm{pre}})^2\|S\|_2^2$.

\emph{(ii)} Since $\sigma_i(AB)\le\sigma_i(A)\|B\|_2$ for any $A$ and $B$ \citep{hornjohnson1991topics,bhatia1997matrix},
$\sigma_i(W_n(I+S))\le\sigma_i(W_n)\|I+S\|_2=\sqrt{1+\|S\|_2^2}\,\sigma_i(W_n)$.
By Lemma~\ref{lem:skew}(b), $I+S$ is invertible with $\|(I+S)^{-1}\|_2\le1$, so applying the same inequality to
$W_n=\big(W_n(I+S)\big)(I+S)^{-1}$ gives
$\sigma_i(W_n)\le\sigma_i(W_n(I+S))\,\|(I+S)^{-1}\|_2\le\sigma_i(W_n(I+S))$.
The first claim then follows from $\sigma_i(W_n)=\sigma_i(W^{\mathrm{pre}})$, which holds because $R_n$ is orthogonal. If $W^{\mathrm{pre}}$ has full column rank,
then $\sigma_d(W_n(I+S))\ge\sigma_d(W^{\mathrm{pre}})>0$, so $W_n(I+S)$ has
full column rank, and dividing the upper bound on $\sigma_1$ by the lower bound on $\sigma_d$ gives the bound on
$\kappa_2(W_n(I+S))$. For the rows, Lemma~\ref{lem:skew}(a) gives
$\|w_i(I+S)\|_2^2=w_i(I+S)(I+S)^\top w_i^\top=w_i(I-S^2)w_i^\top=\|w_i\|_2^2+\|w_iS\|_2^2$.
\end{proof}

\subsection{Proof of Corollary~\ref{cor:tau}}

By the hypothesis, $\|S\|_2$ is evaluated after every optimizer step, and at the first step of a cycle with
$\|S\|_2\ge\tau$ the schedule restores and resets $S$ to $0$ before the next forward pass. Hence every forward pass uses either
an $S$ with $\|S\|_2<\tau$ or $S=0$, and Theorem~\ref{thm:main} gives
$\|A^\top A-I\|_2=\|S\|_2^2\le\tau^2$. For the second statement,
Proposition~\ref{prop:inflight}(i) gives
$\|(I+S)-\mathrm{polar}(I+S)\|_2=\sqrt{1+\|S\|_2^2}-1$, and $\sqrt{1+u}\le1+u/2$ for $u\ge0$ shows that
this is at most $\|S\|_2^2/2\le\tau^2/2$. Since the above proof uses no property of the
loss, the data, the model, or the optimizer, both bounds hold independently of them. \hfill$\square$

\subsection{Stretch of the Tangent Map}

\begin{corollary}[Stretch bound]
\label{cor:stretch}
A transform $A$ with orthogonality defect $\delta$ satisfies $\sigma_{\max}(A)\le\sqrt{1+\delta}$. Under the drift trigger of Corollary~\ref{cor:tau}, every
singular value of $I+S$ lies in $[1,\sqrt{1+\tau^{2}}\,]$ in every
forward pass, and each restoration returns all of them to one.
\end{corollary}

\begin{proof}
The first claim holds because $\sigma_{\max}(A)^{2}=\|A^\top A\|_2\le1+\|A^\top A-I\|_2$. By Lemma~\ref{lem:skew}(b), the singular values of
$I+S$ are $\sqrt{1+\theta_j^{2}}$ over the nonzero eigenvalues
$\pm i\theta_j$ of $S$, and $1$ if $S$ is singular, and $\theta_j\le\|S\|_2\le\tau$ in every forward pass. After a restoration, $S=0$, so every singular value of $I+S$ equals one, and the claim follows.
\end{proof}

\begin{corollary}[Tolerance for a stretch budget]
\label{cor:budget}
Suppose that no forward pass is allowed to apply an $I+S$
that stretches any direction by more than a factor of $1+\varepsilon$. The largest tolerance of the drift trigger that guarantees this for every sequence of updates is
\[
\tau(\varepsilon)=\sqrt{(1+\varepsilon)^{2}-1}=\sqrt{2\varepsilon+\varepsilon^{2}}.
\]
\end{corollary}

\begin{proof}
By Lemma~\ref{lem:skew}(b), $\|I+S\|_2=\sqrt{1+\|S\|_2^{2}}$, which is at most $1+\varepsilon$ if and only if $\|S\|_2\le\tau(\varepsilon)$. Every forward pass has $\|S\|_2<\tau$, so any $\tau\le\tau(\varepsilon)$ satisfies the budget. If $\tau>\tau(\varepsilon)$, an $S$ with $\tau(\varepsilon)<\|S\|_2<\tau$ triggers no restoration and exceeds the budget, whence the corollary holds.
\end{proof}

Because each restoration returns every singular value to one,
the budget applies to each forward pass and does not accumulate over cycles.

\subsection{Numerical Precision of the Restoration}
\label{app:precision}

Theorem~\ref{thm:main} holds in exact arithmetic. In practice, $\mathrm{polar}(I+S)$ is computed in closed form by the
eigendecomposition of each skew-symmetric block or equivalently by an SVD. It is multiplied into the weight in
at least single precision, so a restoration is exact up to rounding errors in that precision. To examine the effect of the precision, we compose the
$46$ restorations of a schedule with $\tau=0.05$ at $d=768$ and $b=32$, each with
$\|S\|_2=0.05$, and measure the deviation of the accumulated transform from orthogonality. The deviation is
$1.5\times10^{-5}$ in float32, which is $170$ times smaller than the bound $\tau^{2}=2.5\times10^{-3}$. It is $5.1\times10^{-2}$ in bfloat16, twenty times the bound,
and $6.0\times10^{-3}$ in float16, which also exceeds the bound. The polar factor and its product with the weight must therefore be computed in single precision or better even when the rest of the model is trained in
half precision, and our implementation does so.

\subsection{Numbers of Restorations}
\label{app:pathlength}

\begin{proposition}[Fewest restorations]
\label{prop:fewest}
Fix the sequence of updates $\Delta S_1,\Delta S_2,\dots$ of $S$. In a cycle that starts after step $u$, $S_t=\sum_{u<j\le t}\Delta S_j$ after step $t$, and the update length $\ell(u,t)=\sum_{u<j\le t}\|\Delta S_j\|_2$ is an upper bound on $\|S_t\|_2$. A schedule that restores after steps $0=s_0<s_1<s_2<\cdots$ is called \emph{length-bounded} if $\ell(s_i,t)<\tau$ whenever $s_i\le t<s_{i+1}$. Within any horizon, the drift trigger, which restores after step $t$ whenever $\|S_t\|_2\ge\tau$, performs no more restorations than any length-bounded schedule. In particular, if $\|\Delta S_j\|_2\le\eta D$ for every $j$, it restores no more often than any fixed period $M$ with $(M-1)\eta D<\tau$.
\end{proposition}

\begin{proof}
Let $c(u)=\min\{t>u:\ell(u,t)\ge\tau\}$, with $c(u)=\infty$ when no such $t$ exists, and let $g_i$ be the step after which the trigger performs its $i$-th restoration, with $g_0=0$. For completeness, we let $s_i=\infty$ (respectively, $g_i=\infty$) when the schedule (respectively, the trigger) performs fewer than $i$ restorations. For a fixed $t$, $\ell(u,t)$ does not increase in $u$, so $u\le u'$ implies $\ell(u,c(u'))\ge\ell(u',c(u'))\ge\tau$ and hence $c(u)\le c(u')$. A length-bounded schedule satisfies $s_{i+1}\le c(s_i)$ by definition. The restorations $g_i$ of the trigger satisfy $g_{i+1}\ge c(g_i)$, because $\|S_t\|_2\le\ell(g_i,t)$ and thus no reading before $c(g_i)$ reaches $\tau$. If $g_i\ge s_i$, then $g_{i+1}\ge c(g_i)\ge c(s_i)\ge s_{i+1}$. Since $g_0=s_0=0$, induction gives $g_i\ge s_i$ for every $i$, so the $i$-th restoration of the trigger never precedes that of the schedule, and the trigger performs at most as many restorations within any horizon. For the second claim, $\|\Delta S_j\|_2\le\eta D$ gives $\ell(u,t)\le(t-u)\eta D$. Under the period $s_i=iM$, every forward pass uses some $S_t$ with $t-s_i\le M-1$, so the period is length-bounded whenever $(M-1)\eta D<\tau$, and the claim follows.
\end{proof}

\begin{proposition}[Restoration count and total update length]
\label{prop:pathlength}
Let $L_T=\sum_{t\le T}\|\Delta S_t\|_2$ be the total update length of a run of $T$ steps. A trigger that restores only when a reading of the drift after an update is at least $\tau$ restores at most $\lfloor L_T/\tau\rfloor$
times, at whichever steps the drift is read, and at
most $\lfloor T/q\rfloor$ times if the drift is read every $q$ steps. A fixed period $M$ restores $\lfloor T/M\rfloor$ times, with
the last step counted, whatever $L_T$ is.
\end{proposition}

\begin{proof}
Every cycle starts at $S=0$. If a cycle ends with a restoration, the reading of the drift at its last step is at least $\tau$, and
by the triangle inequality the update length of this cycle is at least $\tau$. Since the completed cycles are disjoint and the last incomplete cycle
has a nonnegative update length, the number $N_R$ of restorations satisfies $N_R\,\tau\le L_T$, and hence
$N_R\le\lfloor L_T/\tau\rfloor$. The argument does not depend on when the drift is read, and a drift that is read every $q$ steps has at most
$\lfloor T/q\rfloor$ readings. The count of the period follows from its definition.
\end{proof}

For example, applied to the block-diagonal matrix of all adapted matrices, whose spectral norm is the drift that the trigger reads, Proposition~\ref{prop:pathlength} shows that the $10$, $40$, and $1{,}872$ restorations in Figure~\ref{fig:hero}d imply total update
lengths of at least $1.0$, $4.0$, and $187$ at $\tau=0.10$.

\begin{proposition}[Number of restorations of a length-bounded period]
\label{cor:howmany}
In the setting of Proposition~\ref{prop:fewest}, consider a run of $T$ steps, let $\alpha_j=\|\Delta S_j\|_2$, and let $\bar\alpha=L_T/T$ be the mean update norm. For a fixed period $M$ with $2\le M\le T$, let $\bar\alpha_M$ be the largest mean of $\alpha_j$ over the $M-1$ steps $iM<j<(i+1)M$, over all $i$ with $(i+1)M\le T$, and suppose that $\bar\alpha_M>0$. Then, before the end of training, a length-bounded period restores at least $T/(1+\tau/\bar\alpha_M)-1$ times, and the numbers of restorations $K_{\rm period}$ of this period and $K_{\rm trigger}$ of the trigger satisfy
\[
K_{\rm period}\;\ge\;\frac{\bar\alpha_M/\bar\alpha}{1+\bar\alpha_M/\tau}\,K_{\rm trigger}-1.
\]
\end{proposition}

\begin{proof}
Since the period is length-bounded, $\sum_{iM<j<(i+1)M}\alpha_j<\tau$ for every $i$ with $(i+1)M\le T$. For the $i$ that attains $\bar\alpha_M$, this gives $(M-1)\bar\alpha_M<\tau$, so $M<1+\tau/\bar\alpha_M$, and the period restores at least $T/M-1>T/(1+\tau/\bar\alpha_M)-1$ times before the end of training. By Proposition~\ref{prop:pathlength}, $K_{\rm trigger}\le L_T/\tau$. Since $T/(1+\tau/\bar\alpha_M)=(L_T/\tau)(\bar\alpha_M/\bar\alpha)/(1+\bar\alpha_M/\tau)$, the bound follows.
\end{proof}

\begin{figure}[t]
\centering
\includegraphics[width=\linewidth]{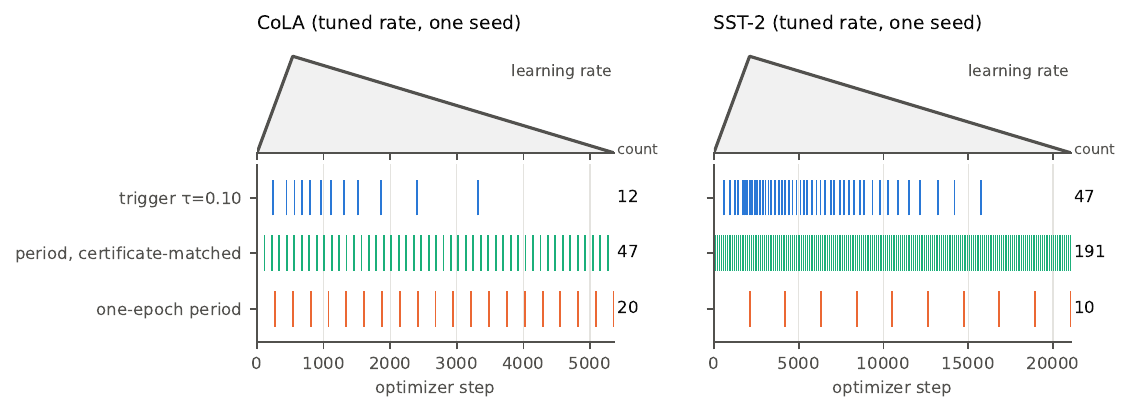}
\caption{Restorations of three schedules at the tuned learning rate for one seed. Each tick is one restoration, and the counts are given at the right. The trigger restores often while the learning rate is high and rarely after it has decayed. A fixed period set to the shortest interval of the trigger (``certificate-matched'' in the figure) restores $47$ times on CoLA and $191$ times on SST-2, compared with $12$ and $47$ times, respectively, for the trigger. The one-epoch period restores $20$ and $10$ times.}
\label{fig:restarts}
\end{figure}

The ratio $\bar\alpha_M/\bar\alpha$ compares the largest mean update norm over one period with the mean over the run, and the factor $1+\bar\alpha_M/\tau$ is close to one when a single update is small compared with $\tau$. If the update norm follows the learning rate, the ratio is about two under warmup followed by linear decay when the period is short compared with the run. In the batch of runs of Table~\ref{tab:tau020}, a fixed period set to the shortest interval between two restorations of the trigger restores $36$--$56$ times on CoLA and $161$--$239$ times on SST-2, compared with $11$--$13$ and $45$--$49$ times, respectively, for the trigger. On SST-2 at twice the tuned rate, it restores $1403$--$2631$ times, compared with $209$--$228$ times for the trigger. Figure~\ref{fig:restarts} shows one seed.

\subsection{Difference from Restoring after Every Step}
\label{app:tracking}

The following lemma compares one cycle of tangent updates with gradient descent that retracts onto the orthogonal group after every step, where both start from the same transform.

\begin{lemma}[Difference over one cycle]
\label{lem:tracking}
Let $F$ be the loss as a function of the transform, with $\|\nabla F(A)\|_2\le G$ and $\|\nabla F(A)-\nabla F(B)\|_2\le\beta\|A-B\|_2$ for all $A$ and $B$ in a convex
neighborhood of $\mathcal{O}(d)$ that contains all iterates below. Fix
$R\in\mathcal{O}(d)$, a step size $\eta$, and a cycle of $M$ steps. Consider
\emph{(a)} the tangent updates
$S_0=0$,
\[
S_{t+1}=S_t-\eta\,\mathrm{skew}\!\bigl(R^\top\nabla F(R(I+S_t))\bigr),
\qquad
R'=R\,\mathrm{polar}(I+S_M),
\]
and \emph{(b)} the updates that retract after every step, $X_0=R$,
\[
X_{t+1}=X_t\,\mathrm{polar}\!\bigl(I-\eta\,H_t\bigr),
\qquad
H_t=\mathrm{skew}\!\bigl(X_t^\top\nabla F(X_t)\bigr).
\]
Suppose that $\|S_t\|_2\le\tau$ for $t<M$,
$\eta G\le\tau\le G/(\beta+2G)$, and $\eta GM\le2\tau$. Then
\[
\bigl\|R\,\mathrm{polar}(I+S_M)-X_M\bigr\|_2
\;\le\;
9\,\tau^{2}\,e^{(2G+\beta)M\eta}
\;\le\;
9e^{2}\tau^{2}.
\]
\end{lemma}

\begin{proof}
Let $A_t=R(I+S_t)$, $U_t=\mathrm{polar}(I+S_t)$, $\tilde X_t=RU_t$,
$s_t=\|S_t\|_2$, and
$\mathcal{D}(X)=X\,\mathrm{skew}(X^\top\nabla F(X))$. We first introduce four facts.
First, the map $\mathrm{skew}$ is linear and $1$-Lipschitz in the spectral norm,
and $\mathrm{skew}(0)=0$. Second, by Lemma~\ref{lem:skew},
$\|(I+Z)-\mathrm{polar}(I+Z)\|_2=\sqrt{1+\|Z\|_2^{2}}-1\le\|Z\|_2^{2}/2$ for a
skew-symmetric $Z$. With $Z=-\eta H_t$, where $\|H_t\|_2\le G$, and since left multiplication by the
orthogonal $X_t$ preserves the norm, $X_{t+1}$ differs from $X_t(I-\eta H_t)$
by at most $\eta^{2}G^{2}/2$. With $Z=S_t$, we have
$\|A_t-\tilde X_t\|_2\le s_t^{2}/2$. Third, the eigenvalues of $U_t$ are
$e^{i\arctan\theta_j}$ over the eigenvalues $\pm i\theta_j$ of $S_t$
(Lemma~\ref{lem:skew}), so $\|U_t-I\|_2\le\max_j\arctan\theta_j\le s_t$.
Fourth, on the group, $\mathcal{D}$ is Lipschitz with constant at most
$2G+\beta$. This is because $\mathcal{D}(X)-\mathcal{D}(X')$ is the sum of $(X-X')\,\mathrm{skew}(X^\top\nabla F(X))$, $X'\,\mathrm{skew}((X-X')^\top\nabla F(X))$, and $X'\,\mathrm{skew}(X'^\top(\nabla F(X)-\nabla F(X')))$, whose norms are at most $G\|X-X'\|_2$, $G\|X-X'\|_2$, and $\beta\|X-X'\|_2$, respectively.

Since $\mathrm{skew}$ is $1$-Lipschitz and $\|\nabla F\|_2\le G$, $\|S_{t+1}\|_2\le\|S_t\|_2+\eta G$, so
$s_t\le t\eta G$ for $t\le M$, and $s_t\le\tau$ for $t<M$ by the hypothesis.

We next derive a recursion for $E_t=\|A_t-X_t\|_2$, where $E_0=0$. Since $X_{t+1}$ differs from $X_t(I-\eta H_t)$ by at most $\eta^2G^2/2$, the two
update rules give
\[
E_{t+1}
\le E_t
+\eta\,\bigl\|R\,\mathrm{skew}(R^\top\nabla F(A_t))-\mathcal{D}(X_t)\bigr\|_2
+\tfrac12\eta^{2}G^{2}.
\]
By the triangle inequality, the norm on the right-hand side is at most $\|R\,\mathrm{skew}(R^\top\nabla F(A_t))-\mathcal{D}(\tilde X_t)\|_2+\|\mathcal{D}(\tilde X_t)-\mathcal{D}(X_t)\|_2$. Let
$g=R^\top\nabla F(\tilde X_t)$ and $U_t=I+\Delta$, where $\|g\|_2\le G$ and
$\|\Delta\|_2\le s_t$. Since $\mathrm{skew}$ is linear,
\[
\mathcal{D}(\tilde X_t)
=R\bigl[\mathrm{skew}(g)+\mathrm{skew}(\Delta^\top g)
+\Delta\,\mathrm{skew}(g)+\Delta\,\mathrm{skew}(\Delta^\top g)\bigr].
\]
Adding and subtracting $\nabla F(\tilde X_t)$ in the first term, we obtain
\[
\bigl\|R\,\mathrm{skew}(R^\top\nabla F(A_t))-\mathcal{D}(\tilde X_t)\bigr\|_2
\le
\beta\,\tfrac{s_t^{2}}{2}+2G\,s_t+G\,s_t^{2},
\]
where the inequality holds since $\mathrm{skew}$ is $1$-Lipschitz, $\nabla F$ is $\beta$-Lipschitz, and $\|A_t-\tilde X_t\|_2\le s_t^2/2$.
Moreover, since $\mathcal{D}$ is Lipschitz with constant $2G+\beta$ on the group and $\|\tilde X_t-X_t\|_2\le E_t+s_t^2/2$, we have
$\|\mathcal{D}(\tilde X_t)-\mathcal{D}(X_t)\|_2
\le(2G+\beta)(E_t+\tfrac{s_t^{2}}{2})$. Combining the two bounds,
\[
E_{t+1}\le\bigl(1+\eta(2G+\beta)\bigr)E_t
+\eta\bigl(2G\,s_t+(2G+\beta)\,s_t^{2}\bigr)+\tfrac12\eta^{2}G^{2}.
\]

The discrete Gr\"onwall inequality then gives
$E_M\le e^{(2G+\beta)M\eta}\,\Sigma$ with
$\Sigma=\sum_{t<M}\bigl[\eta(2Gs_t+(2G+\beta)s_t^{2})+\tfrac12\eta^{2}G^{2}\bigr]$.
Because the cycle starts at $S_0=0$, $s_t\le t\eta G$ and the hypothesis $\eta GM\le2\tau$ give
\[
\sum_{t<M}\eta\,2G\,s_t
\le 2G\eta\cdot\eta G\,\tfrac{M^{2}}{2}
=(\eta GM)^{2}\le4\tau^{2}.
\]
The remaining sums satisfy $\sum_{t<M}\eta(2G+\beta)s_t^{2}\le(2G+\beta)\tau^{2}\,\eta M
\le2(2G+\beta)\tau^{3}/G\le2\tau^{2}$ by $\tau\le G/(\beta+2G)$, and
$\tfrac12 M\eta^{2}G^{2}=\tfrac12(\eta GM)(\eta G)\le\tau^{2}$ by
$\eta G\le\tau$. Hence $E_M\le7\tau^{2}e^{(2G+\beta)M\eta}$.

It remains to add the restoration at the end of the cycle. Since $\|(I+S_M)-\mathrm{polar}(I+S_M)\|_2\le s_M^2/2$ and $s_M\le\eta GM\le2\tau$, we have
$\|R\,\mathrm{polar}(I+S_M)-A_M\|_2\le2\tau^{2}$, so
$\|R'-X_M\|_2\le E_M+2\tau^{2}\le9\tau^{2}e^{(2G+\beta)M\eta}$. The
second inequality of the lemma follows from
$(2G+\beta)M\eta\le(2G+\beta)\,2\tau/G\le2$ under $\tau\le G/(\beta+2G)$.
\end{proof}

\subsection{Change of the Gradient at a Restoration}

\begin{proposition}[Change of the gradient]
\label{cor:frame}
Let $\Phi=\operatorname{polar}(I+S)$ for $S\in{\rm Skew}(d)$. Then
\[
  \|\Phi-I\|_2 = 2\sin\!\bigl(\tfrac12\arctan\|S\|_2\bigr)
  \;\le\; \arctan\|S\|_2 \;\le\; \|S\|_2 .
\]
At a fixed loss gradient, the gradients with respect to $S$ immediately before and after a restoration,
$\operatorname{skew}(W^{\top}\nabla_W\mathcal{L})$ and
$\operatorname{skew}(\Phi^{\top}W^{\top}\nabla_W\mathcal{L})$ with $W$ the weight before the restoration, differ by at most $\arctan\|S\|_2\,\|W^{\top}\nabla_W\mathcal{L}\|_2$.
\end{proposition}

\begin{proof}
By Lemma~\ref{lem:skew}(d), $\Phi$ has the eigenvalues $e^{\pm i\arctan\theta_j}$, and $1$ if $S$ is singular, so $\|\Phi-I\|_2=\max_j|e^{i\arctan\theta_j}-1|=2\sin(\tfrac12\arctan\|S\|_2)$, and the two inequalities are $2\sin(x/2)\le x$ and $\arctan x\le x$ for $x\ge0$. Since $\operatorname{skew}$ is $1$-Lipschitz in the spectral norm, the two gradients differ by at most $\|(\Phi-I)^{\top}W^{\top}\nabla_W\mathcal{L}\|_2\le\|\Phi-I\|_2\,\|W^{\top}\nabla_W\mathcal{L}\|_2\le\arctan\|S\|_2\,\|W^{\top}\nabla_W\mathcal{L}\|_2$, where the last inequality follows from the first claim.
\end{proof}

\subsection{Orthogonality Defect of Truncated Maps}

The following lemma and corollary hold for any approximation of the form $R\,e(Q)$, whatever the orthogonal map $R$, and Lemma~\ref{lem:cnp} gives the exact defect of the maps of OFTv2.

\begin{lemma}[Defect of an approximate orthogonal map]
\label{lem:parity}
Let $Q\in{\rm Skew}(d)$ with $\theta=\|Q\|_2$, let $R$ be any exactly orthogonal
function of $Q$, and let $\widetilde R=R\,e(Q)$, where $e$ is a power series with
real coefficients $a_0=1,a_1,a_2,\dots$ and a radius of convergence larger than $\theta$. Then
\[
    \widetilde R^\top\widetilde R-I \;=\; e(-Q)\,e(Q)-I ,
\]
so the defect depends on the error factor $e$ alone and not on $R$. Let
$c_{2s}=\sum_{j+l=2s}(-1)^{(j-l)/2}\,a_j a_l$ for $s\ge1$. If $c_{2s}\ne0$ for some $s$, let $2s^*$ be the smallest such index. Then, as $\theta\to0$,
\[
    \bigl\|\widetilde R^\top\widetilde R-I\bigr\|_2
    \;=\;
    |c_{2s^*}|\,\theta^{2s^*}+O(\theta^{2s^*+2}).
\]
If $c_{2s}=0$ for all $s\ge1$, the defect is zero.
\end{lemma}

\begin{proof}
Since $R^\top R=I$,
$\widetilde R^\top\widetilde R=e(Q)^\top R^\top R\,e(Q)=e(Q)^\top e(Q)$, and
$e(Q)^\top=e(Q^\top)=e(-Q)$ because $e$ has real coefficients. The matrices $e(-Q)$ and $e(Q)$ are
functions of the same matrix and commute, which gives the first display. Let
$h(z)=e(-z)e(z)-1$. Since $h(-z)=h(z)$, $h$ has only even powers. Over the eigenvalue
pair $\pm i\theta_\kappa$ of $Q$, the defect has the real eigenvalue
$h(i\theta_\kappa)=|e(i\theta_\kappa)|^{2}-1$, and expanding
$e(i\theta)e(-i\theta)$ gives the coefficient
$(-1)^{s}\sum_{j+l=2s}(-1)^{l}a_ja_l$ at order $2s$, which equals $c_{2s}$ because
$(j+l)/2+l$ and $(j-l)/2$ differ by the even number $2l$. The spectral norm is the largest
eigenvalue magnitude, which is attained to leading order at $\theta_\kappa=\theta$ and equals
$|c_{2s^*}|\theta^{2s^*}+O(\theta^{2s^*+2})$.
\end{proof}

\begin{corollary}[Order of the defect for even and odd powers]
\label{cor:parity}
Let $e(z)=1+a_pz^p+O(z^{p+1})$ with $a_p\neq0$. If $p$ is even, the defect is
of order $p$ with leading coefficient $2|a_p|$. If $p$ is odd, the term of order $p$ cancels.
For odd $p\ge3$, the defect is of order $p+1$ with leading coefficient $2|a_{p+1}|$ when
$a_{p+1}\neq0$, and of order $2p$ with leading coefficient $a_p^{2}$ when
$a_{p+1}=\dots=a_{2p}=0$. For $p=1$, the coefficient of order two is $|a_1^{2}-2a_2|$.
\end{corollary}

\begin{proof}
Since $a_1=\dots=a_{p-1}=0$, a pair $(j,l)$ with $0<j+l=2s<p$ has an index in $\{1,\dots,p-1\}$, so $c_{2s}=0$ for $0<2s<p$. If $p$ is even,
the pairs $(0,p)$ and $(p,0)$ give $c_{p}=2(-1)^{p/2}a_p$. For odd $p\ge3$, only $(0,p+1)$ and $(p+1,0)$ reach order $p+1$, which gives
$c_{p+1}=2(-1)^{(p+1)/2}a_{p+1}$, and when $a_{p+1}=\dots=a_{2p}=0$, the pairs $(0,2p)$ and $(2p,0)$ vanish
as well, so the first nonzero pair is $(p,p)$ with $c_{2p}=a_p^{2}$. For $p=1$, the pairs $(0,2)$, $(2,0)$, and $(1,1)$
give $c_{2}=a_1^{2}-2a_2$.
\end{proof}

At truncation order $k$, the reference implementation of OFTv2 accumulates the powers of $Q$ with coefficient two up to $Q^{\mu-1}$ and coefficient one on $Q^\mu$, where $\mu$ depends on $k$. For $k\in\{3,4\}$, its loop over the powers $3,\dots,k-2$ is empty while the last multiplication with coefficient one is always performed, so $\mu=3$. For $k\ge5$, the loop contributes the powers up to $k-2$, and the last multiplication gives $\mu=k-1$. For $k=2$, the implementation returns $I+2Q$ before the last term is added.

\begin{lemma}[Defect of the Cayley--Neumann map]
\label{lem:cnp}
Let $Q\in{\rm Skew}(d)$ have eigenvalues $\pm i\theta_j$, let
$\theta=\|Q\|_2=\max_j\theta_j$, and let $R=(I+Q)(I-Q)^{-1}$. At truncation order $k$, the
reference implementation of OFTv2 returns
\[
    \widetilde R_k
    =
    R\,(I-Q^{\mu}),
    \qquad
    \mu=\begin{cases}3, & k\in\{3,4\},\\[2pt] k-1, & k\ge5,\end{cases}
\]
and $\widetilde R_2=I+2Q$. The orthogonality defect is
\[
    \bigl\|\widetilde R_k^\top\widetilde R_k-I\bigr\|_2
    =
    \begin{cases}
        \theta^{2\mu}, & \mu \text{ odd},\\[3pt]
        2\theta^{\mu}-\theta^{2\mu}, & \mu\equiv0\!\!\pmod 4\text{ and }\theta\le1,\\[3pt]
        2\theta^{\mu}+\theta^{2\mu}, & \mu\equiv2\!\!\pmod 4,
    \end{cases}
\]
and $\|\widetilde R_2^\top\widetilde R_2-I\|_2=4\theta^{2}$. For
$\mu\equiv0\pmod 4$ and $\theta>1$, it is $\max_j|2\theta_j^{\mu}-\theta_j^{2\mu}|$.
\end{lemma}

\begin{proof}
Since
\[
    C(z)\,(1-z^{\mu})
    =
    (1+z)\bigl(1+z+\dots+z^{\mu-1}\bigr)
    =
    1+2z+2z^{2}+\dots+2z^{\mu-1}+z^{\mu},
    \qquad C(z)=\frac{1+z}{1-z},
\]
we have $\widetilde R_k=R\,(I-Q^{\mu})$ with $R=C(Q)$, which is well defined because
$I-Q$ is invertible by Lemma~\ref{lem:skew}(c).

For the defect, $R^\top R=I$ and $(Q^{\mu})^\top=(-1)^{\mu}Q^{\mu}$ give
\[
    \widetilde R_k^\top\widetilde R_k
    =
    \bigl(I-(-1)^{\mu}Q^{\mu}\bigr)\bigl(I-Q^{\mu}\bigr)
    =
    \begin{cases}
        I-Q^{2\mu}, & \mu \text{ odd},\\[3pt]
        (I-Q^{\mu})^{2}, & \mu \text{ even}.
    \end{cases}
\]
Both matrices are polynomials in $Q$, so the spectral norm can be computed from the eigenvalues
$\pm i\theta_j$ of $Q$. For $\mu$ odd, $(i\theta)^{2\mu}=-\theta^{2\mu}$ and the
eigenvalue is $1+\theta^{2\mu}$. For $\mu$ even,
$(i\theta)^{\mu}=(-1)^{\mu/2}\theta^{\mu}$ and the eigenvalue is
$(1\mp\theta^{\mu})^{2}$, whose difference from one is $\mp2\theta^{\mu}+\theta^{2\mu}$. For odd $\mu$ and for $\mu\equiv2\pmod 4$, the absolute difference is increasing in $\theta$, so the maximum
over $j$ is attained at $\theta=\|Q\|_2$. For $\mu\equiv0\pmod 4$, it is
$|2u-u^{2}|$ with $u=\theta_j^{\mu}$, which is increasing on $u\le1$, so the same holds for $\theta\le1$. For $k=2$, $(I-2Q)(I+2Q)=I-4Q^{2}$ has eigenvalue
$1+4\theta^{2}$, so the defect is $4\theta^2$, whence the lemma holds.
\end{proof}

The orders in Lemma~\ref{lem:cnp} also follow from Corollary~\ref{cor:parity}, with $e(z)=1-z^\mu$ and, for $k=2$, with
$e(z)=(1+2z)(1-z)/(1+z)=1-2z^{2}+O(z^{3})$. In particular, the
default $k=5$ has $\mu=4$ and a defect of order four, whereas
$k=3$ has $\mu=3$ and a defect of order six. In all our measurements, $\theta<1$.

\begin{table}[t]
\centering
\scriptsize
\renewcommand{\arraystretch}{1.04}
\setlength{\tabcolsep}{4pt}
\caption{Per-task training time on GLUE in \emph{mm:ss} under the setting of
Table~\ref{tab:glue-test}. The average is computed in seconds and converted back to \emph{mm:ss}.
Times are for whole runs with evaluation, and those of the drift trigger include the readings of the drift. Every method is timed with the block-diagonal skew-symmetric matrix built as in its own code. Bold and underlined values indicate the best and second-best results, respectively, and lower is better.}
\label{tab:glue-runtime-full}
\begin{adjustbox}{max width=\linewidth}
\begin{tabular}{lcc cccccc c}
\toprule
Method & \#Param & Mem.\,(GB) & CoLA & STS-B & MRPC & RTE & SST-2 & QNLI & Avg. \\
\midrule
Full finetuning & 184M & 5.9 & 25:24 & 32:35 & 100:12 & 65:02 & 215:40 & 455:24 & 149:03 \\
\midrule
LoRA$_{r=8}$        & 1.33M & 4.5 & 23:19 & \textbf{29:25} & 88:42 & 59:34 & 208:06 & 418:17 & 137:54 \\
PiSSA$_{r=8}$       & 1.33M & 4.5 & 24:03 & 32:22 & \textbf{81:07} & \textbf{54:56} & \textbf{174:48} & \textbf{395:44} & \textbf{127:10} \\
DoRA$_{r=8}$        & 1.41M & 5.8 & 27:25 & 33:56 & 87:04 & 61:02 & 201:43 & 419:38 & 138:28 \\
LoRA-GA$_{r=8}$     & 1.33M & 4.5 & 23:56 & \underline{29:47} & \underline{81:32} & \underline{55:34} & \underline{176:56} & \underline{398:03} & \underline{127:38} \\
LoRA-XS$_{r=136}$   & 1.33M & 4.2 & 23:33 & 32:40 & 86:26 & 58:55 & 185:32 & 412:56 & 133:20 \\
GiVA$_{r=8}$        & 0.09M & 4.7 & 23:46 & 32:31 & 87:20 & 60:18 & 189:47 & 476:28 & 145:02 \\
\midrule
PSOFT$_{r=46}$      & 0.08M & 4.1 & 54:11 & 48:04 & 96:41 & 65:19 & 259:52 & 495:07 & 169:52 \\
HRA$_{r=16}$        & 1.33M & 15.2 & 144:40 & 107:18 & 129:32 & 88:32 & 616:48 & 614:37 & 283:35 \\
BOFT$^{b=8}_{m=2}$  & 1.41M & 6.3 & 74:09 & 65:06 & 114:10 & 73:12 & 368:58 & 612:32 & 218:01 \\
OFTv2$_{b=32}$      & 1.29M & 4.6 & 35:42 & 36:43 & 95:38 & 63:52 & 218:40 & 455:45 & 151:03 \\
\midrule
\rowcolor{gray!10}
RTOFT-B$_{b=32}$ (no restoration) & 1.29M & 4.6 & \textbf{23:08} & 32:22 & 85:58 & 58:18 & 185:41 & 410:56 & 132:44 \\
\rowcolor{gray!10}
RTOFT-B$_{b=32}$ (one-epoch period) & 1.29M & 4.6 & 23:36 & 32:50 & 86:40 & 59:00 & 185:55 & 411:03 & 133:11 \\
\rowcolor{gray!10}
\textbf{RTOFT-B$_{b=32}$ (ours)} & 1.29M & 4.6 & 23:11 & 33:28 & 86:38 & 58:32 & 185:48 & 411:04 & 133:07 \\
\rowcolor{gray!10}
RTOFT-L$_{r=6}$ (no restoration) & 0.99M & 4.5 & \underline{23:10} & 32:28 & 85:16 & 58:04 & 183:58 & 403:27 & 131:04 \\
\rowcolor{gray!10}
RTOFT-L$_{r=6}$              & 0.99M & 4.5 & 23:33 & 32:33 & 85:25 & 58:11 & 184:23 & 403:28 & 131:16 \\
\bottomrule
\end{tabular}
\end{adjustbox}
\end{table}

\section{Per-Task Training Time on GLUE}\label{app:glue_runtime}

Table~\ref{tab:glue-runtime-full} gives the training time of each task in Table~\ref{tab:glue-test}. All runs use the same hardware, number of epochs, and task-level settings. The three block variants of RTOFT are faster than every orthogonal baseline on all six tasks. For example, on CoLA, OFTv2 takes 35:42, BOFT 74:09, and HRA 144:40, compared with 23:36 for RTOFT with the one-epoch period. Restoration adds little to the training time. With the one-epoch period, the six-task average increases from 132:44 to 133:11, or by 0.3\%, and the trigger takes 133:07. This is because a polar factorization is computed only at a restoration, for $d/b$ blocks of size $b$. RTOFT-L takes 131:04 without restoration and 131:16 with it, and only PiSSA (127:10) and LoRA-GA (127:38) are faster. The training time does not follow the number of trainable parameters. PSOFT has the smallest number in the table and takes 169:52.

\begin{table}[t]
\centering
\small
\setlength{\tabcolsep}{4pt}
\renewcommand{\arraystretch}{1.25}
\caption{Comparison of orthogonal finetuning parameterizations for a pretrained
$W\in\mathbb{R}^{d_{\rm out}\times d}$, where $b$ is the block size, $m$ the number of butterfly
components, $r$ the number of reflections or the rank of the low-rank parameterization, $k$ the truncation order, and $M$ the number of steps between restorations.
``Per-step orthogonalization'' is the work that turns the trainable parameters into an
orthogonal matrix in every forward pass, and ``unrolled backward'' indicates that the backward pass differentiates through this work. ``Per-step cost'' is the cost of one forward pass in training and evaluation. Under the drift trigger, $\|S\|_2$ is read at most once per update without gradient.}
\label{tab:param-comparison}
\begin{adjustbox}{max width=\linewidth}
\begin{tabular}{lccccc}
\toprule
 & OFT & BOFT & HRA & OFTv2 & \textbf{RTOFT (ours)} \\
\midrule
Update
  & $W\,\mathrm{diag}(R_i)$
  & $W\prod_{i=1}^{m}\widetilde B_i$
  & $W\prod_{i=1}^{r}(I-2u_iu_i^\top)$
  & $W\,\mathrm{diag}(R_i)$
  & $W(I+S)$ \\
Orthogonal map
  & Cayley
  & Cayley $\times\,m$
  & reflection chain
  & Cayley--Neumann
  & tangent, $R=I+S$ \\
Per-step orthogonalization
  & $(I-S)^{-1}$
  & $m$ inverses
  & $r$ rank-1 updates
  & at most $k$ block matmuls
  & \textbf{none} \\
Unrolled backward
  & \checkmark & \checkmark & \checkmark & \checkmark & \textbf{--} \\
\# Parameters
  & $\tfrac{d(b-1)}{2}$
  & $\tfrac{dm(b-1)}{2}$
  & $rd$
  & $\tfrac{d(b-1)}{2}$
  & $\tfrac{d(b-1)}{2}$ \ or\ $2dr$ \\
Per-step cost
  & $O(d(b^2{+}b{+}d_{\rm out}))$
  & $O(d((b^2{+}b)m{+}d_{\rm out}))$
  & $O(d(r{+}d_{\rm out}))$
  & $O(d(kb^2{+}d_{\rm out}))$
  & $O(d(b{+}d_{\rm out}))$ \\
Drift reading per update
  & N/A & N/A & N/A & N/A
  & $O(db^2)$ \ or\ $O(dr^2)$ \\
Amortized restoration cost
  & N/A & N/A & N/A & N/A
  & \begin{tabular}[c]{@{}c@{}}$O(d(b^2{+}d_{\rm out}b)/M)$\\ or\ $O(d(r^2{+}d_{\rm out}r)/M)$\end{tabular} \\
Exact orthogonality
  & every step
  & every step
  & every step
  & \emph{never} (truncated)
  & \textbf{at every restoration} \\
\bottomrule
\end{tabular}
\end{adjustbox}
\end{table}

\begin{table}[t]
\centering
\scriptsize
\renewcommand{\arraystretch}{1.06}
\setlength{\tabcolsep}{4pt}
\caption{Comparison of OFTv2 and the RTOFT variants with and without restoration on GLUE under the test-split protocol of
Table~\ref{tab:glue-test}. RTOFT-B uses the block parameterization with block size $b$, and
RTOFT-L uses the low-rank parameterization with rank $r$. The restoration column indicates whether the
one-epoch period is used. Subscripts are standard deviations over the five
seeds.}
\label{tab:variants}
\begin{adjustbox}{max width=\linewidth}
\begin{tabular}{lccccc cccccc c}
\toprule
Variant & Restoration & \#Param & Mem.\,(GB) & Time $\downarrow$
 & CoLA & STS-B & RTE & MRPC & SST-2 & QNLI & Avg. \\
\midrule
OFTv2$_{b=32}$ ($k{=}5$) & \ding{55} & 1.29M & 4.6 & 151:03
 & 71.64$_{1.53}$ & 91.60$_{0.46}$ & 88.48$_{2.01}$ & 92.15$_{3.90}$ & 94.90$_{0.75}$ & 94.06$_{1.00}$ & 88.81 \\
OFTv2$_{b=32}$ ($k{=}5$) & \checkmark & 1.29M & 4.6 & N/A
 & 72.04 & 92.12 & 85.45 & 92.31 & 95.28 & 94.36 & 88.59 \\
\midrule
RTOFT-B$_{b=32}$ & \ding{55} & 1.29M & 4.6 & 132:44
 & 71.72$_{1.48}$ & 91.54$_{0.64}$ & 84.83$_{2.05}$ & 90.85$_{1.07}$ & 93.62$_{1.19}$ & 94.03$_{0.41}$ & 87.77 \\
\rowcolor{gray!10}
RTOFT-B$_{b=32}$ (one-epoch period) & \checkmark & 1.29M & 4.6 & 133:11
 & \textbf{72.52}$_{1.08}$ & \textbf{92.30}$_{0.36}$ & \textbf{88.55}$_{1.62}$ & \textbf{92.78}$_{0.84}$ & \textbf{95.88}$_{0.39}$ & \textbf{94.65}$_{0.37}$ & \textbf{89.45} \\
RTOFT-L$_{r=6}$  & \ding{55} & 0.99M & 4.5 & 131:04
 & 70.44$_{1.18}$ & 91.68$_{0.61}$ & 87.05$_{2.44}$ & 90.20$_{1.25}$ & 95.27$_{0.66}$ & 93.99$_{0.45}$ & 88.11 \\
RTOFT-L$_{r=6}$  & \checkmark & 0.99M & 4.5 & 131:16
 & 71.59$_{0.80}$ & 91.71$_{0.58}$ & 83.74$_{2.37}$ & 89.71$_{1.99}$ & 95.18$_{0.33}$ & 94.59$_{0.54}$ & 87.75 \\
\bottomrule
\end{tabular}
\end{adjustbox}
\end{table}

\begin{table}[t]
\centering
\small
\renewcommand{\arraystretch}{1.06}
\setlength{\tabcolsep}{6pt}
\caption{Standard deviation across the five seeds for every method of
Table~\ref{tab:glue-test-full}, in the same column order, with both RTOFT-B schedules of
Table~\ref{tab:glue-test}. Bold marks the smallest deviation among the orthogonal methods and underline the smallest overall where that is a different entry. The last column averages the six tasks.
The drift trigger has the smallest standard deviation in five of the seven columns and a
smaller one than the one-epoch period in every column.}
\label{tab:glue-sd}
\begin{tabular}{lcccccc c}
\toprule
Method & CoLA & STS-B & RTE & MRPC & SST-2 & QNLI & Mean \\
\midrule
Full finetuning     & 3.56 & 0.40 & 2.10 & 1.28 & 0.71 & 0.35 & 1.40 \\
LoRA$_{r=8}$        & 3.60 & 0.57 & 1.85 & 0.51 & 1.62 & 0.62 & 1.46 \\
PiSSA$_{r=8}$       & 2.48 & 1.16 & 1.45 & 2.42 & 1.47 & 0.86 & 1.64 \\
DoRA$_{r=8}$        & 1.15 & 0.94 & 3.26 & 1.75 & 2.31 & 0.37 & 1.63 \\
LoRA-XS$_{r=136}$   & 2.86 & 0.65 & 2.58 & 1.97 & 0.34 & 0.28 & 1.45 \\
GiVA$_{r=8}$        & \underline{0.30} & 0.71 & 1.54 & 1.44 & 0.59 & 0.74 & 0.89 \\
\midrule
PSOFT$_{r=46}$      & 1.98 & 0.39 & 3.98 & 1.40 & 0.70 & 1.24 & 1.62 \\
HRA$_{r=16}$        & 1.69 & 1.01 & \textbf{0.82} & 1.32 & 0.81 & 0.75 & 1.07 \\
BOFT$_{b=8,m=2}$    & 2.09 & 0.72 & 2.82 & 1.17 & 1.56 & 0.41 & 1.46 \\
OFTv2$_{b=32}$      & 1.53 & 0.46 & 2.01 & 3.90 & 0.75 & 1.00 & 1.61 \\
\midrule
\rowcolor{gray!10}
RTOFT-B$_{b=32}$ (one-epoch period) & 1.08 & 0.36 & 1.62 & 0.84 & 0.39 & 0.37 & 0.78 \\
\rowcolor{gray!10}
\textbf{RTOFT-B$_{b=32}$ (ours)} & \textbf{0.84} & \textbf{0.14} & 0.92 & \textbf{0.31} & \textbf{0.16} & \textbf{0.14} & \textbf{0.42} \\
\bottomrule
\end{tabular}
\end{table}

\section{Training Time per Step}\label{app:launch}

The total training time includes evaluation and checkpointing. We therefore also
measure the time of an optimizer step with evaluation and saving disabled.
Table~\ref{tab:step-time} reports it for eleven configurations on one V100, with three repeats on different nodes. A step of RTOFT takes $1$--$11\%$ longer than a step of LoRA, and it is faster than a step of OFTv2 at every truncation order on the two tasks with shorter sequences. At sequence length $256$, the backbone dominates the time, and RTOFT takes the same time as OFTv2 at $k=2$ and less time than OFTv2 at $k=5$ and $k=3$. Reading $\|S\|_2$ after every step, together with the restorations that it triggers, adds at most $1.6\%$ to the time without restoration, whereas restoring after every step adds $4.5$--$27.5\%$ to the time with the trigger.

\begin{table}[t]
\centering
\small
\caption{Per-step training time relative to LoRA on GLUE. The results are measured over $600$ steps in fp32 with evaluation and checkpointing disabled. Every method except BOFT is run back to back on the same V100, and BOFT is run in a separate back-to-back job with its own LoRA on the same node. RTOFT and OFTv2 build their skew-symmetric matrices from the same constant basis (OFTv2 by caching it), so basis construction does not account for the timing difference. Each entry is the mean over three nodes of the ratio to LoRA on the same node, and the column heading gives the sequence length.}
\label{tab:step-time}
\begin{tabular}{lccc}
\toprule
Method & CoLA ($64$) & SST-2 ($128$) & QNLI ($256$) \\
\midrule
LoRA$_{r=8}$ & 1.000 & 1.000 & 1.000 \\
PiSSA$_{r=8}$ & 0.993 & 0.984 & 0.960 \\
DoRA$_{r=8}$ & 1.274 & 1.113 & 1.029 \\
PSOFT$_{r=46}$ & 2.422 & 1.492 & 1.096 \\
BOFT$_{b=8,m=2}$ & 4.788 & 2.724 & 1.588 \\
OFTv2$_{b=32}$, $k{=}5$ & 1.608 & 1.205 & 1.039 \\
OFTv2$_{b=32}$, $k{=}3$ & 1.428 & 1.181 & 1.088 \\
OFTv2$_{b=32}$, $k{=}2$ & 1.186 & 1.118 & 1.011 \\
\midrule
RTOFT-B$_{b=32}$, drift trigger & 1.105 & 1.068 & 1.013 \\
RTOFT-B$_{b=32}$, no restoration & 1.088 & 1.063 & 1.008 \\
RTOFT-B$_{b=32}$, restoration every step & 1.409 & 1.213 & 1.059 \\
\bottomrule
\end{tabular}
\end{table}

\paragraph{Resolution of the timings.}
Running STS-B twice with the same configuration gives the same score and training times that
differ by $0.94\%$. We
therefore do not draw conclusions from differences in training time below one percent.

\paragraph{Construction of the skew-symmetric matrix.}
The skew-symmetric matrix can be built by scattering its entries or by a multiplication with a constant basis. The two give
bitwise identical scores. The constant basis removes about $935$ kernel launches per step at batches of $2{,}048$, $4{,}096$, and
$8{,}192$ tokens, which lowers the training time by $11.5\%$, $8.4\%$, and $3.1\%$, respectively.
In comparison, applying the transform to the weight instead of the activations lowers the device time by $25$ to $34\%$
but changes the training time by at most $0.4\%$. These results suggest that the training time depends mainly on
the number of kernel launches, and a comparison in which only one method uses the constant basis would include a difference that comes from the implementation. In Table~\ref{tab:step-time}, both RTOFT and OFTv2 use the constant basis, and a step of OFTv2 at $k=5$ takes $1.46$, $1.13$, and $1.03$ times as long as a step of RTOFT at sequence lengths $64$, $128$, and $256$.

\paragraph{Whole runs and single steps.}
The whole-run time of OFTv2 in Table~\ref{tab:glue-runtime-full} is $1.18$ times that of RTOFT on SST-2 and $1.11$ times on QNLI, compared with $1.13$ and $1.03$ times per training step (Table~\ref{tab:step-time}).
To examine whether evaluation explains the difference, we rerun OFTv2 as released, OFTv2 with its transform computed once per evaluation phase and reused for every evaluation batch, and RTOFT on all six GLUE tasks (Table~\ref{tab:eval-cache}). Caching changes the whole-run time of OFTv2 by at most $2.0\%$, so evaluation does not explain the difference, and RTOFT remains faster on every task. In this rerun, the whole-run ratios of OFTv2 as released are $1.15$ on SST-2 and $1.04$ on QNLI, which are close to the ratios per step. Since whole-run ratios vary between runs, we use the per-step ratios in Table~\ref{tab:step-time}, which are measured back to back on one node and averaged over three nodes, as the reference for the cost of a step.

\begin{table}[t]
\centering
\small
\caption{Whole-run time of OFTv2 on GLUE, with training and evaluation, relative to that of RTOFT on the same task. OFTv2 is run as released and with its transform computed once per evaluation phase and reused for every evaluation batch. The results are from one run per method and task. The full SST-2 and QNLI runs are placed on separate nodes, and runs of all three methods shortened to one training epoch on one node give ratios within $0.03$ of these.}
\label{tab:eval-cache}
\begin{tabular}{lccccccc}
\toprule
OFTv2 & CoLA & STS-B & RTE & MRPC & SST-2 & QNLI & Mean \\
\midrule
As released & 1.40 & 1.13 & 1.03 & 1.02 & 1.15 & 1.04 & 1.13 \\
Transform cached during evaluation & 1.38 & 1.11 & 1.03 & 1.02 & 1.14 & 1.04 & 1.12 \\
\bottomrule
\end{tabular}
\end{table}

\section{Subspace and Butterfly Parameterizations and Block Size}\label{app:generality}

\begin{figure}[t]
\centering
\begin{minipage}[t]{0.355\linewidth}
\vspace{0pt}
\centering
\begin{tikzpicture}
\begin{semilogxaxis}[
  width=\linewidth, height=3.3cm,
  xlabel={largest drift $\max_j\|S^j\|_2$}, ylabel={CoLA MCC (\%)},
  xmin=0.002, xmax=1.3, ymin=66.5, ymax=70.5,
  grid=major, grid style={gray!20},
  tick label style={font=\tiny}, label style={font=\scriptsize},
  ]
\draw[dotted, gray!60, line width=0.4pt] (axis cs:0.10,66.5) -- (axis cs:0.10,70.5);
\draw[dotted, gray!60, line width=0.4pt] (axis cs:0.20,66.5) -- (axis cs:0.20,70.5);
\addplot[only marks, mark=*, mark size=2.0pt, color=NavyBlue] coordinates {
  (0.008,69.18) (0.108,69.27) (0.140,69.10) (0.208,69.50) (0.301,68.96) (0.433,68.70) };
\addplot[only marks, mark=triangle*, mark size=2.4pt, color=Maroon] coordinates {
  (0.0032,69.62) (0.0745,67.94) (0.1111,67.55) (0.1629,67.44) (0.2570,67.74) (0.3849,68.21) };
\addplot[only marks, mark=square*, mark size=2.2pt, color=ForestGreen] coordinates {
  (0.0117,68.01) (0.300,68.07) (0.328,67.74) (0.472,67.92) (0.647,67.98) (0.897,67.61) };
\addplot[only marks, mark=o, mark size=2.4pt, line width=0.7pt, color=NavyBlue] coordinates {
  (0.0521,67.59) (0.1018,68.90) (0.2006,69.57) };
\addplot[only marks, mark=triangle, mark size=2.7pt, line width=0.7pt, color=Maroon] coordinates {
  (0.1009,69.53) (0.2002,69.53) };
\addplot[only marks, mark=square, mark size=2.5pt, line width=0.7pt, color=ForestGreen] coordinates {
  (0.1049,68.11) (0.2039,68.21) };
\end{semilogxaxis}
\end{tikzpicture}
\\[-2pt]{\scriptsize (a) CoLA, 5 seeds per point}
\end{minipage}\hfill
\begin{minipage}[t]{0.355\linewidth}
\vspace{0pt}
\centering
\begin{tikzpicture}
\begin{semilogxaxis}[
  width=\linewidth, height=3.3cm,
  xlabel={largest drift $\max_j\|S^j\|_2$}, ylabel={RTE accuracy (\%)},
  xmin=0.003, xmax=1.6, ymin=76.5, ymax=84.5,
  grid=major, grid style={gray!20},
  tick label style={font=\tiny}, label style={font=\scriptsize},
]
\draw[dotted, gray!60, line width=0.4pt] (axis cs:0.10,76.5) -- (axis cs:0.10,84.5);
\draw[dotted, gray!60, line width=0.4pt] (axis cs:0.20,76.5) -- (axis cs:0.20,84.5);
\addplot[only marks, mark=*, mark size=2.0pt, color=NavyBlue] coordinates {
  (0.012,81.17) (0.150,81.77) (0.195,82.17) (0.272,82.53) (0.404,81.97) (0.484,81.73) };
\addplot[only marks, mark=triangle*, mark size=2.4pt, color=Maroon] coordinates {
  (0.0053,82.25) (0.0742,81.06) (0.1358,80.82) (0.2077,80.34) (0.2551,81.53) (0.3590,80.82) };
\addplot[only marks, mark=square*, mark size=2.2pt, color=ForestGreen] coordinates {
  (0.0149,78.18) (0.237,78.18) (0.377,77.94) (0.557,77.94) (0.665,77.46) (1.083,77.70) };
\addplot[only marks, mark=o, mark size=2.4pt, line width=0.7pt, color=NavyBlue] coordinates {
  (0.0561,81.77) (0.1041,82.34) (0.2037,83.01) };
\addplot[only marks, mark=triangle, mark size=2.7pt, line width=0.7pt, color=Maroon] coordinates {
  (0.1026,81.77) (0.2009,82.25) };
\addplot[only marks, mark=square, mark size=2.5pt, line width=0.7pt, color=ForestGreen] coordinates {
  (0.1078,77.94) (0.2064,77.22) };
\end{semilogxaxis}
\end{tikzpicture}
\\[-2pt]{\scriptsize (b) RTE, 3 seeds per point}
\end{minipage}\hfill
\begin{minipage}[t]{0.26\linewidth}
\vspace{0pt}
\centering
\begin{tikzpicture}
\begin{axis}[hide axis, width=1.4cm, height=3.3cm, scale only axis,
  xmin=0, xmax=1, ymin=0, ymax=1,
  legend style={draw=none, fill=none, font=\scriptsize, row sep=2pt,
    column sep=3pt, legend columns=2,
    at={(0.5,0.5)}, anchor=center}, legend cell align=left]
\addlegendimage{only marks, mark=*, mark size=2.0pt, color=NavyBlue}
\addlegendentry{block}
\addlegendimage{only marks, mark=o, mark size=2.4pt, line width=0.7pt, color=NavyBlue}
\addlegendentry{block $\tau$}
\addlegendimage{only marks, mark=triangle*, mark size=2.4pt, color=Maroon}
\addlegendentry{butterfly}
\addlegendimage{only marks, mark=triangle, mark size=2.7pt, line width=0.7pt, color=Maroon}
\addlegendentry{butterfly $\tau$}
\addlegendimage{only marks, mark=square*, mark size=2.2pt, color=ForestGreen}
\addlegendentry{subspace}
\addlegendimage{only marks, mark=square, mark size=2.5pt, line width=0.7pt, color=ForestGreen}
\addlegendentry{subspace $\tau$}
\end{axis}
\end{tikzpicture}
\end{minipage}
\caption{Score against the largest drift for three parameterizations at a fixed budget of ten epochs. Filled markers are fixed
periods, where the rightmost one of each parameterization is without restoration, and open markers are drift triggers with $\tau=0.05$ (block only), $0.10$, and $0.20$.
Within each parameterization, the score changes by at most $2.2$ points over the whole range, and each tolerance gives about the same largest drift for all three. The
largest drift $\max_j\|S^j\|_2$ over the steps $j$ of a run is the square root of the largest
orthogonality defect $\|A^\top A-I\|_2$.}
\label{fig:law-full}
\end{figure}
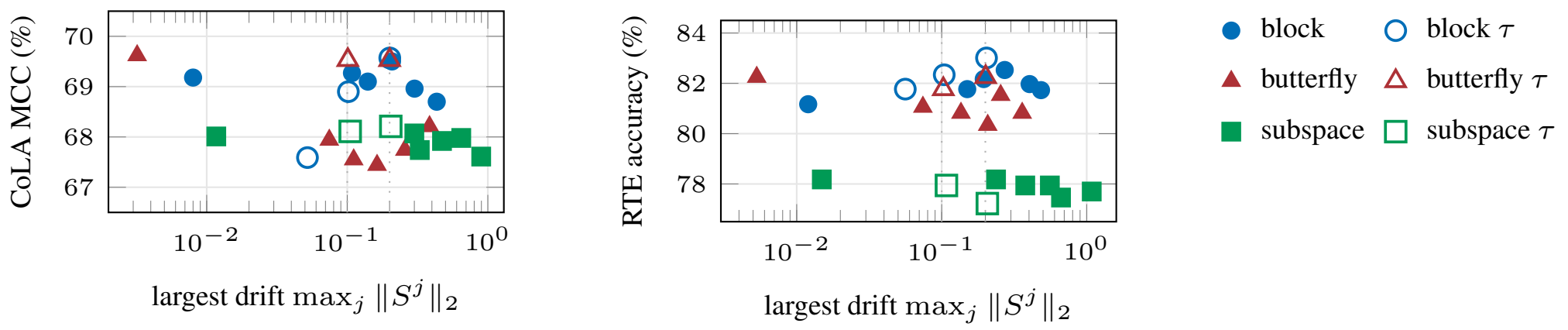

The drift in Figure~\ref{fig:saw} is the same quantity for all
three parameterizations. For the block parameterization, $A=I+S$ and
$\|A^\top A-I\|_2=\|S\|_2^2$ (Lemma~\ref{lem:skew}). For the subspace
parameterization, $A=I+P_r^\top S_rP_r$ (Section~\ref{sec:related}). The linear terms cancel, so $A^\top A-I=-P_r^\top S_r^2P_r$ and
$\|A^\top A-I\|_2=\|S_r\|_2^2$. For the butterfly parameterization, the
factors reduce to one skew-symmetric matrix to first order, and the drift is the spectral norm of this matrix, which is the one used in the forward pass. In each case,
the figure plots the largest drift over the adapted matrices and the steps of a run.

\begin{table}[t]
\centering
\small
\renewcommand{\arraystretch}{1.06}
\setlength{\tabcolsep}{5pt}
\caption{Comparison of PSOFT and RTOFT with the subspace parameterization, in the PSOFT configuration of Table~\ref{tab:glue-test} and over
five seeds. With the one-epoch period, RTOFT matches PSOFT and
trains $1.19\times$ as fast. Without restoration, it loses $1.16$ points and falls
below PSOFT. With the drift trigger at $\tau=0.10$ and the same hyperparameters and seeds, the six-task average is $0.21\pm0.10$ points higher than with the one-epoch period, and the training time stays within $1\%$. The increase is given as the mean and standard error of the paired difference, which is positive on four of five seeds.}
\label{tab:psoft}
\begin{tabular}{lc cccccc c}
\toprule
Method & Time $\downarrow$ & CoLA & STS-B & RTE & MRPC & SST-2 & QNLI & Avg. \\
\midrule
PSOFT$_{r=46}$ & 169:52 & 71.60 & 91.42 & 85.75 & 90.53 & 95.18 & \textbf{93.60} & 88.01 \\
\quad RTOFT, no restoration & \textbf{141:44} & 70.54 & 89.88 & 84.27 & 90.50 & 94.61 & 92.36 & 87.03 \\
\rowcolor{gray!10}
\quad RTOFT, one epoch & 142:20 & 72.05 & 91.43 & 86.02 & \textbf{91.22} & 94.86 & 93.56 & 88.19 \\
\rowcolor{gray!10}
\quad RTOFT, trigger & 142:51 & \textbf{72.60} & \textbf{91.52} & \textbf{86.49} & 91.04 & \textbf{95.23} & 93.51 & \textbf{88.40} \\
\bottomrule
\end{tabular}
\end{table}

\begin{table}[t]
\centering
\small
\renewcommand{\arraystretch}{1.06}
\setlength{\tabcolsep}{3.2pt}
\caption{Comparison of BOFT and RTOFT with the butterfly parameterization over
five seeds, with standard deviations as subscripts. Without restoration, RTOFT
falls below BOFT. With the one-epoch period, it exceeds BOFT and
trains $1.41\times$ as fast. With the drift trigger at $\tau=0.10$ and the same hyperparameters and seeds, the average is $0.28\pm0.12$ points higher than with the one-epoch period, and the training time stays within $1\%$. The increase is given as the mean and standard error of the paired difference, which is positive on four of five seeds.}
\label{tab:tboft}
\begin{tabular}{lc cccccc c}
\toprule
Method & Time $\downarrow$ & CoLA & STS-B & RTE & MRPC & SST-2 & QNLI & Avg. \\
\midrule
BOFT$_{b=8,m=2}$ & 218:01 & 68.85 & 91.85 & 86.78 & 89.91 & 95.33 & \textbf{93.85} & 87.76 \\
\quad RTOFT, no restoration & \textbf{153:01} & 68.35$_{2.58}$ & 90.91$_{1.25}$ & 84.41$_{3.14}$ & 90.72$_{1.76}$ & 94.88$_{1.30}$ & 93.16$_{2.28}$ & 87.07 \\
\rowcolor{gray!10}
\quad RTOFT, one epoch & 154:10 & 70.23$_{1.49}$ & 92.25$_{0.43}$ & 87.13$_{2.63}$ & \textbf{91.27}$_{1.42}$ & 95.21$_{0.96}$ & 93.74$_{1.29}$ & 88.31 \\
\rowcolor{gray!10}
\quad RTOFT, trigger & 155:12 & \textbf{70.89}$_{1.21}$ & \textbf{92.33}$_{0.35}$ & \textbf{87.49}$_{2.18}$ & 91.24$_{1.14}$ & \textbf{95.78}$_{0.83}$ & 93.80$_{1.01}$ & \textbf{88.59} \\
\bottomrule
\end{tabular}
\end{table}

BOFT is a product of $m$ orthogonal factors with permutations between them. To first order, $(I+S_1)(I+S_2)=I+S_1+S_2$, and a sum of skew-symmetric matrices is
again skew-symmetric, so the tangent update reduces the product to one factor. Restoration recovers a product over the cycles, because each restoration multiplies
$\operatorname{polar}(I+S)$ into $W$ and the next cycle starts from the rotated weight. For the subspace parameterization, $R=I+P_r^\top(R_r-I)P_r$
with a fixed $P_r$. Replacing $R_r$ by $I_r+S_r$ gives the forward
pass $W(I+P_r^\top S_rP_r)$, and since $P_rP_r^\top=I_r$, a restoration needs only the polar factor of an
$r\times r$ matrix.

\paragraph{Block size.}
Table~\ref{tab:blocksize} summarizes the results for different values of the block size $b$. For accuracy, we train RTOFT on CoLA, STS-B, and RTE with the drift trigger at $\tau=0.10$ and tune the learning rate of each $b$ on the grid in Appendix~\ref{app:impl}. For cost, we measure the time of a step of RTOFT and of OFTv2 at each $b$ on CoLA, SST-2, and QNLI under the protocol of Table~\ref{tab:step-time}. At the same number of trainable parameters, a step of OFTv2 at its default $k=5$ takes $1.31$ to $1.47$ times as long as a step of RTOFT on CoLA, $1.07$ to $1.13$ times on SST-2, and $1.03$ to $1.05$ times on QNLI. The drift grows faster with larger blocks, and doubling $b$ multiplies the number of restorations by $1.7$ to $3.3$. Since the trigger adjusts this number, the same tolerance keeps the defect bound $\tau^2$ at every $b$. The three-task average increases by $0.67$ points from $b=16$ to $b=32$, and further increasing $b$ to $64$, with twice the number of parameters, leads to an improvement of $0.10$ points. We set $b=32$ in Table~\ref{tab:glue-test}, where RTOFT and OFTv2 have the same block size. We also consider a model with $b=32$ in some layers and $b=64$ in the others. In a separate batch of five seeds on the same three tasks, this mixed setting with $1.95$M adapter parameters scores $0.70$ points above $b=64$ everywhere ($2.61$M) and $0.44$ points above $b=32$ everywhere. Compared with $b=32$ everywhere, $b=64$ everywhere is $0.26$ points lower in this batch and $0.10$ points higher in Table~\ref{tab:blocksize}, so a larger block in every layer gives no consistent gain.

\begin{table}[t]
\centering
\small
\caption{Block size $b$ for RTOFT with the drift trigger at $\tau=0.10$. Scores are means over three seeds (five for $b=32$, from Table~\ref{tab:glue-test}), with the learning rate of each $b$ tuned on the same grid. Restoration counts are ranges over three seeds at every $b$, including $b=32$. Step-time ratios use the protocol of Table~\ref{tab:step-time} and at $b=32$ agree with it to within $0.04$. Trainable parameters exclude the classification head, as in Table~\ref{tab:glue-test}.}
\label{tab:blocksize}
\begin{tabular}{lcccc}
\toprule
 & $b=16$ & $b=32$ & $b=64$ & $b=128$ \\
\midrule
Trainable parameters & 0.62M & 1.29M & 2.61M & 5.27M \\
\midrule
\multicolumn{5}{l}{\textit{OFTv2 step time $\div$ RTOFT step time}} \\
\quad CoLA & 1.37 & 1.42 & 1.47 & 1.31 \\
\quad SST-2 & 1.12 & 1.13 & 1.07 & 1.09 \\
\quad QNLI & 1.05 & 1.04 & 1.03 & 1.04 \\
\midrule
\multicolumn{5}{l}{\textit{Restorations per run}} \\
\quad CoLA & 6 & 11--13 & 39--40 & 110--116 \\
\quad STS-B & 17--18 & 47--52 & 132--142 & 244--329 \\
\quad RTE & 5--7 & 10 & 25--26 & 63--65 \\
\midrule
\multicolumn{5}{l}{\textit{Score}} \\
\quad CoLA & 73.07 & 73.55 & 74.07 & 73.64 \\
\quad STS-B & 92.32 & 92.44 & 92.65 & 90.68 \\
\quad RTE & 87.77 & 89.20 & 88.75 & 89.75 \\
Three-task average & 84.39 & 85.06 & 85.16 & 84.69 \\
\bottomrule
\end{tabular}
\end{table}

\section{Computational Cost and Reachable Rotations}\label{app:cost}

\paragraph{Block parameterization.}
Let $b$ divide $d$ and
$S={\rm blockdiag}(S_1,\dots,S_{d/b})$ with $S_i\in{\rm Skew}(b)$. This parameterization has
$d(b-1)/2$ trainable parameters, the same number as block OFT and
OFTv2 at block size $b$. A forward pass of the tangent update costs
$O\!\left(d(b+d_{\rm out})\right)$, of which $O(db)$ is for the product $Sx$ and
$O(dd_{\rm out})$ is for $W$, and no $O(db^2)$ construction is needed. In comparison, OFTv2 at truncation order $k$ performs at most $k$ multiplications of block matrices in every forward pass, which cost $O(dkb^2)$, and differentiates through them in the backward pass. The drift trigger adds at most one reading of $\|S\|_2=\max_i\|S_i\|_2$ per optimizer step. The spectral norm of a $b\times b$ skew-symmetric block costs $O(b^3)$ by an eigendecomposition or an SVD, so a reading costs at most $O(db^2)$. It needs no gradient and is not repeated for evaluation batches. A restoration is separable over the blocks,
${\rm polar}(I+S)={\rm blockdiag}({\rm polar}(I+S_1),\dots,{\rm polar}(I+S_{d/b}))$.
The $d/b$ polar factors cost $O(db^2)$ in total, and multiplying them into
$W\in\mathbb{R}^{d_{\rm out}\times d}$ costs $O(dd_{\rm out}b)$. With $M$ steps between restorations, restoration therefore adds
$O\!\left(d(b^2+d_{\rm out}b)/M\right)$ per step. With one restoration per epoch, this
is $0.3\%$ of the total training time on GLUE (Table~\ref{tab:glue-runtime-full}).

\paragraph{Reading the drift and restoring at the size of a 70B model.}
Reading the drift and restoring need no gradient, do not depend on the sequence length or the batch size, and run layer by layer on the device that holds each layer. Table~\ref{tab:70b} reports their time at the layer sizes of Llama-3.1-70B on one H200 with $b=32$. The shares are relative to the forward and backward time of the adapted layers alone and are thus upper bounds on the shares of a full training step. A restoration takes $0.82$ s, which is $0.12\%$ of the step time of the adapted layers if it happens every $50$ steps. The share is about $0.4\%$ if it happens every $15$ steps, as on segmentation-to-image. In a screened reading, we first bound the spectral norm of each block by the norms of its rows and then compute the exact norm only for blocks whose bound reaches $\tau$. Under data parallelism, the replicas read identical values and the trigger needs no communication. Under tensor parallelism with shards aligned to the blocks, it needs one scalar max-reduction per step.

\begin{table}[t]
\centering
\small
\caption{Drift reading and restoration over all $80$ layers at Llama-3.1-70B layer sizes on one H200, $b=32$. Shares are of the step time of the adapted layers alone.}
\label{tab:70b}
\begin{tabular}{lcc}
\toprule
Operation & Time & Share \\
\midrule
Drift reading, screened (step without restoration) & 25 ms & $\leq0.18\%$ \\
Drift reading, every block exact (worst case) & 313 ms & $\leq2.3\%$ \\
One restoration of every adapted matrix & 0.82 s & $0.12\%$ if every $50$ steps \\
\bottomrule
\end{tabular}
\end{table}

\paragraph{Reachable rotations.}
The rotation angles of a polar factor ${\rm polar}(I+S)$ are $\arctan\theta_j$ and hence smaller than $\pi/2$. The following proposition shows that a product of three such factors can still be any rotation.

\begin{proposition}[Rotations reached by restorations]
\label{prop:reach}
Let $\mathcal{G}=\{\operatorname{polar}(I+S):S\in\mathcal{S}\}$, where $\mathcal{S}$ is the set of block-diagonal skew-symmetric
matrices defined above, and let $\mathcal{B}$ be the corresponding
group of block-diagonal rotations. Then $\mathcal{G}$ is the set of elements
of $\mathcal{B}$ whose rotation angles all lie in $(-\pi/2,\pi/2)$, and
every element of $\mathcal{B}$ is a product of three elements of $\mathcal{G}$.
\end{proposition}

\begin{proof}
For a skew-symmetric $S$ with eigenvalues $\pm i\theta_j$, $\operatorname{polar}(I+S)$ has
eigenvalues $e^{\pm i\arctan\theta_j}$, and $\arctan$ maps $\mathbb{R}$ onto
$(-\pi/2,\pi/2)$. Conversely, if every angle $\varphi_j$ of $R\in\mathcal{B}$ lies in this interval,
let $S$ act on the eigenplanes of $R$ with $\theta_j=\tan\varphi_j$. Each eigenplane lies inside
a block, so $S$ is block-diagonal and skew-symmetric, and ${\rm polar}(I+S)=R$ by Lemma~\ref{lem:skew}(d).
This shows the first claim. For the second, any $R\in\mathcal{B}$ lies in a
maximal torus, in which it is given by angles $\varphi_j\in(-\pi,\pi]$. The element
with angles $\varphi_j/3$ lies in $\mathcal{G}$ since
$|\varphi_j/3|<\pi/2$, and its cube is $R$ because the three factors commute.
\end{proof}

\section{RTOFT Variants}\label{app:variants}

\begin{table}[t]
\centering
\scriptsize
\renewcommand{\arraystretch}{1.06}
\setlength{\tabcolsep}{2.7pt}
\caption{Table~\ref{tab:glue-test} with the additional methods LoRA-XS and GiVA, under the test-split protocol of \citet{wu2025efficient}. Every method is re-run in the same pipeline. Cells are means over the same five seeds, with standard deviations in Table~\ref{tab:glue-sd}, and \#Param and Adapter are as defined in Table~\ref{tab:glue-test}. Time is the mean wall-clock time per task on one V100 in \emph{mm:ss}, and Rel.\ Time is normalized to LoRA. The protocol is given in Appendix~\ref{app:impl}. Bold and underlined values indicate the best and second-best results, respectively, and tied second-best results are unmarked.}
\label{tab:glue-test-full}
\begin{adjustbox}{max width=\linewidth}
\begin{tabular}{lcccc c cccccc c}
\toprule
& \multicolumn{5}{c}{\textbf{Efficiency}}
& \multicolumn{7}{c}{\textbf{GLUE performance}} \\
\cmidrule(lr){2-6}
\cmidrule(lr){7-13}
Method
& \#Param
& Adapter
& Mem.\,(GB)
& Time $\downarrow$
& Rel.\ Time $\downarrow$
& CoLA
& STS-B
& RTE
& MRPC
& SST-2
& QNLI
& Avg. \\
\midrule

Full finetuning
& 184M
& 184M
& 5.9
& 149:03
& 1.08$\times$
& 68.20
& 91.34
& 83.35
& 90.48
& 95.18
& 92.94
& 86.92 \\

\midrule
\multicolumn{13}{l}{\emph{Low-rank adaptation}} \\

LoRA$_{r=8}$
& 1.33M
& 1.33M
& 4.5
& 137:54
& 1.00$\times$
& 71.98
& 91.42
& 84.53
& 90.63
& 95.01
& 93.92
& 87.91 \\

PiSSA$_{r=8}$
& 1.33M
& 1.33M
& 4.5
& \textbf{127:10}
& \textbf{0.92$\times$}
& 67.52
& 91.43
& 86.17
& 90.25
& 94.69
& 92.90
& 87.16 \\

DoRA$_{r=8}$
& 1.41M
& 1.41M
& 5.8
& 138:28
& 1.00$\times$
& 67.99
& 91.92
& 87.68
& 89.24
& 95.25
& 93.99
& 87.68 \\

LoRA-XS$_{r=136}$
& 1.33M
& 23.89M
& \underline{4.2}
& 133:20
& 0.97$\times$
& 64.67
& 91.22
& 85.00
& 87.59
& 93.69
& 94.10
& 86.05 \\

GiVA$_{r=8}$
& \underline{0.09M}
& 1.41M
& 4.7
& 145:02
& 1.05$\times$
& 72.18
& 91.51
& 85.36
& 90.58
& 95.17
& 93.81
& 88.10 \\

\midrule
\multicolumn{13}{l}{\emph{Orthogonal finetuning}} \\

PSOFT$_{r=46}$
& \textbf{0.08M}
& 7.71M
& \textbf{4.1}
& 169:52
& 1.23$\times$
& 71.60
& 91.42
& 85.75
& 90.53
& 95.18
& 93.60
& 88.01 \\

HRA$_{r=16}$
& 1.33M
& 1.33M
& 15.2
& 283:35
& 2.06$\times$
& 70.05
& 90.47
& 86.46
& \textbf{92.82}
& 94.71
& 93.15
& 87.94 \\

BOFT$_{b=8,m=2}$
& 1.41M
& 1.41M
& 6.3
& 218:01
& 1.58$\times$
& 68.85
& 91.85
& 86.78
& 89.91
& 95.33
& 93.85
& 87.76 \\

OFTv2$_{b=32}$ ($k{=}5$)
& 1.29M
& \textbf{1.29M}
& 4.6
& 151:03
& 1.10$\times$
& 71.64
& 91.60
& 88.48
& 92.15
& 94.90
& 94.06
& 88.81 \\

\midrule
\rowcolor{gray!10}
RTOFT-B$_{b=32}$ (one-epoch period)
& 1.29M
& \textbf{1.29M}
& 4.6
& 133:11
& 0.97$\times$
& \underline{72.52}
& \underline{92.30}
& \underline{88.55}
& \underline{92.78}
& \underline{95.88}
& \underline{94.65}
& \underline{89.45} \\

\rowcolor{gray!10}
\textbf{RTOFT-B$_{b=32}$ (ours)}
& 1.29M
& \textbf{1.29M}
& 4.6
& \underline{133:07}
& 0.97$\times$
& \textbf{73.55}
& \textbf{92.44}
& \textbf{89.20}
& 92.65
& \textbf{96.75}
& \textbf{94.80}
& \textbf{89.90} \\

\bottomrule
\end{tabular}
\end{adjustbox}
\end{table}

\begin{table}[t]
\vspace{-10pt}
\centering
\fontsize{7}{8.4}\selectfont
\renewcommand{\arraystretch}{1.05}
\setlength{\tabcolsep}{2.2pt}
\begin{tabular}{@{}l r ccc c@{}}
\toprule
Method & \#P.\,(M) & \multicolumn{3}{c}{S2I $\uparrow$} & L2F $\downarrow$ \\
\cmidrule(lr){3-5}\cmidrule(l){6-6}
 & & mIoU & mAcc & aAcc & Error \\
\midrule
DreamBooth & 859.52 & 7.72 & 14.40 & 33.61 & 146.19 \\
ControlNet & 361.30 & 20.88 & 30.91 & 61.42 & 7.61 \\
T2I-Adapter & 77.00 & 16.38 & 26.31 & 51.63 & 23.75 \\
LoRA$_{r=16}$ & 2.50 & 24.58 & 37.74 & 60.31 & 8.01 \\
\midrule
OFT$_{4\,\mathrm{blocks}}$ & 26.40 & 27.18 & 40.23 & 62.58 & 7.03 \\
COFT$_{4\,\mathrm{blocks}}$ & 26.40 & 26.81 & 39.95 & 62.82 & 6.97 \\
OFTv2$_{c=2}$ & 26.34 & 27.79 & 40.78 & 64.36 & 6.61 \\
BOFT$^{m=4}_{bn=16}$ & 26.40 & 29.01 & 41.43 & 67.94 & 5.60 \\
HRA$_{r=8,\lambda=10^{-3}}$ & 0.89 & 29.19 & 40.80 & 67.73 & 5.51 \\
\midrule
RTOFT-B$_{2\,\mathrm{blocks}}$ (no restoration) & 26.35 & 26.32 & 40.91 & 64.90 & 6.86 \\
\rowcolor{gray!6}
RTOFT-B$_{2\,\mathrm{blocks}}$ (one-epoch period) & 26.35 & \underline{29.45} & \underline{42.26} & \underline{67.97} & \underline{5.36} \\
\rowcolor{gray!12}
\textbf{RTOFT-B$_{2\,\mathrm{blocks}}$ (ours)} & 26.35 & \textbf{30.18} & \textbf{42.78} & \textbf{69.19} & \textbf{5.31} \\
\bottomrule
\end{tabular}
\caption{Comparison of all methods on controllable generation with Stable Diffusion (S2I: segmentation-to-image on ADE20K; L2F: landmark-to-face on CelebA-HQ). RTOFT-B (ours) uses the drift trigger at $\tau=0.10$, and the one-epoch row differs only in its restoration schedule.}
\label{tab:diffusion-full}
\vspace{-10pt}
\end{table}

\paragraph{Low-rank parameterization.}

RTOFT-L uses the low-rank parameterization
\begin{equation}
    S
    =
    UV^\top-VU^\top,
    \qquad
    U,V\in\mathbb{R}^{d\times r},
    \label{eq:lowrank-S}
\end{equation}
which is skew-symmetric by construction and has $2dr$ trainable
parameters. We initialize $U$ with Kaiming initialization and set $V=0$, so
that the initial adapter is exactly the identity while both factors remain
trainable.

The matrix $S$ need not be formed, because its product with a vector is
\[
    Sx
    =
    U(V^\top x)-V(U^\top x),
\]
which costs $O(dr)$. The total cost of a forward pass is then $O\!\left(d(r+d_{\rm out})\right)$.

A polar decomposition of $I+S$ costs $O(d^3)$ in general, and the low-rank
structure reduces this cost. Let
\[
    Y=[\,U\mid V\,]\in\mathbb{R}^{d\times2r},
    \qquad
    Y=Q_YR_Y
\]
be a thin QR factorization. Since both the row and column spaces of $S$ lie in
${\rm col}(Y)$,
\[
    S
    =
    Q_YM_SQ_Y^\top,
    \qquad
    M_S
    =
    Q_Y^\top SQ_Y
    \in{\rm Skew}(2r).
\]
Any orthogonal completion $[\,Q_Y\;Q_\perp\,]$ therefore block-diagonalizes
$I+S$ into
\[
    {\rm blockdiag}
    (I_{2r}+M_S,I_{d-2r}).
\]
Hence
\begin{equation}
    {\rm polar}(I+S)
    =
    I
    +
    Q_Y
    \bigl(\Phi_s-I_{2r}\bigr)
    Q_Y^\top,
    \qquad
    \Phi_s
    \coloneqq
    {\rm polar}(I_{2r}+M_S).
    \label{eq:lr-polar}
\end{equation}

The product with $W$ can be computed without forming a
$d\times d$ matrix, as
$W\,{\rm polar}(I+S)=W+(WQ_Y)(\Phi_s-I_{2r})Q_Y^\top$.
A restoration therefore costs $O(dr^2)$ for the QR and projections,
$O(r^3)$ for the polar factor of size $2r$, and
$O(dd_{\rm out}r)$ for the product with $W$, or
$O(d(r^2+d_{\rm out}r)/M)$ per step with $M$ steps between restorations.

\paragraph{Restoration of the low-rank parameterization.}

The block parameterization has no trainable state other than $S$, whereas the low-rank parameterization has
two factors. At a restoration, RTOFT-L rotates $U$ together with the weight and resets only $V$:
\begin{equation}
    U
    \leftarrow
    \Phi^\top U,
    \qquad
    V
    \leftarrow0,
    \qquad
    \Phi={\rm polar}(I+S).
    \label{eq:lr-warm}
\end{equation}
The tangent map is then the identity again, and at $V=0$ the gradient with respect to $V$ is
$\nabla_V\mathcal{L}|_{V=0}=-2\,{\rm skew}(W^\top\nabla\mathcal{L}(W))\,U$,
so the retained $U$ determines the directions of the next steps. The following proposition gives the gradient right after a restoration.

\begin{proposition}[Gradient after a reset]
\label{prop:reset-signal}
Let $g={\rm skew}\bigl(W^\top\nabla\mathcal{L}(W)\bigr)$ be the Riemannian gradient of the loss
at the current weight, and consider the tangent update $W(I+S)$ right after a restoration.
\begin{enumerate}\itemsep1pt
\item[(i)] Suppose that $S=S(\xi)$ is linear in its parameters $\xi$ with
Jacobian $J$, as for the block, subspace, and butterfly parameterizations. At the reset
$\xi=0$, $\nabla_\xi\mathcal{L}=J^\top g$, which is zero if and only if $g$ is orthogonal to the range of $J$, that is, to the support.
\item[(ii)] Suppose that $S=UV^\top-VU^\top$. Then
$\nabla_U\mathcal{L}=2\,gV$ and $\nabla_V\mathcal{L}=-2\,gU$. At the reset $V=0$, $\nabla_U\mathcal{L}=0$, and a gradient step changes $V$ only, in
the direction $gU$, provided that $gU\neq0$. If $gU=0$, or if both factors are reset to zero, both gradients
are zero.
\end{enumerate}
\end{proposition}

\begin{proof}
For a skew-symmetric $E$, $\tfrac{d}{dt}\mathcal{L}\bigl(W(I+tE)\bigr)\big|_{t=0}
=\langle\nabla\mathcal{L}(W),WE\rangle=\langle W^\top\nabla\mathcal{L}(W),E\rangle=\langle g,E\rangle$, where the last equality holds because
the symmetric part of $W^\top\nabla\mathcal{L}(W)$ is orthogonal to $E$. In (i), the
differential of $S$ is $J\,d\xi$, so $\nabla_\xi\mathcal{L}=J^\top g$, which is zero
if and only if $g\perp{\rm range}(J)$. In (ii), the differential is
$dS=dU\,V^\top+U\,dV^\top-dV\,U^\top-V\,dU^\top$. Since $g^\top=-g$,
$\langle g,dU\,V^\top-V\,dU^\top\rangle=2\langle gV,dU\rangle$ and
$\langle g,U\,dV^\top-dV\,U^\top\rangle=-2\langle gU,dV\rangle$, which give the two gradients.
The remaining claims follow by setting $V=0$ or $U=V=0$.
\end{proof}

\paragraph{Storage of the adapter.}
Instead of multiplying the polar factors into the weight, RTOFT can keep
the pretrained weight and accumulate the factors in the adapter by
$R\leftarrow R\,{\rm polar}(I+S)$ and $S\leftarrow0$. The final weight is then $W^{\rm pre}R$
after the last restoration. For the block parameterization, every factor is block-diagonal with the same
blocks, and so is $R$, which has $db$ entries. By Lemma~\ref{lem:skew}(c) applied to each block, as in the proof of Theorem~\ref{thm:main}, every block $R_i$ of $R$ lies in $\mathcal{SO}(b)$. Since the matrix exponential maps ${\rm Skew}(b)$ onto $\mathcal{SO}(b)$, $R_i=\exp(Z_i)$ with $Z_i\in{\rm Skew}(b)$. Hence $R$ can be stored as the $d(b-1)/2$ free entries of the $Z_i$, which is the number reported as Adapter for the block rows of Tables~\ref{tab:glue-test} and~\ref{tab:commonsense}. Computing the $Z_i$ after the last restoration and rebuilding $R$ before it is multiplied into the weight each require $d/b$ matrix functions of size $b\times b$ and cost $O(db^2)$. For the subspace parameterization, every factor
has the form $I+P_r^\top(R_r-I_r)P_r$ with the same fixed $P_r$ and $R_r\in\mathcal{O}(r)$, and so does $R$. In
both cases, the adapter can be stored separately from the pretrained weight and merged once, like a low-rank adapter. For the
butterfly parameterization, the polar factor need not have the butterfly form, so it is multiplied into the weight.

\subsection{Relation to Low-Rank Adaptation}
\label{sec:lora-connection}

Substituting \eqref{eq:lowrank-S} into the tangent update gives
\begin{align}
    W(I+UV^\top-VU^\top)
    &=
    W+(WU)V^\top-(WV)U^\top.
    \label{eq:lora-form}
\end{align}
RTOFT-L is thus an additive update of rank at most $2r$ whose left factors $WU$ and $WV$ lie in the column space of $W$. It has $2dr$ trainable parameters, whereas LoRA of rank $r$ has
$(d+d_{\rm out})r$. Since $I+S$ is invertible for every skew-symmetric $S$,
\[
    {\rm col}\!\left(W(I+S)\right)
    =
    {\rm col}(W),
\]
so the update stays in the column space of the pretrained weight.

\subsection{Comparison of the Variants}

Table~\ref{tab:variants} compares the block and the low-rank variants with and without restoration. For RTOFT-B, restoration increases the six-task average from $87.77$ to $89.45$ and improves every task, with the largest gains on RTE and SST-2. It also reduces the standard deviation over seeds on every task, from $1.14$ to $0.78$ on average. For RTOFT-L, the five tasks other than RTE gain $1.20$ points in total, and RTE decreases from $87.05$ to
$83.74$. The average changes from $88.11$ to $87.75$. The two variants differ in the gradient after a reset (Proposition~\ref{prop:reset-signal}). In RTOFT-B, the gradient after $S\leftarrow0$ is the block-diagonal part of ${\rm skew}(W^\top\nabla\mathcal{L})$, so $S$ moves in the next step. In RTOFT-L, $\nabla_U\mathcal{L}=2\,{\rm skew}(W^\top\nabla\mathcal{L})\,V$ is proportional to $V$, which is reset to zero at every restoration. The training time is about the same with and without restoration for both variants.

\begin{table}[t]
\centering
\scriptsize
\renewcommand{\arraystretch}{1.04}
\setlength{\tabcolsep}{5pt}
\caption{Comparison of finetuning methods on PEFT-Arena \citep{huang2026peftarena}
(mathematics domain, Qwen2.5-7B, supervised finetuning). Figure~\ref{fig:hero}(c) shows a
part of this table. The best value in each column is in bold, and the strongest OFT setting reported by
PEFT-Arena is underlined. The OFT rows keep the label of PEFT-Arena,
whose implementation of this baseline is the OFTv2 construction. The three RTOFT-B rows use the one-epoch period, the drift
trigger at $\tau=0.10$ with $\|S\|_2$ read after every step, and no restoration, and each is the median of three
runs.}
\label{tab:arena-full}
\begin{tabular}{llcrr}
\toprule
Method & Setting & Tr.\ Param & Target (\%) $\uparrow$ & General (\%) $\rightarrow 0$ \\
\midrule
Base            & N/A       & 7.61B   & 35.30 \fmt{+0.00}  & 46.97 \fmt{+0.00} \\
Full finetuning & N/A       & 7.61B   & \textbf{50.63} \fmt{+15.33} & 34.22 \fmt{-12.75} \\
\midrule
OFT   & block 16  & 8.49M   & 42.33 \fmt{+7.03}  & 42.58 \fmt{-4.39} \\
OFT   & block 32  & 17.55M  & 46.93 \fmt{+11.63} & \underline{44.37} \fmt{-2.60} \\
OFT   & block 64  & 35.68M  & 46.23 \fmt{+10.93} & 35.97 \fmt{-11.00} \\
OFT   & block 128 & 71.92M  & \underline{47.77} \fmt{+12.47} & 36.98 \fmt{-9.99} \\
\midrule
LoRA     & r4a8    & 10.09M & 42.33 \fmt{+7.03}  & 41.66 \fmt{-5.31} \\
LoRA     & r8a16   & 20.19M & 42.47 \fmt{+7.17}  & 39.22 \fmt{-7.75} \\
LoRA     & r16a32  & 40.37M & 44.87 \fmt{+9.57}  & 34.91 \fmt{-12.06} \\
LoRA     & r32a64  & 80.74M & 45.37 \fmt{+10.07} & 38.21 \fmt{-8.76} \\
AdaLoRA  & r8a16   & 30.28M & 40.43 \fmt{+5.13}  & 35.41 \fmt{-11.56} \\
PiSSA    & r8a16   & 20.19M & 44.53 \fmt{+9.23}  & 24.78 \fmt{-22.19} \\
MiLoRA   & r8a16   & 20.19M & 42.60 \fmt{+7.30}  & 37.62 \fmt{-9.35} \\
KeepLoRA & r8      & 20.19M & 40.53 \fmt{+5.23}  & 43.75 \fmt{-3.22} \\
MiSS     & r8      & 11.12M & 43.17 \fmt{+7.87}  & 39.12 \fmt{-7.85} \\
MiSS     & r64     & 89.00M & 46.93 \fmt{+11.63} & 32.77 \fmt{-14.20} \\
VeRA     & r256    & 1.44M  & 39.43 \fmt{+4.13}  & \textbf{47.25} \fmt{+0.28} \\
DoRA     & r8a16   & 21.58M & 42.33 \fmt{+7.03}  & 40.25 \fmt{-6.72} \\
IA$^3$   & N/A     & 1.82M  & 35.13 \fmt{-0.17}  & 44.71 \fmt{-2.26} \\
\midrule
\rowcolor{gray!10}
RTOFT-B (one-epoch period) & block 32 & 17.55M & 47.21 \fmt{+11.91} & 44.44 \fmt{-2.53} \\
\rowcolor{gray!10}
\textbf{RTOFT-B (ours)} & block 32 & 17.55M & 47.19 \fmt{+11.89} & 44.62 \fmt{-2.35} \\
RTOFT-B (no restoration) & block 32 & 17.55M & 43.83 \fmt{+8.53} & 33.62 \fmt{-13.35} \\
\bottomrule
\end{tabular}
\end{table}

\subsection{Variation over Runs on PEFT-Arena}

On PEFT-Arena, the OFTv2 baseline (OFT at block size $32$) and RTOFT-B with the period and with the trigger, all with $17.55$M parameters, are each run three
times. On the target, the mean and standard deviation over the three runs are $47.18\pm0.09$ for the trigger, $47.12\pm0.16$ for the period, and $46.79\pm0.50$ for OFTv2. On general
capability, they are $44.62\pm0.16$, $44.45\pm0.13$, and
$44.33\pm0.07$, respectively. The mean of the trigger is the highest on both. On the target, the
ranges of the three overlap, and on general capability, the lowest run of the trigger is above the highest run of OFTv2. The trigger restores $38$ to $41$ times per run, and the largest drift in a forward pass is $0.0999$. The readings and the restorations take $33$ to $37$ seconds per run, compared with about $12.2$ hours of training, or about $0.08\%$ of the training time. RTOFT-B without restoration is also run three times, with $43.83\pm0.35$ on the target and
$33.62\pm0.42$ on general capability. Its highest run on general capability is more than ten points below
the lowest run of the period. Table~\ref{tab:arena-full} reports the median of the three runs for these rows, and
every other row is a single run under the protocol of the benchmark.

\begin{table}[t]
\centering
\scriptsize
\renewcommand{\arraystretch}{1.06}
\setlength{\tabcolsep}{6pt}
\caption{Training time for controllable generation in \emph{h:mm:ss} under the setting of
Table~\ref{tab:diffusion}, also given relative to LoRA. All methods are
trained in one pipeline on the same hardware with the same number of steps, and the peak memory is $53$GB for each method.}
\label{tab:diffusion-runtime}
\begin{tabular}{lll cc cc}
\toprule
Method & Config & \#Param\,(M) & S2I time $\downarrow$ & $\times$LoRA & L2F time $\downarrow$ & $\times$LoRA \\
\midrule
LoRA & $r=16$ & 2.50 & \textbf{21:09:37} & \textbf{1.000} & \textbf{15:34:26} & \textbf{1.000} \\
OFT & 4 blocks & 26.40 & 23:28:10 & 1.109 & 17:57:13 & 1.153 \\
COFT & 4 blocks & 26.40 & 23:31:53 & 1.112 & 18:00:33 & 1.156 \\
OFTv2 & $c=2$ & 26.34 & 22:58:47 & 1.086 & 17:54:42 & 1.150 \\
BOFT & $m=4$, $bn=16$ & 26.40 & 30:29:12 & 1.441 & 24:29:31 & 1.573 \\
HRA & $r=8$, $\lambda=10^{-3}$ & 0.89 & 32:21:08 & 1.529 & 26:20:44 & 1.692 \\
\midrule
\rowcolor{gray!10}
RTOFT-B (one-epoch period) & 2 blocks & 26.35 & 21:26:06 & 1.013 & 15:51:08 & 1.018 \\
\rowcolor{gray!10}
RTOFT-B (one-epoch period) & 4 blocks & 13.15 & \underline{21:11:35} & \underline{1.002} & 15:50:56 & 1.018 \\
RTOFT-B (no restoration) & 2 blocks & 26.35 & 21:25:58 & 1.013 & \underline{15:50:52} & \underline{1.018} \\
\rowcolor{gray!10}
\textbf{RTOFT-B (ours)} & 2 blocks & 26.35 & 21:35:36 & 1.020 & 15:58:10 & 1.025 \\
\bottomrule
\end{tabular}
\end{table}

\begin{table}[t]
\centering
\small
\renewcommand{\arraystretch}{1.15}
\setlength{\tabcolsep}{5pt}
\caption{Hyperparameters for RTOFT on DeBERTaV3-base under the validation-score protocol of
Table~\ref{tab:glue-val}, at block size $b=16$. The setup follows \citet{liu2024parameter} and
\citet{yuan2024hra}.}
\label{tab:hp-glue-val}
\begin{adjustbox}{max width=\linewidth}
\begin{tabular}{lcccccccc}
\toprule
 & MNLI & SST-2 & CoLA & QQP & QNLI & RTE & MRPC & STS-B \\
\midrule
Batch size       & 32 & 32 & 32 & 32 & 32 & 32 & 32 & 32 \\
Epochs           & 8 & 10 & 34 & 12 & 12 & 34 & 60 & 11 \\
Learning rate    & 2E-04 & 2E-04 & 9E-04 & 1E-03 & 8E-05 & 1E-03 & 1E-03 & 1.5E-03 \\
Warm-up steps    & 1000 & 500 & 100 & 1000 & 500 & 50 & 50 & 50 \\
Dropout          & 0 & 0 & 0 & 0 & 0 & 0 & 0 & 0 \\
Max.\ seq.\ len. & 256 & 128 & 64 & 320 & 512 & 320 & 320 & 128 \\
\bottomrule
\end{tabular}
\end{adjustbox}
\end{table}

\begin{table}[t]
\centering
\small
\renewcommand{\arraystretch}{1.15}
\setlength{\tabcolsep}{7pt}
\caption{Learning rates for RTOFT on DeBERTaV3-base under the test-split protocol of
Table~\ref{tab:glue-test}, at block size $b=32$. The split and the remaining settings follow
\citet{wu2025efficient}, and the
learning-rate grid is given in Appendix~\ref{app:impl}. The rates are selected with the one-epoch period, and the drift trigger uses the same rates, so the two differ only in the restoration schedule.}
\label{tab:hp-glue-test}
\begin{tabular}{lcccccc}
\toprule
 & CoLA & STS-B & MRPC & RTE & SST-2 & QNLI \\
\midrule
Learning rate & 4E-04 & 1E-03 & 8E-04 & 5E-04 & 4E-04 & 1E-04 \\
Epochs & 20 & 20 & 30 & 30 & 10 & 5 \\
Max.\ seq.\ len. & 64 & 128 & 256 & 256 & 128 & 256 \\
Batch size & 32 & 32 & 32 & 32 & 32 & 32 \\
\bottomrule
\end{tabular}
\end{table}

\begin{table}[t]
\centering
\small
\renewcommand{\arraystretch}{1.15}
\setlength{\tabcolsep}{8pt}
\caption{Configuration for the commonsense reasoning experiments of
Table~\ref{tab:commonsense}. Following SVFT
\citep{lingam2024svft}, we use the Commonsense-15K protocol, which is based on the eight-task LLM-Adapters benchmark \citep{hu2023llmadapters}.
Every compared method uses this configuration and the same adapter placement.}
\label{tab:hp-commonsense}
\begin{tabular}{ll}
\toprule
Setting & Value \\
\midrule
Backbone            & Qwen3.5-9B (base) \\
Training data       & \texttt{commonsense\_15k} (15K examples) \\
Target modules      & \texttt{q\_proj}, \texttt{k\_proj}, \texttt{v\_proj},
                      \texttt{in\_proj\_qkv}, \texttt{up\_proj}, \texttt{down\_proj} \\
Batch size / micro-batch & 64 / 2 \\
Epochs              & 3 \\
Warm-up steps       & 100 \\
Weight decay        & 0 \\
Cutoff length       & 512 \\
Validation set size & 120 \\
Evaluation / checkpoint interval & every 233 steps \\
Train on inputs     & yes \\
\bottomrule
\end{tabular}
\end{table}

\section{Additional Ablation Studies}\label{app:ablation-detail}

In this section, we provide details on the ablation studies in Section~\ref{sec:ablation} and on the measurements in Section~\ref{sec:audit}.

\paragraph{Learning-rate schedule, scale, and number of epochs.}
We vary the learning-rate schedule, its scale, and the number of epochs in six settings on CoLA (Appendix~\ref{app:impl}). Over eighteen runs under the
one-epoch period, the largest drift ranges from $0.12$ to $0.32$. Under the
trigger with $\tau=0.20$, it stays between $0.2000$ and $0.2022$, and the number of restorations ranges from
one to ten per run.

\paragraph{Projection of checkpoints trained without restoration.}
We evaluate runs of twenty epochs with and without restoration at five points of training. The drift without
restoration reaches $0.48$--$0.66$, whereas it decreases to $0.005$ under the one-epoch period.
Projecting the checkpoints trained without restoration onto the orthogonal group at any of the five points
does not improve the score, which changes by $-1.36$ on average over fifteen evaluations on CoLA.
On PEFT-Arena, the same projection of the final model changes the general capability from $-13.35$ to $-13.11$.

\paragraph{Learning rate.}
Table~\ref{tab:lr-stress} reports runs of ten epochs on SST-2 under the test-split protocol at $b=32$, with the learning rate set to a multiple of the tuned rate in Table~\ref{tab:hp-glue-test} and with $\tau=0.10$ for the trigger. These runs are separate from those of Table~\ref{tab:glue-test}. OFTv2, the one-epoch period, and the trigger use five seeds, and restoring after every step and the period with the same number of restorations as the trigger use three. The one-epoch period and the trigger are above OFTv2 at every rate, and OFTv2 decreases from $95.08$ at $0.5\times$ to $93.58$ at $2\times$. At $2\times$, the trigger reaches $95.42$ and restoring after every step reaches $95.53$, whereas the one-epoch period reaches $93.73$. The best seed of the one-epoch period is below the worst seed of the trigger. A fixed period with the same number of restorations as the trigger reaches
$94.44$. Relative to the tuned rate, the score decreases by less than one point when the model is restored after every step or at the trigger, and by $2.45$ points under the one-epoch period. This suggests that the larger loss comes from the drift between restorations, and that the timing of the restorations matters more than their number.

\par\noindent\begin{minipage}{\linewidth}
\centering
\small
\captionof{table}{Accuracy on SST-2 at learning rates from half to twice the tuned rate. Learning rates are given as multiples of the tuned SST-2 rate of Table~\ref{tab:hp-glue-test}. The results of OFTv2, the one-epoch period, and the drift trigger ($\tau=0.10$) are averaged over five seeds, and those of per-step restoration and the period with the same number of restorations as the trigger are averaged over three seeds.}
\label{tab:lr-stress}
\begin{tabular}{lccccc}
\toprule
Schedule & $0.5\times$ & $0.75\times$ & $1\times$ & $1.5\times$ & $2\times$ \\
\midrule
OFTv2 & 95.08 & 94.88 & 94.80 & 94.27 & 93.58 \\
One-epoch period & 95.26 & 95.99 & 96.18 & 95.42 & 93.73 \\
Drift trigger ($\tau=0.10$) & 95.33 & 95.93 & 96.23 & 95.44 & 95.42 \\
Per-step restoration & -- & -- & 96.20 & -- & 95.53 \\
Period, trigger's number of restorations & -- & -- & -- & -- & 94.44 \\
\bottomrule
\end{tabular}
\end{minipage}
\vspace{0.6em}

\paragraph{Tolerance.}
We rerun the six GLUE tasks at $\tau=0.20$ and at $\tau=0.10$ over five paired seeds at the rates of Table~\ref{tab:hp-glue-test} (Table~\ref{tab:tau020}). Five of the six tasks change by less than one standard error, and SST-2 increases by $0.37\pm0.26$ at $\tau=0.20$. The six-task average increases by $0.38$, and the trigger restores three to five times less often. On commonsense reasoning with Qwen3.5-9B over five paired seeds, the eight-task average changes by $-0.01$ ($95\%$ CI $[-0.07,0.06]$) at $\tau=0.05$ and by $0.00$ ($[-0.05,0.04]$) at $\tau=0.20$ relative to $\tau=0.10$. 

\par\noindent\begin{minipage}{\linewidth}
\centering
\small
\captionof{table}{Comparison of $\tau=0.20$ and $\tau=0.10$ on GLUE in a separate batch of runs. The rows give the paired difference over five seeds (mean $\pm$ standard error) and the range of the number of restorations over the seeds.}
\label{tab:tau020}
\begin{adjustbox}{max width=\linewidth}
\begin{tabular}{lcccccc}
\toprule
 & CoLA & STS-B & MRPC & RTE & SST-2 & QNLI \\
\midrule
$\tau{=}0.20$ minus $\tau{=}0.10$ & $+1.27_{\pm1.54}$ & $0.00_{\pm0.08}$ & $-0.39_{\pm1.08}$ & $+1.01_{\pm1.17}$ & $+0.37_{\pm0.26}$ & $-0.01_{\pm0.02}$ \\
Restorations at $\tau{=}0.10$ & 11--13 & 47--53 & 28--31 & 10--11 & 45--49 & 6 \\
Restorations at $\tau{=}0.20$ & 3--4 & 10--12 & 6--7 & 3--4 & 10--11 & 2 \\
\bottomrule
\end{tabular}
\end{adjustbox}
\end{minipage}
\vspace{0.6em}

\paragraph{Penalty instead of restoration.}
Table~\ref{tab:l2} summarizes the results when restoration is replaced by an $L_2$ penalty of weight $\lambda$ on $S$. With a small weight, the drift at the end of training is $0.17$--$0.20$, above $\tau$. A weight large enough to keep the drift below $\tau$ on CoLA reduces the score by $13.9$ points on CoLA and by $6.3$ points on SST-2. Under the trigger, the drift at the end of training is $0.035$--$0.077$, and the score on CoLA changes by less than $0.1$ after the final restoration.

\par\noindent\begin{minipage}{\linewidth}
\centering
\small
\captionof{table}{Comparison of an $L_2$ penalty on $S$ and the drift trigger at the tuned rate, in the same batch of runs as Table~\ref{tab:tau020}. The columns give the largest drift at the end of training on CoLA and the paired difference, penalty minus trigger, scored before the final restoration. The paired differences are means $\pm$ standard errors over five seeds on CoLA and three on SST-2.}
\label{tab:l2}
\begin{tabular}{lccc}
\toprule
 & Final drift (CoLA) & CoLA & SST-2 \\
\midrule
Drift trigger ($\tau{=}0.10$) & $0.035$--$0.077$ & -- & -- \\
$L_2$, $\lambda{=}0.01$ & $0.17$--$0.20$ & $-0.35_{\pm1.54}$ & $-1.53_{\pm0.67}$ \\
$L_2$, $\lambda{=}0.1$ & $0.11$--$0.13$ & $-7.55_{\pm1.84}$ & $-2.98_{\pm0.40}$ \\
$L_2$, $\lambda{=}1$ & $0.03$ & $-13.93_{\pm1.18}$ & $-6.34_{\pm0.86}$ \\
\bottomrule
\end{tabular}
\end{minipage}
\vspace{0.6em}

\paragraph{Trigger for each matrix.}
Since the defect bound holds for each matrix (Corollary~\ref{cor:tau}), the trigger need not use the maximum over all adapted matrices. We further consider a trigger under which the $\|S\|_2$ of each adapted matrix is read after every step and the matrix is restored on its own when its $\|S\|_2$ reaches $\tau$. Compared with the trigger of Table~\ref{tab:glue-test} on the same five seeds and on CoLA, STS-B, and RTE, this trigger needs $60\%$ fewer restorations of single matrices ($10{,}721$ versus $26{,}712$, summed over tasks and seeds). The training time changes by at most $0.2\%$. The mean paired differences, separate minus joint trigger, are $-1.46$ on CoLA ($t=-1.3$), $-0.12$ on STS-B ($t=-2.4$), $+0.57$ on RTE ($t=0.5$), and $-0.34$ on the three-task average ($t=-0.6$), and none is significant at the $5\%$ level. The main experiments use one trigger for all matrices, which is simpler to implement.

\paragraph{Block defects.}
Table~\ref{tab:audit-full} gives the minimum, median, and maximum block defect at each
truncation order for the checkpoints of Table~\ref{tab:audit}. At the
default $k{=}5$, the block defects range from $1.3\times10^{-7}$ to $1.36\times10^{-2}$, which is
five orders of magnitude.

\begin{center}
\captionsetup{type=table}
\small
\setlength{\tabcolsep}{3pt}
\begin{adjustbox}{max width=\linewidth}
\begin{tabular}{l ccc ccc ccc}
\toprule
& \multicolumn{3}{c}{CoLA} & \multicolumn{3}{c}{RTE} & \multicolumn{3}{c}{STS-B} \\
\cmidrule(lr){2-4} \cmidrule(lr){5-7} \cmidrule(lr){8-10}
$k$ & min & median & max & min & median & max & min & median & max \\
\midrule
2 & $7.7\times10^{-4}$ & $5.4\times10^{-3}$ & $3.8\times10^{-2}$ & $3.9\times10^{-3}$ & $2.7\times10^{-2}$ & $1.9\times10^{-1}$ & $1.6\times10^{-2}$ & $6.4\times10^{-2}$ & $2.6\times10^{-1}$ \\
3 & $2.3\times10^{-11}$ & $7.9\times10^{-9}$ & $2.7\times10^{-6}$ & $2.7\times10^{-9}$ & $9.3\times10^{-7}$ & $3.2\times10^{-4}$ & $1.8\times10^{-7}$ & $1.2\times10^{-5}$ & $8.1\times10^{-4}$ \\
4 & $2.3\times10^{-11}$ & $7.9\times10^{-9}$ & $2.7\times10^{-6}$ & $2.7\times10^{-9}$ & $9.3\times10^{-7}$ & $3.2\times10^{-4}$ & $1.8\times10^{-7}$ & $1.2\times10^{-5}$ & $8.1\times10^{-4}$ \\
$\mathbf{5}$ & $1.3\times10^{-7}$ & $6.2\times10^{-6}$ & $3.0\times10^{-4}$ & $3.1\times10^{-6}$ & $1.5\times10^{-4}$ & $7.3\times10^{-3}$ & $4.8\times10^{-5}$ & $8.1\times10^{-4}$ & $1.36\times10^{-2}$ \\
6 & $3.5\times10^{-18}$ & $5.6\times10^{-14}$ & $8.9\times10^{-10}$ & $9.5\times10^{-15}$ & $1.5\times10^{-10}$ & $2.4\times10^{-6}$ & $8.5\times10^{-12}$ & $9.8\times10^{-9}$ & $1.1\times10^{-5}$ \\
\bottomrule
\end{tabular}
\end{adjustbox}
\captionof{table}{Minimum, median, and maximum block orthogonality defect at each truncation order over all adapted blocks of checkpoints from the reference implementation.}
\label{tab:audit-full}
\end{center}

\paragraph{Measured orders of the defect.}
The following table gives $\mu$ in Lemma~\ref{lem:cnp}, the measured order of the defect in $\|Q\|_2$, and the number of block multiplications at each truncation order. Consistent with Lemma~\ref{lem:cnp}, the measured order is $2$ at $k=2$, $2\mu$ when $\mu$ is odd, and $\mu$ at $k=5$.

\begin{center}
\small
\begin{tabular}{lccccc}
\toprule
$k$ & $2$ & $3$ & $4$ & $5$ \emph{(default)} & $6$ \\
\midrule
$\mu$ & N/A & $3$ & $3$ & $4$ & $5$ \\
measured order of $\|\widetilde R_k^\top\widetilde R_k-I\|_2$ & $2.00$ & $6.00$ & $6.00$ & $4.00$ & $10.00$ \\
block multiplications & $0$ & $2$ & $2$ & $3$ & $4$ \\
\bottomrule
\end{tabular}
\end{center}

The measured orders are the slopes of the defect against $\|Q\|_2$ on a logarithmic scale. The exact
Cayley branch of the same implementation has a defect of $1.1\times10^{-15}$, which is machine
precision.\footnote{The exact branch of the implementation computes $(I-Q)(I+Q)^{-1}$, which is the transpose of the limit
$(I+Q)(I-Q)^{-1}$ of the series, so a checkpoint trained with one of the two has its rotations inverted when it is evaluated with the other.}

\begin{figure}[h]
\centering
\begin{tikzpicture}
\begin{axis}[
  width=0.70\linewidth, height=5.0cm, scale only axis,
  xlabel={epoch}, ylabel={largest block defect},
  xmin=0, xmax=40, ymode=log, ymin=2e-6, ymax=1.1e-1,
  xtick={1,9,16,25,32,39},
  ytick={1e-5,1e-4,1e-3,1e-2},
  yticklabels={$10^{-5}$,$10^{-4}$,$10^{-3}$,$10^{-2}$},
  grid=major, grid style={gray!18},
  tick label style={font=\tiny}, label style={font=\scriptsize}, clip=false,
  legend style={font=\tiny, at={(0.02,0.98)}, anchor=north west, draw=none,
    fill=none, row sep=0pt},
]
\addplot[draw=none, fill=BrickRed!7, forget plot] coordinates
  {(9,1e-2)(40,1e-2)(40,1.1e-1)(9,1.1e-1)} \closedcycle;
\draw[dashed, BrickRed, thick] (axis cs:0,1e-2)--(axis cs:40,1e-2);
\node[anchor=north west, font=\tiny, color=BrickRed] at (axis cs:21.5,8.8e-3)
  {$\tau^{2}=10^{-2}$};
\addplot[mark=*, mark size=1.0pt, color=BrickRed, line width=0.7pt] coordinates
{(1,3.223e-04)(2,9.069e-04)(3,1.911e-03)(4,3.367e-03)(5,4.014e-03)(6,5.395e-03)
 (7,7.401e-03)(8,9.789e-03)(9,1.113e-02)(10,1.099e-02)(11,1.159e-02)(12,1.327e-02)
 (13,1.370e-02)(14,1.344e-02)(15,1.301e-02)(16,1.335e-02)(17,1.570e-02)(18,1.529e-02)
 (19,1.617e-02)(20,1.673e-02)(21,1.763e-02)(22,1.878e-02)(23,1.955e-02)(24,1.939e-02)
 (25,1.983e-02)(26,2.153e-02)(27,2.299e-02)(28,2.325e-02)(29,2.425e-02)(30,2.412e-02)
 (31,2.359e-02)(32,2.347e-02)(33,2.343e-02)(34,2.355e-02)(35,2.375e-02)(36,2.413e-02)
 (37,2.422e-02)(38,2.405e-02)(39,2.416e-02)};
\addlegendentry{$k=5$, the default}
\addplot[mark=square*, mark size=1.0pt, color=NavyBlue, line width=0.7pt] coordinates
{(1,3.599e-06)(2,1.531e-05)(3,4.955e-05)(4,1.126e-04)(5,1.506e-04)(6,2.414e-04)
 (7,3.645e-04)(8,5.524e-04)(9,6.754e-04)(10,6.788e-04)(11,7.407e-04)(12,9.435e-04)
 (13,9.575e-04)(14,9.496e-04)(15,9.125e-04)(16,9.366e-04)(17,1.081e-03)(18,1.146e-03)
 (19,1.233e-03)(20,1.270e-03)(21,1.379e-03)(22,1.500e-03)(23,1.580e-03)(24,1.584e-03)
 (25,1.629e-03)(26,1.846e-03)(27,2.022e-03)(28,2.058e-03)(29,2.173e-03)(30,2.158e-03)
 (31,2.100e-03)(32,2.080e-03)(33,2.070e-03)(34,2.090e-03)(35,2.104e-03)(36,2.150e-03)
 (37,2.168e-03)(38,2.153e-03)(39,2.165e-03)};
\addlegendentry{$k=3$, one block multiplication cheaper}
\addplot[only marks, mark=o, mark size=2.6pt, line width=0.7pt, color=BrickRed]
  coordinates {(9,1.113e-02)};
\node[anchor=north west, font=\tiny, color=BrickRed, align=left]
  at (axis cs:9.7,8.6e-03) {crosses at $23\%$ of training};
\node[anchor=west, font=\tiny, color=BrickRed] at (axis cs:39.4,2.416e-02)
  {$2.4\,\tau^{2}$};
\node[anchor=west, font=\tiny, color=NavyBlue] at (axis cs:39.4,2.165e-03)
  {$0.22\,\tau^{2}$};
\draw[gray!65, line width=0.5pt] (axis cs:16,2e-6)--(axis cs:16,1.335e-02);
\end{axis}
\end{tikzpicture}
\caption{Largest block defect
$\|\widetilde R_k^\top\widetilde R_k-I\|_2$ at every epoch of an OFTv2 run on
STS-B at the default order $k=5$ and at $k=3$, which needs one block
multiplication fewer. The dashed line is the defect bound $\tau^{2}=10^{-2}$ of the drift trigger at $\tau=0.10$.
Both curves use the same thirty-nine checkpoints
and differ only in the truncation order. The defect at $k=5$ exceeds the bound from epoch $9$ on, and the defect at $k=3$ stays below it at all
checkpoints. The vertical line marks the checkpoint that the run
returns, which is the one in Table~\ref{tab:audit}.}
\label{fig:oftv2-traj}
\end{figure}
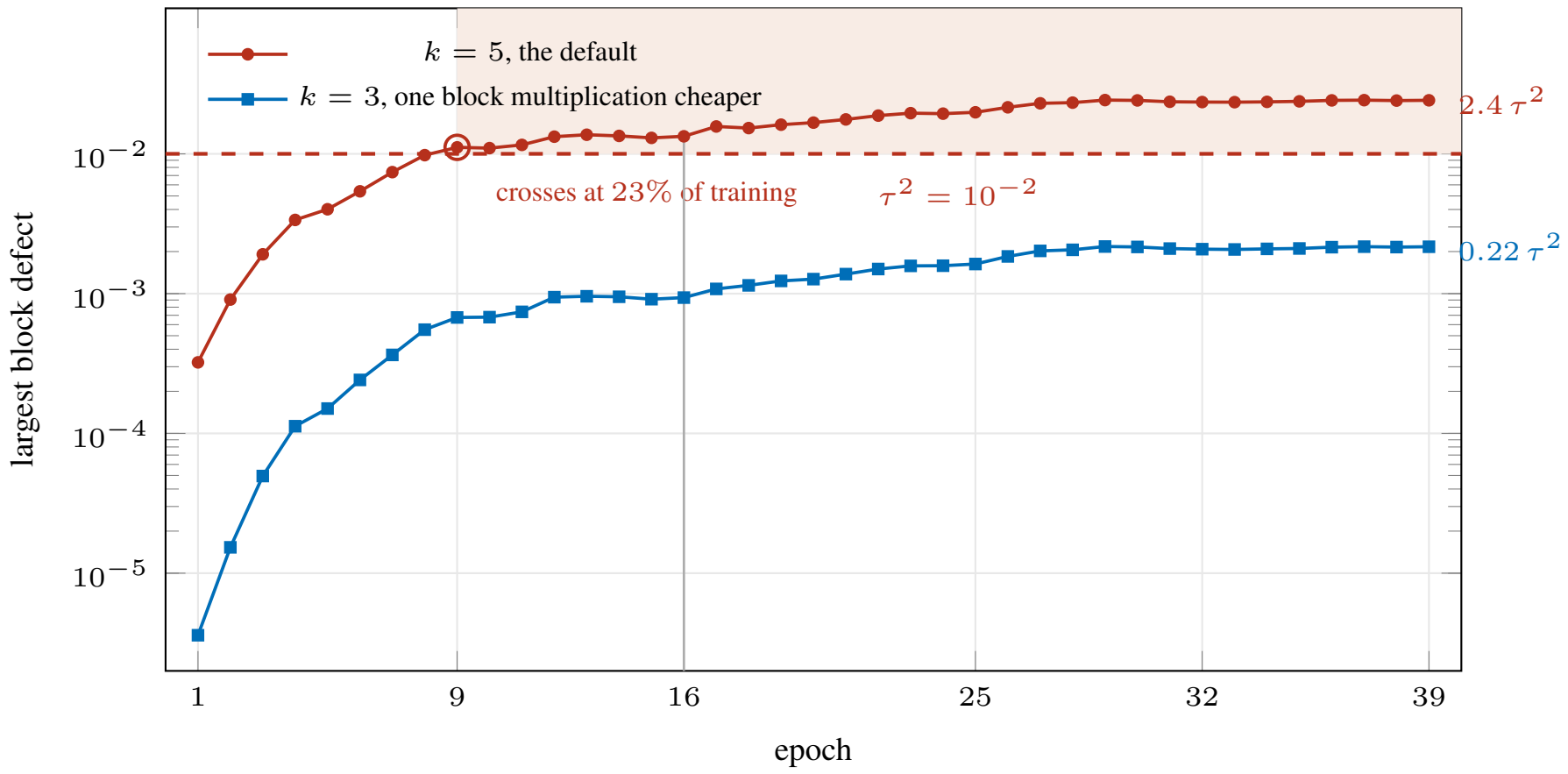

By Lemma~\ref{lem:cnp}, the ratio of the defects at $k=5$ and $k=3$ is to leading order proportional to
$\|Q\|_2^{-2}$, so it changes with $\|Q\|_2$ during training. Over the
thirty-nine checkpoints of Figure~\ref{fig:oftv2-traj}, the ratio decreases from $89.6$ to
$11.2$, which corresponds to a growth of $\|Q\|_2$ by a factor of $2.83$. Consistent with Lemma~\ref{lem:cnp}, the growth of the two defects corresponds to factors of $2.94$
and $2.91$, and the three values agree to within $4\%$.

\paragraph{Finetuning with $k=3$.}
We train OFTv2 at $k=3$ under
the same protocol, learning-rate grid, and five seeds as at the default $k=5$ on
the six GLUE tasks of Table~\ref{tab:glue-test}. The paired difference and the training time in Section~\ref{sec:audit} are those of these runs.

\paragraph{Defect during training.}
Figure~\ref{fig:oftv2-traj} shows the largest block defect at every
epoch of one OFTv2 run on STS-B. At the default order, it increases on average, exceeds $\tau^{2}=10^{-2}$ after $23\%$
of training, and stays above this value.

\paragraph{Selection of the schedule.}
Following the practice of tuning schedules on shorter runs \citep{khaled2025muonbp},
we compare six fixed periods and three tolerances on CoLA and RTE over ten
epochs (Figure~\ref{fig:period}). The learning rates are those of the main experiments, and the seeds are the same for every setting. We use the one-standard-error rule, fixed before
the runs, which selects the cheapest among the settings within one standard error
of the best score of a task. On CoLA, the six periods range from $2680$
restorations to none, and their largest drifts differ by a factor of fifty. Nevertheless, their scores lie
between $68.70$ and $69.50$, within one standard error of each other. Restoring
after every step, with a largest drift of $0.008$, does not score higher than restoring once an
epoch, with a largest drift of $0.208$. On RTE, the run without restoration scores $81.73$, below the threshold of $82.69$ of the rule.

The rule selects the trigger with $\tau=0.20$. On CoLA, it admits every setting
except $\tau=0.05$, and on RTE, it admits only $\tau=0.20$. This tolerance has the highest score on both tasks, $69.57$ and $83.01$, with $3.8$ and $3.0$ restorations on average, respectively, compared with ten for the one-epoch period.
By Theorem~\ref{thm:main}, the largest
defect read after an update at $\tau=0.20$ is $0.040$ on CoLA and $0.041$ on RTE, compared with $0.19$ and $0.23$ without
restoration. The largest drifts of the one-epoch period on the two tasks, $0.208$ and $0.272$, differ by thirty percent, whereas those at
$\tau=0.20$, $0.2006$ and $0.2037$, differ by $1.5\%$.

The main experiments use the trigger with $\tau=0.10$ and report the one-epoch period for comparison.
Under the one-standard-error rule, a trigger is selected over a fixed period.
The tolerance of the main experiments is set from the stretch budget by Corollary~\ref{cor:budget}. Its defect bound is $10^{-2}$, compared with $0.04$ at $\tau=0.20$, and it needs more restorations. At the twenty
epochs of the main experiments, restoration with the one-epoch period improves CoLA by $0.80$ points, from $71.72$ to $72.52$ (Table~\ref{tab:variants}).

\paragraph{Drift within a run.}
Under the one-epoch period, the drift is not constant within a run. On CoLA, it is $0.21$ in
the first epoch and $0.011$ in the tenth, and RTE
shows the same pattern. A fixed period therefore restores too rarely when the drift is large and continues to
restore when it is small. The drift follows the
decaying learning rate, so a period that suits one learning-rate schedule need not suit another. With a constant learning
rate, the largest drift under the one-epoch period increases from $0.208$ to $0.372$ on CoLA and from $0.272$ to
$0.461$ on RTE, whereas that under the trigger does not change.

Under the trigger, the interval between restorations adapts within a run. On CoLA
under the validation-score protocol with $9112$ steps, the interval at $\tau=0.15$ increases from $140$ to $1320$ steps as the learning rate decays.
The same is observed at the other tolerances, where the interval increases by factors of $3.1$ and $7.4$. A larger
tolerance gives longer intervals, and the mean
interval increases from $425$ steps at $\tau=0.15$ to $4140$ steps at $\tau=0.40$.

\paragraph{Drift across models.}
With a period of one epoch and $b=32$, the
largest $\|S\|_2$ in a run is $0.170$ on commonsense reasoning,
$0.208$ on CoLA, $0.272$ on RTE, and $0.799$ on PEFT-Arena. The largest defects therefore range from
$0.029$ to $0.638$, or by a factor of twenty-two. On segmentation-to-image with two blocks per matrix, the largest $\|S\|_2$ is $1.65$. At $b=32$, the drift is the largest on PEFT-Arena, which also loses the most without restoration, $10.82$ points of general
capability. The drift is the smallest on commonsense reasoning, which loses $0.39$ points of accuracy. At a drift of $0.799$, $I+S$ stretches some direction by $28\%$ (Lemma~\ref{lem:skew}), and restoration still gives the largest gain in this setting.

With the trigger,
Corollary~\ref{cor:tau} bounds the defect of every forward pass by $\tau^2$. Tolerances of $0.05$, $0.10$, and $0.20$ give largest readings of
$0.0521$, $0.1018$, and $0.2006$ on CoLA (Figure~\ref{fig:period}). They
exceed $\tau$ by $4.2\%$, $1.8\%$, and $0.3\%$ of $\tau$,
because $\|S\|_2$ is read after each step and can exceed $\tau$ by the growth of one step. The scores of the three tolerances are $67.59\pm1.95$, $68.90\pm2.37$, and $69.57\pm2.10$,
compared with $68.70$ to $69.50$ for the six periods, and every
pairwise difference is within the variation over seeds. From Qwen3.5-9B on commonsense reasoning to Qwen2.5-7B on PEFT-Arena, the largest drift under the trigger changes only from $0.1002$ to $0.102$, so the defect changes by a factor of $1.04$, compared with twenty-two under the period. Under a period, as at a truncation order, the defect is known only after the run, whereas the trigger fixes its bound before training.

\paragraph{Cost and variation over seeds.}
The trigger needs one reading of $\|S\|_2$ per step and more restorations in the settings with a fast drift. Its training time over the six GLUE tasks nevertheless matches that of the period, because a restoration needs one polar factorization for $d/b$ blocks of size $b$, and the readings and restorations add at most $1.6\%$ to the time of a step (Appendix~\ref{app:launch}). On the GLUE test split, the trigger is better than the period on five
of the six tasks, and under the validation-score protocol, it is better on all
eight tasks, by $0.09$ on QQP to $1.50$ on RTE. The standard deviation over seeds under the trigger is smaller than that under the period on every task of the test split, $0.42$ compared with $0.78$ on average (Table~\ref{tab:glue-sd}). Under the validation-score protocol, it is no larger on any task, $0.52$ compared with $0.58$ on average.

\paragraph{Implementations behind Table~\ref{tab:order-restart}.}
The $k{=}2$ results in Table~\ref{tab:order-restart} use our implementation, and the $k{=}5$ results use the reference
implementation with our restoration added, so each row uses one implementation. Running $k{=}2$ in the reference implementation changes its score by $0.18$ points, a
fifth of the $0.86$-point difference between the rows. Restoration improves $k{=}5$ on
five of the six tasks, by $0.16$ to $0.52$ points, and reduces RTE by $3.03$ points to $85.45$. Without RTE, restoration improves the average by $0.35$ at $k{=}5$ and by $1.27$ at
$k{=}2$.

\section{Additional Qualitative Results}\label{app:qualitative}

\begin{figure}[t]
\centering
\begin{subfigure}{0.492\linewidth}
\centering
\includegraphics[width=\linewidth]{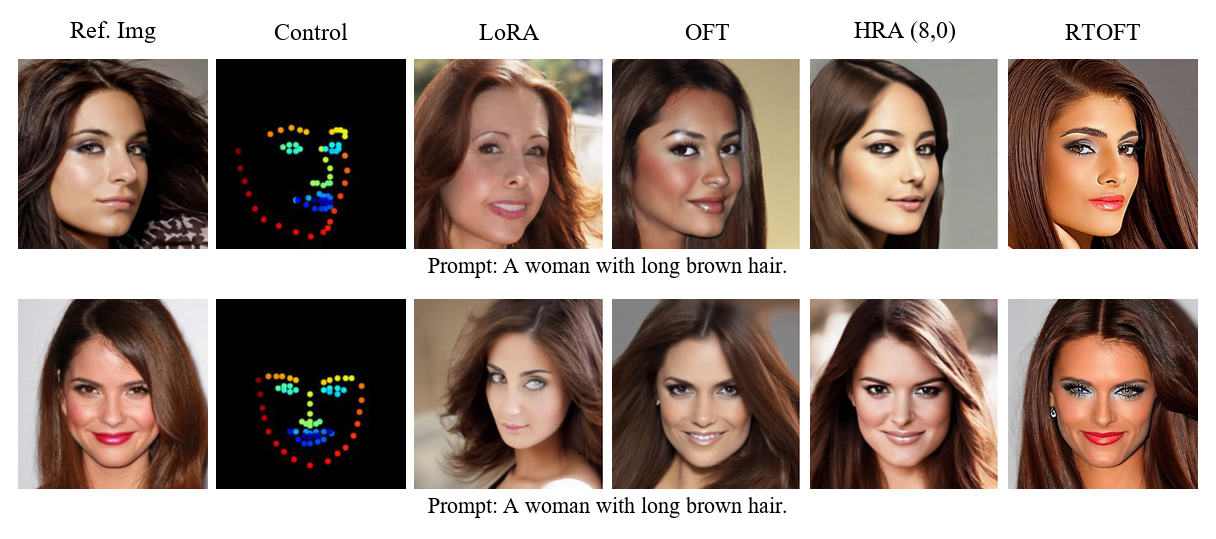}
\caption{Landmark-to-face on CelebA-HQ.}
\end{subfigure}\hfill
\begin{subfigure}{0.492\linewidth}
\centering
\includegraphics[width=\linewidth]{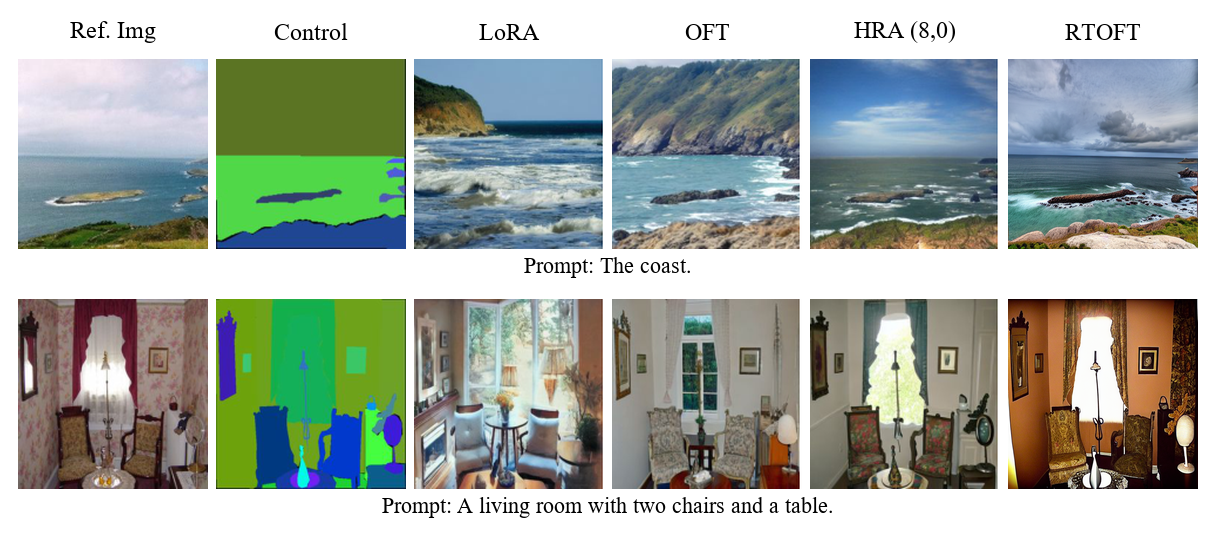}
\caption{Segmentation-to-image on ADE20K.}
\end{subfigure}
\caption{Examples of controllable generation with Stable Diffusion, two for each task.
All examples are in Figures~\ref{fig:l2f-full} and~\ref{fig:s2i-full}. The methods use the
settings of Table~\ref{tab:diffusion}, except for HRA. HRA\,(8,0) denotes
HRA$_{r=8,\lambda=0}$, with which these images are generated, and the tables
report HRA at $\lambda=10^{-3}$, the value used by \citet{yuan2024hra} in their controllable-generation experiments. RTOFT denotes RTOFT-B$_{2\,\mathrm{blocks}}$, which splits each adapted matrix into two diagonal blocks. }
\label{fig:qualitative}
\end{figure}

\begin{figure}[p]
\centering
\includegraphics[width=\linewidth,height=0.92\textheight,keepaspectratio]{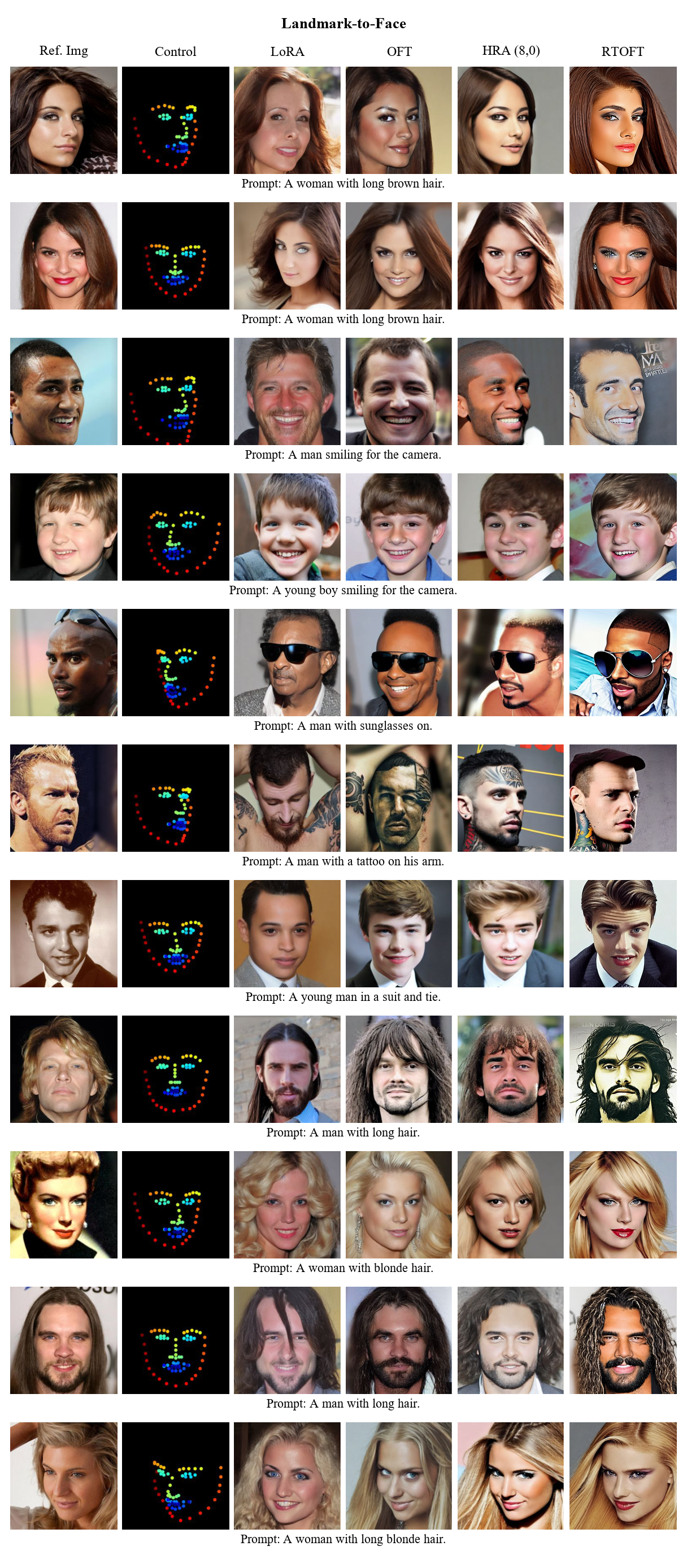}
\caption{Complete landmark-to-face comparison on CelebA-HQ.}
\label{fig:l2f-full}
\end{figure}

\begin{figure}[p]
\centering
\includegraphics[width=\linewidth,height=0.92\textheight,keepaspectratio]{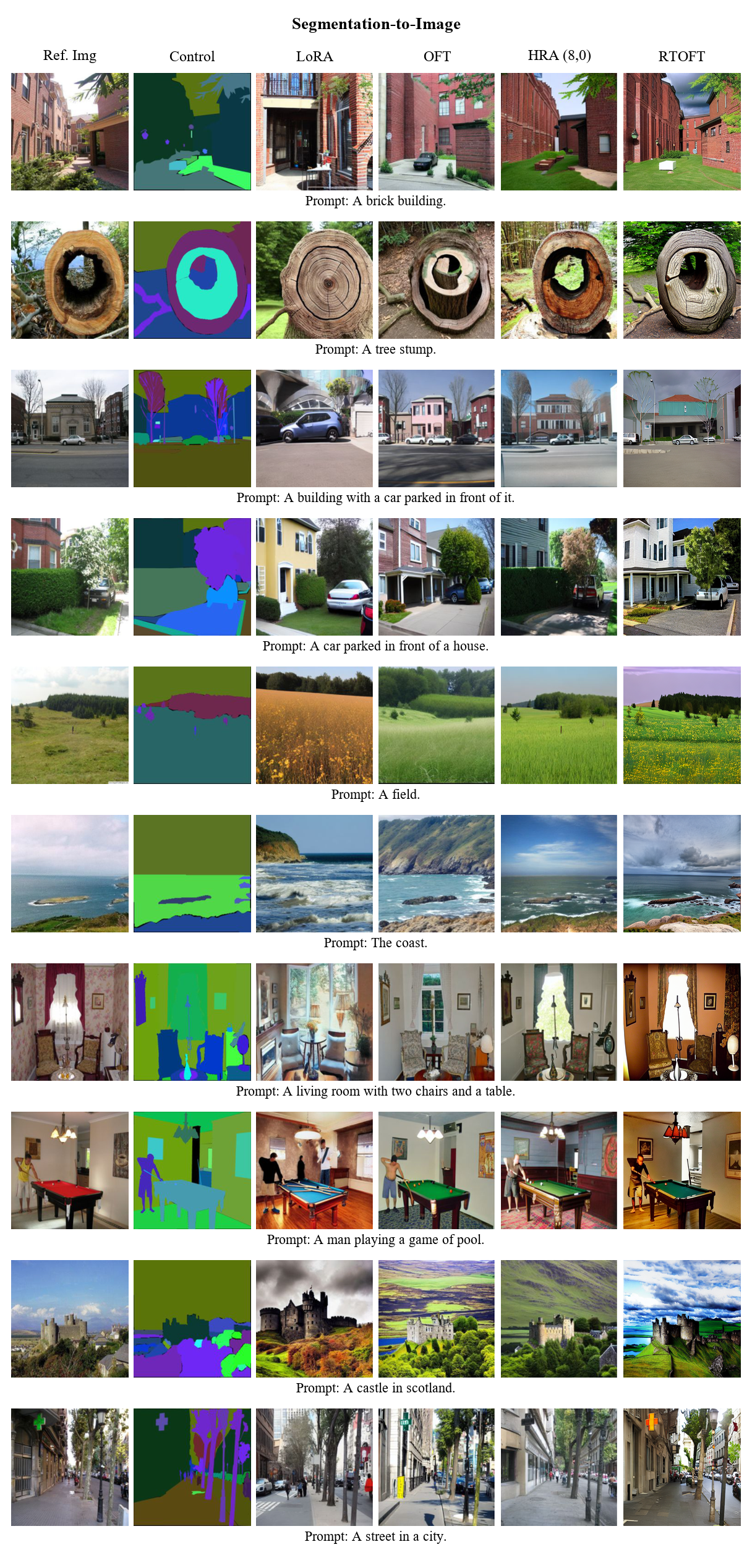}
\caption{Complete segmentation-to-image comparison on ADE20K.}
\label{fig:s2i-full}
\end{figure}

Figures~\ref{fig:l2f-full} and~\ref{fig:s2i-full} show all generated examples, from which those in
Figure~\ref{fig:qualitative} are taken.

\section{Implementation Details}\label{app:impl}

\subsection{Natural Language Understanding}

The backbone is DeBERTaV3-base \citep{he2021debertav3}, and we report two protocols.

The validation-score protocol of Table~\ref{tab:glue-val} follows \citet{liu2024parameter} and \citet{zhang2023adalora} and reports the best score on the original validation split, as in the published results. The baseline rows are taken from the literature, and we run OFTv2 \citep{qiu2025orthogonal} and RTOFT under both schedules in one pipeline at block size $b=16$ on the same five seeds. The full setup is given in Table~\ref{tab:hp-glue-val} and follows the GLUE configuration of \citet{liu2024parameter} and \citet{yuan2024hra}.

The test-split protocol of Table~\ref{tab:glue-test} follows \citet{wu2025efficient}, which adopts it from earlier work on GLUE evaluation \citep{wu2024red,wu2024reft,bini2025delora}. The public validation set of each task is split into two halves under a fixed seed. The first half is used to select the checkpoint, and the second half is used for the reported score. The learning rates are given in Table~\ref{tab:hp-glue-test}, at block size $b=32$. Weight decay is $0$ on every GLUE task. The drift trigger reads $\|S\|_2$ after every optimizer step and restores before the next forward pass whenever the reading reaches $\tau$.

CoLA is scored by Matthews correlation, STS-B by Pearson correlation, and the other four tasks by accuracy. Each reported average is the arithmetic mean over the tasks in that table.

A paired difference $x\pm s$ on GLUE is the mean $x$ of the differences over the five shared seeds, computed before rounding, and its standard error $s$, so that the $t$ statistic with four degrees of freedom is $t(4)=x/s$. The subscripts in the GLUE tables are standard deviations over seeds. The paired differences in the six-task average between RTOFT-B and OFTv2, PSOFT, and LoRA have $t(4)=4.3$,
$6.9$, and $5.1$, respectively, and all are positive on every seed. All methods share the learning-rate search of the test-split protocol. The range is from $5\times 10^{-5}$ to $5\times 10^{-3}$. For each method, we examine learning rates in $\{10^{-4},5\times10^{-4},10^{-3}\}$ and then one or two values between the best two, and use its own best value. Every score in Tables~\ref{tab:glue-test} and~\ref{tab:glue-test-full} is a mean over five seeds, whose standard deviations are in Table~\ref{tab:glue-sd}. The training time and peak memory are measured on one NVIDIA V100.

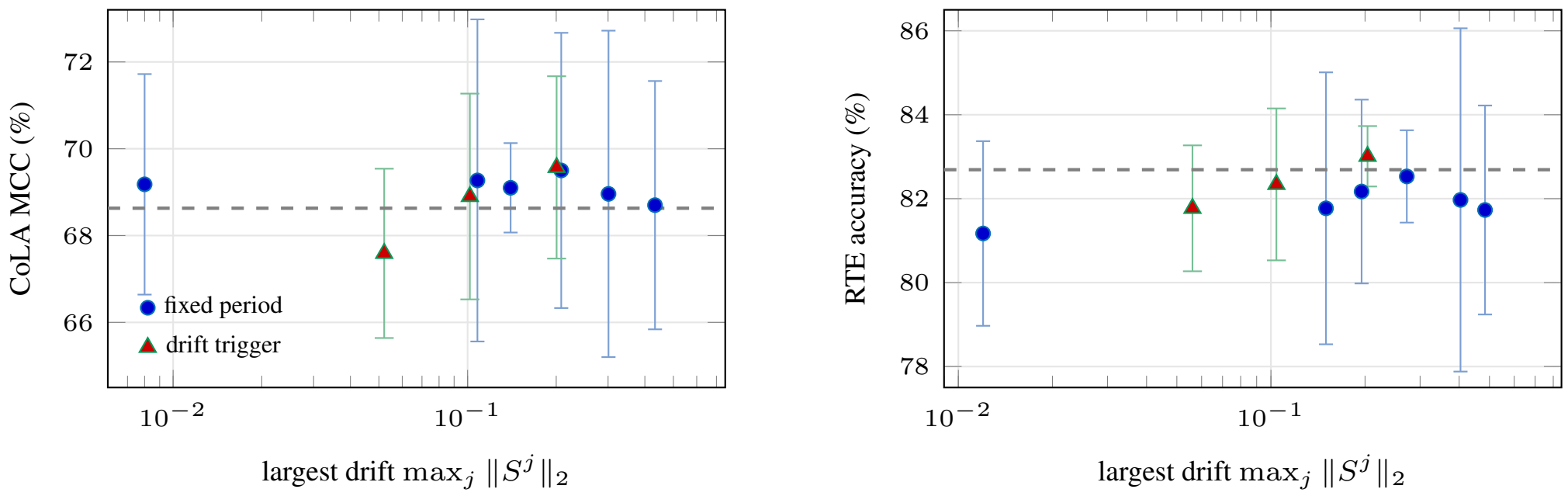
\begin{figure}[t]
\centering
\begin{minipage}{0.49\linewidth}
\centering
\begin{tikzpicture}
\begin{semilogxaxis}[
  width=\linewidth, height=4.8cm,
  xlabel={largest drift $\max_j\|S^j\|_2$}, ylabel={CoLA MCC (\%)},
  xmin=0.006, xmax=0.75, ymin=64.5, ymax=73.2,
  grid=major, grid style={gray!20},
  tick label style={font=\tiny}, label style={font=\scriptsize},
  legend style={font=\tiny, at={(0.03,0.03)}, anchor=south west, draw=none, fill=none},
]
\addplot+[only marks, mark=*, mark size=1.7pt, color=NavyBlue,
  error bars/.cd, y dir=both, y explicit, error bar style={color=NavyBlue!55}]
coordinates {
  (0.008,69.18) +- (0,2.54)
  (0.108,69.27) +- (0,3.71)
  (0.140,69.10) +- (0,1.03)
  (0.208,69.50) +- (0,3.17)
  (0.301,68.96) +- (0,3.76)
  (0.433,68.70) +- (0,2.86)
};
\addplot+[only marks, mark=triangle*, mark size=2.3pt, color=ForestGreen,
  error bars/.cd, y dir=both, y explicit, error bar style={color=ForestGreen!55}]
coordinates {
  (0.0521,67.59) +- (0,1.95)
  (0.1018,68.90) +- (0,2.37)
  (0.2006,69.57) +- (0,2.10)
};
\addplot[dashed, gray, thick] coordinates {(0.006,68.63) (0.75,68.63)};
\legend{fixed period, drift trigger}
\end{semilogxaxis}
\end{tikzpicture}
\\[-2pt]{\scriptsize (a) CoLA, 5 seeds}
\end{minipage}\hfill
\begin{minipage}{0.49\linewidth}
\centering
\begin{tikzpicture}
\begin{semilogxaxis}[
  width=\linewidth, height=4.8cm,
  xlabel={largest drift $\max_j\|S^j\|_2$}, ylabel={RTE accuracy (\%)},
  xmin=0.009, xmax=0.85, ymin=77.5, ymax=86.5,
  grid=major, grid style={gray!20},
  tick label style={font=\tiny}, label style={font=\scriptsize},
]
\addplot+[only marks, mark=*, mark size=1.7pt, color=NavyBlue,
  error bars/.cd, y dir=both, y explicit, error bar style={color=NavyBlue!55}]
coordinates {
  (0.012,81.17) +- (0,2.20)
  (0.150,81.77) +- (0,3.24)
  (0.195,82.17) +- (0,2.19)
  (0.272,82.53) +- (0,1.10)
  (0.404,81.97) +- (0,4.09)
  (0.484,81.73) +- (0,2.49)
};
\addplot+[only marks, mark=triangle*, mark size=2.3pt, color=ForestGreen,
  error bars/.cd, y dir=both, y explicit, error bar style={color=ForestGreen!55}]
coordinates {
  (0.0561,81.77) +- (0,1.50)
  (0.1041,82.34) +- (0,1.81)
  (0.2037,83.01) +- (0,0.72)
};
\addplot[dashed, gray, thick] coordinates {(0.009,82.69) (0.85,82.69)};
\end{semilogxaxis}
\end{tikzpicture}
\\[-2pt]{\scriptsize (b) RTE, 3 seeds}
\end{minipage}
\caption{Score against the largest drift for fixed periods and drift triggers over ten epochs
(RTOFT-B, $b{=}32$; mean $\pm$ standard deviation). The periods range
from $2680$ restorations to none, and the tolerances are $0.05$, $0.10$, and $0.20$. The dashed line is one standard error below the best score of each task.}
\label{fig:period}
\end{figure}

The sweep of restoration periods in Figure~\ref{fig:period} uses RTOFT-B at $b=32$ over ten epochs with the learning rate of the main experiments for each task, five seeds on CoLA, and three seeds on RTE. The period is counted in optimizer steps. Each restoration is performed before the next update, and no restoration is performed at the last step. The largest drift $\max_j\|S^j\|_2$ is the maximum over the run of the block spectral norms read immediately before each restoration, where $S^j$ is $S$ after step $j$.

\subsection{Commonsense Reasoning}\label{app:commonsense}
All methods are tuned on one learning-rate grid from $10^{-4}$ to $5\times 10^{-3}$, which is searched as on GLUE, and each method uses its own best value. Full finetuning is also run at $10^{-5}$ and $2\times10^{-5}$, below this grid, and Table~\ref{tab:commonsense} reports the better rate, $2\times10^{-5}$.
The backbone is Qwen3.5-9B (base), with 8.95B total parameters.\footnote{\url{https://huggingface.co/Qwen/Qwen3.5-9B}} We evaluate on the eight-task suite of Table~\ref{tab:commonsense}, comprising BoolQ, PIQA, SIQA, HellaSwag, WinoGrande, ARC-e, ARC-c and OBQA. All methods are trained under the configuration in Table~\ref{tab:hp-commonsense} with the same adapter placement.

We follow the Commonsense-15K protocol used by SVFT \citep{lingam2024svft} and PSOFT \citep{wu2025efficient}. This protocol takes a 15K instruction subset of the LLM-Adapters commonsense benchmark \citep{hu2023llmadapters} and evaluates on the same eight tasks. We run every method in Table~\ref{tab:commonsense} under this protocol on the same five seeds as in the other experiments. HRA \citep{yuan2024hra} runs out of memory at this scale. Training time is in \emph{mm:ss}, as in the other experiments except those on diffusion.

\paragraph{BoolQ.}
The lower score of the one-epoch period on BoolQ comes from a bias toward the majority class.
Answers can be extracted from all $3270$ responses. The model trained with the
period answers ``true'' on up to $90.9\%$ of the items ($2973$ of $3270$), compared with $62.2\%$
of the labels ($2033$ of $3270$) and $66.3\%$ for the trigger ($2167$).

\subsection{\texorpdfstring{Commonsense Reasoning on Llama-3.1-8B}{Commonsense Reasoning on Llama-3.1-8B}}\label{app:commonsense-llama}
To examine whether the comparison in Section~\ref{sec:commonsense} depends on the model family, we repeat the comparison between the drift trigger and the fixed one-epoch period on Llama-3.1-8B (base).\footnote{\url{https://huggingface.co/meta-llama/Llama-3.1-8B}} Both use RTOFT-B at $b=32$ under the configuration of Table~\ref{tab:hp-commonsense-llama}, are run from the same training script over five paired seeds on NVIDIA H200 GPUs, and are the same except for the restoration schedule. The trigger restores when $\|S\|_2$, read after every optimizer step, reaches $\tau=0.10$, and the period restores once per epoch. Evaluation uses the eight-task suite of Table~\ref{tab:commonsense}.

Under this configuration, the trigger exceeds the period on the eight-task average by $1.02\pm 0.32$ (mean $\pm$ standard error of the paired difference; $95\%$ CI $[0.13,1.91]$, positive on every seed; Table~\ref{tab:commonsense-llama}). It is also the more stable schedule, with a standard deviation of the average over seeds of $0.09$, compared with $0.69$. The largest gain is on HellaSwag, $10.81$ ($[3.92,17.70]$, positive on every seed), where the standard deviation over seeds is $0.38$ for the trigger, compared with $5.44$ for the period. On PIQA, the trigger is also higher, with a gain of $1.68$ ($[0.96,2.39]$, positive on every seed). Confidence intervals are $t$-based with four degrees of freedom.

\par\noindent\begin{minipage}{\linewidth}
\centering
\small
\renewcommand{\arraystretch}{1.15}
\setlength{\tabcolsep}{8pt}
\captionof{table}{Configuration for the Llama-3.1-8B commonsense experiments of Table~\ref{tab:commonsense-llama}, on the Commonsense-15K training set of Table~\ref{tab:hp-commonsense}. The two schedules share every other setting.}
\label{tab:hp-commonsense-llama}
\begin{tabular}{ll}
\toprule
Setting & Value \\
\midrule
Backbone            & Llama-3.1-8B (base) \\
Adapter             & RTOFT-B, block size $b=32$ \\
Training data       & \texttt{commonsense\_15k} (15K examples) \\
Target modules      & \texttt{q\_proj}, \texttt{k\_proj}, \texttt{v\_proj},
                      \texttt{up\_proj}, \texttt{down\_proj} \\
Learning rate       & $10^{-4}$ \\
Batch size / micro-batch & 64 / 2 \\
Epochs              & 3 \\
Warm-up steps       & 100 \\
Weight decay        & 0 \\
Cutoff length       & 512 \\
Restoration         & drift trigger at $\tau=0.10$, read after every step; \\
                    & or the fixed one-epoch period \\
Seeds               & five, paired across the two schedules \\
Hardware            & NVIDIA H200 \\
\bottomrule
\end{tabular}
\end{minipage}
\vspace{0.6em}

\par\noindent\begin{minipage}{\linewidth}
\centering
\small
\captionof{table}{Commonsense reasoning with Llama-3.1-8B (base), RTOFT-B at $b=32$ under the configuration of Table~\ref{tab:hp-commonsense-llama}. The two schedules use the same training script over five paired seeds. Accuracy cells are means over five seeds with standard deviations as subscripts. The difference row gives trigger minus period as mean $\pm$ standard error over the five seed pairs, and the last row counts the seed pairs in which the trigger is higher. Confidence intervals quoted in the text are $t$-based with four degrees of freedom.}
\label{tab:commonsense-llama}
\setlength{\tabcolsep}{3.2pt}
\begin{adjustbox}{max width=\linewidth}
\begin{tabular}{lccccccccc}
\toprule
Schedule & BoolQ & PIQA & SIQA & HellaSwag & WinoG. & ARC-e & ARC-c & OBQA & Avg. \\
\midrule
One-epoch period & 67.34$_{1.73}$ & 82.82$_{0.50}$ & 69.72$_{1.12}$ & 70.30$_{5.44}$ & 66.53$_{0.56}$ & 89.33$_{0.36}$ & 75.82$_{0.07}$ & 77.29$_{0.42}$ & 74.89$_{0.69}$ \\
Drift trigger ($\tau{=}0.10$) & 67.70$_{0.29}$ & 84.50$_{0.23}$ & 69.23$_{0.21}$ & 81.11$_{0.38}$ & 63.39$_{0.73}$ & 89.68$_{0.23}$ & 75.70$_{0.23}$ & 76.00$_{0.24}$ & 75.91$_{0.09}$ \\
\midrule
Trigger minus period & $+0.35_{\pm0.81}$ & $+1.68_{\pm0.26}$ & $-0.48_{\pm0.55}$ & $+10.81_{\pm2.48}$ & $-3.14_{\pm0.48}$ & $+0.35_{\pm0.24}$ & $-0.12_{\pm0.11}$ & $-1.29_{\pm0.16}$ & $+1.02_{\pm0.32}$ \\
Seed pairs, trigger higher & 2/5 & 5/5 & 2/5 & 5/5 & 0/5 & 4/5 & 2/5 & 0/5 & 5/5 \\
\bottomrule
\end{tabular}
\end{adjustbox}
\end{minipage}
\vspace{0.6em}

\subsection{PEFT-Arena Stability--Plasticity Evaluation}
We adhere to the released protocol of \citet{huang2026peftarena} without modification for the Qwen2.5-7B backbone, the mathematics domain, and the supervised-finetuning regime. Target performance is measured on Math-500, AMC23 and AIME24, and general performance on IFEval, NQ and BBH, each reported together with its change against the un-adapted base model. Every row uses the same data, budget, optimizer, and evaluation code. We run RTOFT-B at block size 32, at which its number of trainable parameters equals that of OFT at block size 32. For comparison with restoration during training, under the period or the trigger, we consider OFTv2 and training without restoration. The latter is evaluated with and without a polar projection at the end (Section~\ref{sec:restart-necessary}).

\subsection{Controllable Generation}

We follow the protocol of \citet{qiu2023oft}, \citet{liu2024parameter}, and \citet{yuan2024hra} on Stable Diffusion \citep{rombach2022ldm}. The two tasks are segmentation-to-image on ADE20K and landmark-to-face on CelebA-HQ, and the control signal is injected in the same way for all methods. Control fidelity on segmentation-to-image is measured by segmenting the generated images and comparing the result with the conditioning map (mIoU, mAcc, aAcc). On landmark-to-face, it is the landmark error between the control signal and the prediction. The baselines are OFTv2 \citep{qiu2025orthogonal}, DreamBooth \citep{ruiz2023dreambooth}, ControlNet \citep{zhang2023controlnet}, T2I-Adapter \citep{mou2024t2iadapter}, LoRA \citep{hu2022lora}, OFT and its constrained variant \citep{qiu2023oft}, BOFT \citep{liu2024parameter}, and HRA \citep{yuan2024hra}. Following
\citet{liu2024parameter}, all methods are run on one grid, with
budgets of roughly $1$--$2$M, $10$M, $20$M, and $26$M trainable parameters and learning rates from $10^{-4}$ to
$10^{-3}$. We report the best entry of each method on this grid. The best entry of LoRA is $r=16$, and
HRA is reported at $r=8,\lambda=10^{-3}$, the configuration used by \citet{yuan2024hra} in their controllable-generation experiments. All rows use the same hardware,
number of steps, and batch size.
The training times in Table~\ref{tab:diffusion-runtime} are in \emph{h:mm:ss}.

The drift trigger in Table~\ref{tab:diffusion} uses $\tau=0.10$ and reads
$\|S\|_2$ after every optimizer step. Each adapted matrix is split into two diagonal blocks of half its input dimension ($160$, $320$, or $640$ in the UNet and $384$ on the text side), and the exact spectral norm is computed only for the blocks whose row-norm upper bound reaches $\tau$. All other settings are the same as those of the one-epoch
period. The trigger restores $1{,}872$ times in $27{,}820$ steps ($6.7\%$ of the readings).
The largest reading is $0.12$, and the drift in every forward pass is below $\tau$.
The peak memory is the same as that of the period, and the training time is $21{:}35{:}36$, compared with
$21{:}26{:}06$, which is $0.7\%$ longer.

\subsection{Ablation Protocols}

The projections in Appendix~\ref{app:ablation-detail} use the tangent parameters of runs without restoration and with the one-epoch period
on CoLA and STS-B under the test-split protocol and with multiple seeds. The tangent parameters are saved at $20$, $40$, $60$, $80$, and $100\%$ of training. Each saved model is evaluated
as it is and after a polar projection.

The comparison of learning-rate schedules in Appendix~\ref{app:ablation-detail} uses the one-epoch period and the trigger with $\tau=0.20$, which have about the
same largest drift on CoLA. We let the learning-rate schedule vary in $\{$cosine, linear, constant$\}$ and its scale in $\{0.5\times,2\times\}$, and also consider $20$ epochs, with multiple seeds each, while keeping the protocol of the ten-epoch sweep
and reading the drift at every step.

\input{app_restoration}

\end{document}

%% file: figures/fig2_evidence.tex
\definecolor{rtInk}{HTML}{1E293B}
\definecolor{rtMute}{HTML}{64748B}
\definecolor{rtFaint}{HTML}{A3ADBC}
\definecolor{rtGrid}{HTML}{E9EDF3}
\definecolor{rtBlue}{HTML}{2563EB}
\definecolor{rtBlueD}{HTML}{1D4ED8}
\definecolor{rtBlueM}{HTML}{93B4F7}
\definecolor{rtBlueL}{HTML}{E4EDFE}
\definecolor{rtGold}{HTML}{F59E0B}
\definecolor{rtGoldE}{HTML}{D97706}
\definecolor{rtGoldD}{HTML}{B45309}
\definecolor{rtGoldL}{HTML}{FEF3D6}
\definecolor{rtGray}{HTML}{A9B2C0}
\newcommand{\rtTitle}[2]{{\fontsize{7}{8.4}\selectfont\textbf{(#1)}\;#2}}
\newcommand{\rtBar}[5]{%
  \makebox[#4cm][r]{\ifdim #1pt<0pt\textcolor{#2}{\rule[-0.2pt]{\fpeval{-(#1)*#3}cm}{3.6pt}}\fi}%
  \textcolor{rtFaint}{\rule[-1.4pt]{0.3pt}{6pt}}%
  \makebox[#5cm][l]{\ifdim #1pt>0pt\textcolor{#2}{\rule[-0.2pt]{\fpeval{(#1)*#3}cm}{3.6pt}}\fi}}
\newcommand{\rtT}[2]{\rtBar{#1}{#2}{0.030}{0.04}{0.47}}   %
\newcommand{\rtG}[2]{\rtBar{#1}{#2}{0.0225}{0.51}{0.03}}   %
\begin{tikzpicture}[x=1cm, y=1cm, line join=round,
    font=\fontsize{6}{7.2}\selectfont,
    >={Stealth[length=3pt, width=2.6pt]},
    rtbox/.style={rounded corners=2.2pt, line width=0.55pt, inner xsep=2.5pt, inner ysep=2.3pt,
                  align=center, minimum width=3.31cm}]
\useasboundingbox (0,0) rectangle (13.97,6.2);

\node[anchor=north west, inner sep=0pt] at (0,6.2) {\rtTitle{a}{GLUE: score and training time}};
\begin{axis}[
    at={(0.58cm,0.72cm)}, anchor=south west, scale only axis,
    width=3.62cm, height=5.0cm,
    xmin=0.855, xmax=1.66, ymin=85.75, ymax=90.3,
    axis x line*=bottom, axis y line*=left,
    axis line style={rtFaint, line width=0.4pt},
    tick style={rtFaint, line width=0.35pt}, major tick length=1.8pt,
    xtick={1.0,1.2,1.4,1.6},
    xticklabels={1.0\texttimes, 1.2\texttimes, 1.4\texttimes, 1.6\texttimes},
    ytick={86,87,88,89,90},
    ymajorgrids, grid style={rtGrid, line width=0.35pt},
    tick label style={font=\fontsize{5.5}{6}\selectfont, text=rtMute},
    xlabel={training time (relative to LoRA)}, ylabel={GLUE avg.\ (\%)},
    label style={font=\fontsize{5.8}{6.5}\selectfont, text=rtMute},
    xlabel shift=-4.5pt, ylabel shift=-5pt,
    clip=false,
  ]
  \draw[rtFaint, line width=0.4pt, dash pattern=on 1.2pt off 1.1pt] (axis cs:1,85.75) -- (axis cs:1,90.3);
  \addplot[only marks, mark=*, mark size=1.35pt, color=rtGray, mark options={draw=white, line width=0.25pt}]
    coordinates {(1.00,87.91) (0.92,87.16) (1.00,87.68) (0.97,86.05) (1.05,88.10)};
  \addplot[only marks, mark=triangle*, mark size=1.7pt, color=rtInk!70, mark options={draw=white, line width=0.25pt}]
    coordinates {(1.08,86.92)};
  \addplot[only marks, mark=square*, mark size=1.25pt, color=rtGold, mark options={draw=rtGoldE, line width=0.4pt}]
    coordinates {(1.10,88.81) (1.23,88.01) (1.58,87.76)};
  \addplot[only marks, mark=triangle*, mark size=1.5pt, color=rtGold, mark options={rotate=-90, draw=rtGoldE, line width=0.4pt}]
    coordinates {(1.655,87.94)};
  \draw[->, rtBlue, line width=0.6pt, shorten <=2.6pt, shorten >=3.6pt] (axis cs:1.10,88.81) -- (axis cs:0.97,89.90);
  \fill[rtBlueL] (axis cs:0.97,89.90) circle[radius=0.16cm];
  \node[star, star points=5, star point ratio=2.3, fill=rtBlue, draw=white, line width=0.4pt,
        inner sep=0pt, minimum size=0.25cm] at (axis cs:0.97,89.90) {};
  \begin{scope}[every node/.style={inner sep=0.8pt, font=\fontsize{5.5}{6}\selectfont}]
    \node[anchor=west, text=rtBlueD, font=\fontsize{6}{7}\selectfont\bfseries] at (axis cs:0.995,89.93) {RTOFT};
    \node[anchor=west, text=rtBlueD] at (axis cs:1.062,89.34) {\textminus12\% time, +1.09};
    \node[anchor=west, text=rtGoldD] at (axis cs:1.115,88.83) {OFTv2};
    \node[anchor=north, text=rtGoldD] at (axis cs:1.23,87.95) {PSOFT};
    \node[anchor=north, text=rtGoldD] at (axis cs:1.58,87.70) {BOFT};
    \node[anchor=south east, text=rtGoldD] at (axis cs:1.665,88.02) {HRA 2.06\texttimes};
    \node[anchor=east, text=rtMute] at (axis cs:0.99,87.93) {LoRA};
    \node[anchor=west, text=rtMute] at (axis cs:1.06,88.12) {GiVA};
    \node[anchor=west, text=rtMute] at (axis cs:1.01,87.62) {DoRA};
    \node[anchor=north, text=rtMute] at (axis cs:0.92,87.10) {PiSSA};
    \node[anchor=west, text=rtMute] at (axis cs:1.09,86.92) {Full FT};
    \node[anchor=west, text=rtMute] at (axis cs:0.985,86.05) {LoRA-XS};
  \end{scope}
\end{axis}

\begin{scope}[shift={(4.59,0)}]
  \node[anchor=north west, inner sep=0pt] at (0,6.2) {\rtTitle{b}{One step and one restoration}};
  \node[rtbox, draw=rtInk, fill=white] (w) at (1.86,5.43)
      {$W_n{=}\,W^{\mathrm{pre}}R_n,\;R_n{\in}\,\mathcal{SO}(d)$};
  \node[rtbox, draw=rtBlue, fill=rtBlueL] (in) at (1.86,4.4)
      {{\color{rtBlueD}\textbf{tangent update}, every step}\\[0.6pt]
       $W_n(I{+}S)$, \ $S^\top\!=-S$\\[0.6pt]
       one multiply, \ $O(db)$};
  \node[rtbox, draw=rtGoldE, fill=rtGoldL] (re) at (1.86,2.36)
      {{\color{rtGoldD}\textbf{restore}, only when triggered}\\[0.6pt]
       $W_{n+1}=W_n\,\mathrm{polar}(I{+}S)$\\[0.6pt]
       $S\leftarrow0$, \ $O(db^2{+}d\,d_{\mathrm{out}}b)$};
  \draw[->, rtInk, line width=0.5pt] (w) -- (in);
  \draw[->, rtGoldE, line width=0.6pt] (in) -- (re);
  \node[anchor=west, text=rtGoldD, inner sep=0.5pt] at (1.95,3.53) {$\|S\|_2\ge\tau$};
  \node[anchor=west, text=rtMute, inner sep=0.5pt] at (1.95,3.21) {$\delta(I{+}S)=\|S\|_2^2$};
  \draw[->, rtBlue, line width=0.55pt] ([yshift=-5pt]in.east) -- ++(0.17,0) |- ([yshift=5pt]in.east);
  \node[rotate=-90, anchor=south, inner sep=1.2pt, text=rtBlueD] at ([xshift=0.17cm]in.east) {repeat};
  \path let \p1=(w.west), \p2=(re.west) in coordinate (lx) at ({min(\x1,\x2)-0.16cm},0);
  \draw[->, rtGoldE, line width=0.55pt] (re.west) -- (re.west -| lx) -- (w.west -| lx) -- (w.west);
  \node[rotate=90, anchor=south, inner sep=1.2pt, text=rtGoldD]
      at ($(lx |- re.north)!0.5!(lx |- w.south)$) {orthogonal again};
  \node[rtbox, draw=rtFaint, fill=white, dash pattern=on 1.4pt off 1.1pt] (cmp) at (1.86,0.66)
      {{\color{rtMute}prior OFT, at \emph{every} step:}\\[0.6pt]
       build $\widetilde R_k(Q)$, backprop through it\\[0.6pt]
       $O(dkb^2)$ per step};
\end{scope}

\node[anchor=north west, inner sep=0pt] at (8.62,6.2) {\rtTitle{c}{PEFT-Arena}};
\node[anchor=north west, inner sep=0pt, text=rtMute] at (8.62,5.86)
    {Qwen2.5-7B, math; change vs.\ base ($35.30$ / $46.97$)};
\node[anchor=north west, inner sep=0pt] (tab) at (8.56,5.52) {%
  \fontsize{5.6}{6.6}\selectfont\renewcommand{\arraystretch}{1.0}\setlength{\tabcolsep}{1.3pt}%
  \begin{tabular}{@{}l r r@{\;}l r@{\;}l@{}}
    \toprule
    Method & Param & \multicolumn{2}{c}{Target $\uparrow$} & \multicolumn{2}{c}{General $\uparrow$} \\
    \midrule
    Full FT                      & 7.61B  & $+15.33$ & \rtT{15.33}{rtInk!55} & $-12.75$ & \rtG{-12.75}{rtInk!55} \\
    OFTv2 {\color{rtMute}b32}    & 17.55M & $+11.63$ & \rtT{11.63}{rtGold}   & $-2.60$  & \rtG{-2.60}{rtGold} \\
    \midrule
    LoRA {\color{rtMute}r8}      & 20.19M & $+7.17$  & \rtT{7.17}{rtGray}    & $-7.75$  & \rtG{-7.75}{rtGray} \\
    PiSSA {\color{rtMute}r8}     & 20.19M & $+9.23$  & \rtT{9.23}{rtGray}    & $-22.19$ & \rtG{-22.19}{rtGray} \\
    DoRA {\color{rtMute}r8}      & 21.58M & $+7.03$  & \rtT{7.03}{rtGray}    & $-6.72$  & \rtG{-6.72}{rtGray} \\
    KeepLoRA {\color{rtMute}r8}  & 20.19M & $+5.23$  & \rtT{5.23}{rtGray}    & $-3.22$  & \rtG{-3.22}{rtGray} \\
    VeRA {\color{rtMute}r256}    & 1.44M  & $+4.13$  & \rtT{4.13}{rtGray}    & $+0.28$  & \rtG{0.28}{rtGray} \\
    IA$^3$                       & 1.82M  & $-0.17$  & \rtT{-0.17}{rtGray}   & $-2.26$  & \rtG{-2.26}{rtGray} \\
    \midrule
    \rowcolor{rtBlueL}
    \textbf{RTOFT} $\tau$ (ours) & 17.55M & $\mathbf{+11.89}$ & \rtT{11.89}{rtBlue} & $\mathbf{-2.35}$ & \rtG{-2.35}{rtBlue} \\
    RTOFT period                & 17.55M & $+11.91$ & \rtT{11.91}{rtBlueM}  & $-2.53$  & \rtG{-2.53}{rtBlueM} \\
    RTOFT no restoration        & 17.55M & $+8.53$  & \rtT{8.53}{rtBlueM}   & $-13.35$ & \rtG{-13.35}{rtBlueM} \\
    \bottomrule
  \end{tabular}};
\node[anchor=north west, inner sep=0pt] at (8.62,1.76) {\rtTitle{d}{Restorations and largest drift}};
\node[anchor=north east, inner sep=0pt, text=rtMute] at (13.97,1.74) {same $\tau=0.10$};
\node[anchor=south west, inner sep=0pt, text=rtMute, font=\fontsize{5.5}{6}\selectfont] at (10.95,1.08) {restorations (log scale)};
\node[anchor=south, inner sep=0pt, text=rtMute, font=\fontsize{5.5}{6}\selectfont] at (13.33,1.08) {max $\|S\|_2$};
\draw[rtBlueM, line width=0.45pt, dash pattern=on 1.2pt off 1pt] (13.14,0.06) -- (13.14,1.06);
\foreach \y/\m/\t/\n/\len/\dr in {
    0.86/Qwen3.5-9B/commonsense/10/0.43/0.10,
    0.53/Qwen2.5-7B/PEFT-Arena/40/0.69/0.10,
    0.20/Stable Diffusion/S2I/{1{,}872}/1.40/0.12} {
  \node[anchor=west, inner sep=0pt, text=rtInk, font=\fontsize{5.5}{6.2}\selectfont]
    at (8.62,\y) {\m\ {\color{rtMute}\t}};
  \fill[rtBlue] (10.95,\y-0.068) rectangle ++(\len,0.136);
  \node[anchor=west, inner sep=1.2pt, text=rtBlueD] at (10.95+\len,\y) {$\n$};
  \fill[rtBlue] ({12.94+(\dr-0.08)/0.06*0.6},\y) circle[radius=0.052];
  \node[anchor=west, inner sep=1.2pt, text=rtMute, font=\fontsize{5.5}{6}\selectfont]
    at ({13.0+(\dr-0.08)/0.06*0.6},\y) {\dr};
}
\end{tikzpicture}

%% file: app_restoration.tex
\section{Additional Evidence on Restoration}\label{app:restoration-roles}

In this section, we provide additional results on whether the gain of restoration comes from orthogonality itself or from its effect on the training path, on its interaction with the moment estimates of Adam, and on the parameterizations for which it helps.

\paragraph{Orthogonality and the training path.}
Each restoration makes the transform exactly orthogonal and also changes the training path that follows. Two kinds of comparison separate these effects. In the first, we make the model orthogonal, or nearly so, without restoring it during training. We project the final model or the checkpoints of a run without restoration onto the orthogonal group (Section~\ref{sec:restart-necessary} and Appendix~\ref{app:ablation-detail}) or use an $L_2$ penalty on $S$ (Table~\ref{tab:l2}). The projections do not recover the gain of restoration, and the penalty either leaves the drift above $\tau$ or reduces the score. In the second, we vary when restoration happens during training (Table~\ref{tab:lr-stress}). At twice the tuned SST-2 learning rate, restoring after every step performs as well as the trigger, and a fixed period with the same number of restorations recovers less than half of the gain of the trigger over the one-epoch period. The gain of restoration therefore comes from keeping the training path near the orthogonal group without limiting the total rotation, and the timing of the restorations relative to the drift matters more than their number. The total rotation is not limited because each restoration multiplies the accumulated rotation into $W$ and resets $S$ to zero, whereas a penalty on $S$ limits it. Moreover, if the loss has a bounded, Lipschitz gradient near the orthogonal group, the tolerance is small, and the number of steps in a cycle times the step size and the gradient bound is at most $2\tau$, each cycle of gradient descent on $S$ ends within $O(\tau^2)$ of gradient descent that starts from the same transform and retracts onto the group after every step (Lemma~\ref{lem:tracking}).

\paragraph{Optimizer state.}
A restoration keeps the moments of Adam unchanged, and at a fixed loss gradient, the gradient with respect to $S$ changes by at most $\arctan\|S\|_2\,\|W^\top\nabla\mathcal{L}\|_2$ (Proposition~\ref{cor:frame}). In our runs, a restoration happens at a drift between $0.10$ and $0.12$ (Figure~\ref{fig:hero}(d)), so this change is at most about $0.12\,\|W^\top\nabla\mathcal{L}\|_2$. Doubling $\tau$ to $0.20$ roughly doubles the bound, and no GLUE task decreases by more than one standard error (Table~\ref{tab:tau020}). Resetting the moments instead would discard the accumulated momentum. An exact transformation of the moments cannot be computed from the stored state, because the new gradient depends on the symmetric part of $W^\top\nabla\mathcal{L}$, which the moments do not contain.

\paragraph{When restoration helps.}
Whether restoration helps depends on how much drift there is to remove and on the gradient after a reset (Proposition~\ref{prop:reset-signal}). In the block, subspace, and butterfly parameterizations, $S$ is linear in its parameters, so the gradient right after a reset is the Riemannian gradient $g=\operatorname{skew}(W^\top\nabla\mathcal{L})$ restricted to the support. Restoration improves the block, subspace, and butterfly variants on all six tasks (Tables~\ref{tab:variants}, \ref{tab:psoft}, and~\ref{tab:tboft}). In RTOFT-L, the gradient with respect to $U$ is zero right after a reset (Appendix~\ref{app:variants}). The amount of drift matters as well. The default $k=5$ map of OFTv2 has a smaller defect than the tangent map (Section~\ref{sec:order-restart}), and restoration does not improve its six-task average.

%% file: main.bbl
\begin{thebibliography}{54}
\providecommand{\natexlab}[1]{#1}
\providecommand{\url}[1]{\texttt{#1}}
\expandafter\ifx\csname urlstyle\endcsname\relax
  \providecommand{\doi}[1]{doi: #1}\else
  \providecommand{\doi}{doi: \begingroup \urlstyle{rm}\Url}\fi

\bibitem[Ablin \& Peyr\'{e}(2022)Ablin and Peyr\'{e}]{ablin2022landing}
Pierre Ablin and Gabriel Peyr\'{e}.
\newblock Fast and accurate optimization on the orthogonal manifold without retraction.
\newblock In \emph{International Conference on Artificial Intelligence and Statistics (AISTATS)}, 2022.

\bibitem[Ablin et~al.(2024)Ablin, Vary, Gao, and Absil]{ablin2024infeasible}
Pierre Ablin, Simon Vary, Bin Gao, and P.-A. Absil.
\newblock Infeasible deterministic, stochastic, and variance-reduction algorithms for optimization under orthogonality constraints.
\newblock \emph{Journal of Machine Learning Research}, 25\penalty0 (389):\penalty0 1--38, 2024.

\bibitem[Absil et~al.(2008)Absil, Mahony, and Sepulchre]{absil2008optimization}
P.-A. Absil, Robert Mahony, and Rodolphe Sepulchre.
\newblock \emph{Optimization Algorithms on Matrix Manifolds}.
\newblock Princeton University Press, Princeton, NJ, 2008.
\newblock ISBN 978-0-691-13298-3.

\bibitem[Ba{\l}azy et~al.(2025)Ba{\l}azy, Banaei, Aberer, and Tabor]{wang2024loraxs}
Klaudia Ba{\l}azy, Mohammadreza Banaei, Karl Aberer, and Jacek Tabor.
\newblock {LoRA-XS}: Low-rank adaptation with extremely small number of parameters.
\newblock In \emph{ECAI 2025: 28th European Conference on Artificial Intelligence}, pp.\  3194--3201, 2025.
\newblock arXiv:2405.17604.

\bibitem[B{\'e}cigneul \& Ganea(2019)B{\'e}cigneul and Ganea]{becigneul2019riemannian}
Gary B{\'e}cigneul and Octavian-Eugen Ganea.
\newblock Riemannian adaptive optimization methods.
\newblock In \emph{International Conference on Learning Representations (ICLR)}, 2019.

\bibitem[Ben~Zaken et~al.(2022)Ben~Zaken, Goldberg, and Ravfogel]{benzaken2022bitfit}
Elad Ben~Zaken, Yoav Goldberg, and Shauli Ravfogel.
\newblock {BitFit}: Simple parameter-efficient fine-tuning for transformer-based masked language-models.
\newblock In \emph{Proceedings of the 60th Annual Meeting of the Association for Computational Linguistics (Volume 2: Short Papers)}, pp.\  1--9, 2022.

\bibitem[Bhatia(1997)]{bhatia1997matrix}
Rajendra Bhatia.
\newblock \emph{Matrix Analysis}, volume 169 of \emph{Graduate Texts in Mathematics}.
\newblock Springer, 1997.

\bibitem[Bini et~al.(2025)Bini, Girrbach, and Akata]{bini2025delora}
Massimo Bini, Leander Girrbach, and Zeynep Akata.
\newblock Decoupling angles and strength in low-rank adaptation.
\newblock In \emph{International Conference on Learning Representations (ICLR)}, 2025.
\newblock arXiv:2503.18225.

\bibitem[Boumal(2023)]{boumal2023intromanifolds}
Nicolas Boumal.
\newblock \emph{An Introduction to Optimization on Smooth Manifolds}.
\newblock Cambridge University Press, Cambridge, 2023.
\newblock ISBN 978-1-009-16615-7.
\newblock \doi{10.1017/9781009166164}.

\bibitem[Fan \& Hoffman(1955)Fan and Hoffman]{fan1955metric}
Ky~Fan and Alan~J. Hoffman.
\newblock Some metric inequalities in the space of matrices.
\newblock \emph{Proceedings of the American Mathematical Society}, 6\penalty0 (1):\penalty0 111--116, 1955.

\bibitem[Gangwar et~al.(2026)Gangwar, Deshmukh, Shavlovsky, Li, Mittal, Ying, and Kani]{giva}
Neeraj Gangwar, Rishabh Deshmukh, Michael Shavlovsky, Hancao Li, Vivek Mittal, Lexing Ying, and Nickvash Kani.
\newblock {GiVA}: Gradient-informed bases for vector-based adaptation.
\newblock In \emph{Proceedings of the 29th International Conference on Artificial Intelligence and Statistics (AISTATS)}, 2026.
\newblock arXiv:2604.21901.

\bibitem[Hairer et~al.(2006)Hairer, Lubich, and Wanner]{hairer2006geometric}
Ernst Hairer, Christian Lubich, and Gerhard Wanner.
\newblock \emph{Geometric Numerical Integration: Structure-Preserving Algorithms for Ordinary Differential Equations}, volume~31 of \emph{Springer Series in Computational Mathematics}.
\newblock Springer, 2 edition, 2006.

\bibitem[He et~al.(2023)He, Gao, and Chen]{he2021debertav3}
Pengcheng He, Jianfeng Gao, and Weizhu Chen.
\newblock {DeBERTaV3}: Improving {DeBERTa} using {ELECTRA}-style pre-training with gradient-disentangled embedding sharing.
\newblock In \emph{International Conference on Learning Representations (ICLR)}, 2023.
\newblock arXiv:2111.09543.

\bibitem[Higham(1997)]{higham1997orthonormality}
Desmond~J. Higham.
\newblock Time-stepping and preserving orthonormality.
\newblock \emph{BIT Numerical Mathematics}, 37\penalty0 (1):\penalty0 24--36, 1997.

\bibitem[Higham(1986)]{higham1986polar}
Nicholas~J. Higham.
\newblock Computing the polar decomposition---with applications.
\newblock \emph{SIAM Journal on Scientific and Statistical Computing}, 7\penalty0 (4):\penalty0 1160--1174, 1986.

\bibitem[Horn \& Johnson(1991)Horn and Johnson]{hornjohnson1991topics}
Roger~A. Horn and Charles~R. Johnson.
\newblock \emph{Topics in Matrix Analysis}.
\newblock Cambridge University Press, 1991.

\bibitem[Houlsby et~al.(2019)Houlsby, Giurgiu, Jastrzebski, Morrone, de~Laroussilhe, Gesmundo, Attariyan, and Gelly]{houlsby2019parameter}
Neil Houlsby, Andrei Giurgiu, Stanislaw Jastrzebski, Bruna Morrone, Quentin de~Laroussilhe, Andrea Gesmundo, Mona Attariyan, and Sylvain Gelly.
\newblock Parameter-efficient transfer learning for {NLP}.
\newblock In \emph{Proceedings of the 36th International Conference on Machine Learning}, volume~97 of \emph{Proceedings of Machine Learning Research}, pp.\  2790--2799, 2019.

\bibitem[Hu et~al.(2022)Hu, Shen, Wallis, Allen-Zhu, Li, Wang, Wang, and Chen]{hu2022lora}
Edward~J. Hu, Yelong Shen, Phillip Wallis, Zeyuan Allen-Zhu, Yuanzhi Li, Shean Wang, Lu~Wang, and Weizhu Chen.
\newblock {LoRA}: Low-rank adaptation of large language models.
\newblock In \emph{International Conference on Learning Representations (ICLR)}, 2022.
\newblock arXiv:2106.09685.

\bibitem[Hu et~al.(2023)Hu, Wang, Lan, Xu, Lim, Bing, Xu, Poria, and Lee]{hu2023llmadapters}
Zhiqiang Hu, Lei Wang, Yihuai Lan, Wanyu Xu, Ee-Peng Lim, Lidong Bing, Xing Xu, Soujanya Poria, and Roy Ka-Wei Lee.
\newblock {LLM-Adapters}: An adapter family for parameter-efficient fine-tuning of large language models.
\newblock In \emph{Proceedings of the 2023 Conference on Empirical Methods in Natural Language Processing}. Association for Computational Linguistics, 2023.
\newblock arXiv:2304.01933.

\bibitem[Huang et~al.(2026)Huang, Peng, Qiu, Kang, Wen, Sch{\"o}lkopf, and Liu]{huang2026peftarena}
Yangyi Huang, Ruotian Peng, Zeju Qiu, Jiale Kang, Yandong Wen, Bernhard Sch{\"o}lkopf, and Weiyang Liu.
\newblock {PEFT-Arena}: Understanding parameter-efficient finetuning from a stability-plasticity perspective.
\newblock \emph{arXiv preprint arXiv:2605.28819}, 2026.

\bibitem[Javaloy \& Vergari(2026)Javaloy and Vergari]{javaloy2026simple}
Adri\'an Javaloy and Antonio Vergari.
\newblock An embarrassingly simple way to optimize orthogonal matrices at scale.
\newblock In \emph{International Conference on Machine Learning}, 2026.

\bibitem[Jia et~al.(2017)Jia, Tao, Gao, and Xu]{jia2017svb}
Kui Jia, Dacheng Tao, Shenghua Gao, and Xiangmin Xu.
\newblock Improving training of deep neural networks via singular value bounding.
\newblock In \emph{IEEE Conference on Computer Vision and Pattern Recognition (CVPR)}, 2017.

\bibitem[Kang \& Yin(2026)Kang and Yin]{kang2026miss}
Jiale Kang and Qingyu Yin.
\newblock {MiSS}: Revisiting the trade-off in {LoRA} with an efficient shard-sharing structure.
\newblock In \emph{International Conference on Learning Representations (ICLR)}, 2026.
\newblock arXiv:2409.15371.

\bibitem[Khaled et~al.(2026)Khaled, Ozkara, Yu, Hong, and Park]{khaled2025muonbp}
Ahmed Khaled, Kaan Ozkara, Tao Yu, Mingyi Hong, and Youngsuk Park.
\newblock {MuonBP}: Faster {Muon} via block-periodic orthogonalization.
\newblock In \emph{International Conference on Learning Representations (ICLR)}, 2026.

\bibitem[Kopiczko et~al.(2024)Kopiczko, Blankevoort, and Asano]{kopiczko2024vera}
Dawid~Jan Kopiczko, Tijmen Blankevoort, and Yuki~M. Asano.
\newblock {VeRA}: Vector-based random matrix adaptation.
\newblock In \emph{International Conference on Learning Representations (ICLR)}, 2024.
\newblock arXiv:2310.11454.

\bibitem[Lezcano-Casado(2019)]{lezcanocasado2019trivializations}
Mario Lezcano-Casado.
\newblock Trivializations for gradient-based optimization on manifolds.
\newblock In \emph{Advances in Neural Information Processing Systems (NeurIPS)}, 2019.
\newblock arXiv:1909.09501.

\bibitem[Li et~al.(2020)Li, Fuxin, and Todorovic]{li2020cayley}
Jun Li, Li~Fuxin, and Sinisa Todorovic.
\newblock Efficient {R}iemannian optimization on the {S}tiefel manifold via the {C}ayley transform.
\newblock In \emph{International Conference on Learning Representations (ICLR)}, 2020.

\bibitem[Lingam et~al.(2024)Lingam, Tejaswi, Vavre, Shetty, Gudur, Ghosh, Dimakis, Choi, Bojchevski, and Sanghavi]{lingam2024svft}
Vijay Lingam, Atula Tejaswi, Aditya Vavre, Aneesh Shetty, Gautham~Krishna Gudur, Joydeep Ghosh, Alex Dimakis, Eunsol Choi, Aleksandar Bojchevski, and Sujay Sanghavi.
\newblock {SVFT}: Parameter-efficient fine-tuning with singular vectors.
\newblock In \emph{Advances in Neural Information Processing Systems (NeurIPS)}, 2024.
\newblock arXiv:2405.19597.

\bibitem[Liu et~al.(2022)Liu, Tam, Muqeeth, Mohta, Huang, Bansal, and Raffel]{liu2022ia3}
Haokun Liu, Derek Tam, Mohammed Muqeeth, Jay Mohta, Tenghao Huang, Mohit Bansal, and Colin Raffel.
\newblock Few-shot parameter-efficient fine-tuning is better and cheaper than in-context learning.
\newblock In \emph{Advances in Neural Information Processing Systems 35 (NeurIPS 2022)}, 2022.
\newblock Introduces $(IA)^3$. arXiv:2205.05638.

\bibitem[Liu et~al.(2024{\natexlab{a}})Liu, Wang, Yin, Molchanov, Wang, Cheng, and Chen]{liu2024dora}
Shih-Yang Liu, Chien-Yi Wang, Hongxu Yin, Pavlo Molchanov, Yu-Chiang~Frank Wang, Kwang-Ting Cheng, and Min-Hung Chen.
\newblock {DoRA}: Weight-decomposed low-rank adaptation.
\newblock In \emph{International Conference on Machine Learning (ICML)}, 2024{\natexlab{a}}.
\newblock Oral. arXiv:2402.09353.

\bibitem[Liu et~al.(2018)Liu, Lin, Liu, Liu, Yu, Dai, and Song]{liu2018mhe}
Weiyang Liu, Rongmei Lin, Zhen Liu, Lixin Liu, Zhiding Yu, Bo~Dai, and Le~Song.
\newblock Learning towards minimum hyperspherical energy.
\newblock In \emph{Advances in Neural Information Processing Systems (NeurIPS)}, 2018.

\bibitem[Liu et~al.(2024{\natexlab{b}})Liu, Qiu, Feng, Xiu, Xue, Yu, Feng, Liu, Heo, Peng, Wen, Black, Weller, and Sch{\"o}lkopf]{liu2024parameter}
Weiyang Liu, Zeju Qiu, Yao Feng, Yuliang Xiu, Yuxuan Xue, Longhui Yu, Haiwen Feng, Zhen Liu, Juyeon Heo, Songyou Peng, Yandong Wen, Michael~J. Black, Adrian Weller, and Bernhard Sch{\"o}lkopf.
\newblock Parameter-efficient orthogonal finetuning via butterfly factorization.
\newblock In \emph{International Conference on Learning Representations (ICLR)}, 2024{\natexlab{b}}.
\newblock arXiv:2311.06243.

\bibitem[Luo et~al.(2026)Luo, Zhou, Zhang, Wan, Wei, and Zhang]{luo2026keeplora}
Mao-Lin Luo, Zi-Hao Zhou, Yi-Lin Zhang, Yuanyu Wan, Tong Wei, and Min-Ling Zhang.
\newblock {KeepLoRA}: Continual learning with residual gradient adaptation.
\newblock In \emph{International Conference on Learning Representations (ICLR)}, 2026.
\newblock arXiv:2601.19659.

\bibitem[Meng et~al.(2024)Meng, Wang, and Zhang]{meng2024pissa}
Fanxu Meng, Zhaohui Wang, and Muhan Zhang.
\newblock {PiSSA}: Principal singular values and singular vectors adaptation of large language models.
\newblock In \emph{Advances in Neural Information Processing Systems 37 (NeurIPS 2024)}, 2024.
\newblock Spotlight. arXiv:2404.02948.

\bibitem[Mou et~al.(2024)Mou, Wang, Xie, Wu, Zhang, Qi, and Shan]{mou2024t2iadapter}
Chong Mou, Xintao Wang, Liangbin Xie, Yanze Wu, Jian Zhang, Zhongang Qi, and Ying Shan.
\newblock {T2I-Adapter}: Learning adapters to dig out more controllable ability for text-to-image diffusion models.
\newblock In \emph{Proceedings of the AAAI Conference on Artificial Intelligence}, volume~38, pp.\  4296--4304, 2024.
\newblock \doi{10.1609/aaai.v38i5.28226}.
\newblock arXiv:2302.08453.

\bibitem[Nguyen et~al.(2026)Nguyen, Choi, and Tong]{nguyen2026loco}
An~Nguyen, Jaesik Choi, and Anh Tong.
\newblock {LoCO}: Low-rank compositional rotation fine-tuning.
\newblock In \emph{Proceedings of the International Joint Conference on Artificial Intelligence (IJCAI)}, 2026.
\newblock arXiv:2605.15916.

\bibitem[Parlett \& Scott(1979)Parlett and Scott]{parlett1979lanczos}
Beresford~N. Parlett and David~S. Scott.
\newblock The {Lanczos} algorithm with selective orthogonalization.
\newblock \emph{Mathematics of Computation}, 33\penalty0 (145):\penalty0 217--238, 1979.
\newblock \doi{10.1090/S0025-5718-1979-0514820-3}.

\bibitem[Pfeiffer et~al.(2021)Pfeiffer, Kamath, R{\"u}ckl{\'e}, Cho, and Gurevych]{pfeiffer2021adapterfusion}
Jonas Pfeiffer, Aishwarya Kamath, Andreas R{\"u}ckl{\'e}, Kyunghyun Cho, and Iryna Gurevych.
\newblock {AdapterFusion}: Non-destructive task composition for transfer learning.
\newblock In \emph{Proceedings of the 16th Conference of the European Chapter of the Association for Computational Linguistics: Main Volume}, pp.\  487--503, 2021.

\bibitem[Qiu et~al.(2023)Qiu, Liu, Feng, Xue, Feng, Liu, Zhang, Weller, and Sch{\"o}lkopf]{qiu2023oft}
Zeju Qiu, Weiyang Liu, Haiwen Feng, Yuxuan Xue, Yao Feng, Zhen Liu, Dan Zhang, Adrian Weller, and Bernhard Sch{\"o}lkopf.
\newblock Controlling text-to-image diffusion by orthogonal finetuning.
\newblock In \emph{Advances in Neural Information Processing Systems 36 (NeurIPS 2023)}, 2023.
\newblock arXiv:2306.07280.

\bibitem[Qiu et~al.(2025{\natexlab{a}})Qiu, Buchholz, Xiao, Dax, Sch\"{o}lkopf, and Liu]{qiu2025poet}
Zeju Qiu, Simon Buchholz, Tim~Z. Xiao, Maximilian Dax, Bernhard Sch\"{o}lkopf, and Weiyang Liu.
\newblock Reparameterized {LLM} training via orthogonal equivalence transformation.
\newblock In \emph{Advances in Neural Information Processing Systems (NeurIPS)}, 2025{\natexlab{a}}.
\newblock arXiv:2506.08001.

\bibitem[Qiu et~al.(2025{\natexlab{b}})Qiu, Liu, Weller, and Sch{\"o}lkopf]{qiu2025orthogonal}
Zeju Qiu, Weiyang Liu, Adrian Weller, and Bernhard Sch{\"o}lkopf.
\newblock Orthogonal finetuning made scalable.
\newblock In \emph{Proceedings of the 2025 Conference on Empirical Methods in Natural Language Processing}, pp.\  31946--31963, Suzhou, China, November 2025{\natexlab{b}}. Association for Computational Linguistics.
\newblock \doi{10.18653/v1/2025.emnlp-main.1627}.
\newblock arXiv:2506.19847.

\bibitem[Rombach et~al.(2022)Rombach, Blattmann, Lorenz, Esser, and Ommer]{rombach2022ldm}
Robin Rombach, Andreas Blattmann, Dominik Lorenz, Patrick Esser, and Bj{\"o}rn Ommer.
\newblock High-resolution image synthesis with latent diffusion models.
\newblock In \emph{Proceedings of the IEEE/CVF Conference on Computer Vision and Pattern Recognition (CVPR)}, 2022.
\newblock arXiv:2112.10752.

\bibitem[Ruiz et~al.(2023)Ruiz, Li, Jampani, Pritch, Rubinstein, and Aberman]{ruiz2023dreambooth}
Nataniel Ruiz, Yuanzhen Li, Varun Jampani, Yael Pritch, Michael Rubinstein, and Kfir Aberman.
\newblock {DreamBooth}: Fine tuning text-to-image diffusion models for subject-driven generation.
\newblock In \emph{Proceedings of the IEEE/CVF Conference on Computer Vision and Pattern Recognition (CVPR)}, pp.\  22500--22510, 2023.
\newblock arXiv:2208.12242.

\bibitem[Simon(1984)]{simon1984lanczos}
Horst~D. Simon.
\newblock The {Lanczos} algorithm with partial reorthogonalization.
\newblock \emph{Mathematics of Computation}, 42\penalty0 (165):\penalty0 115--142, 1984.
\newblock \doi{10.1090/S0025-5718-1984-0725988-X}.

\bibitem[Wang et~al.(2025)Wang, Li, Wang, Chen, and Chen]{wang2025milora}
Hanqing Wang, Yixia Li, Shuo Wang, Guanhua Chen, and Yun Chen.
\newblock {MiLoRA}: Harnessing minor singular components for parameter-efficient {LLM} finetuning.
\newblock In \emph{Proceedings of the 2025 Conference of the Nations of the Americas Chapter of the Association for Computational Linguistics: Human Language Technologies (Volume 1: Long Papers)}, pp.\  4823--4836, Albuquerque, New Mexico, 2025. Association for Computational Linguistics.
\newblock arXiv:2406.09044.

\bibitem[Wang et~al.(2024)Wang, Yu, and Li]{wang2024lorga}
Shaowen Wang, Linxi Yu, and Jian Li.
\newblock {LoRA-GA}: Low-rank adaptation with gradient approximation.
\newblock In \emph{Advances in Neural Information Processing Systems 37 (NeurIPS 2024)}, 2024.
\newblock arXiv:2407.05000.

\bibitem[Wu et~al.(2026)Wu, Hu, Min, and Wang]{wu2025efficient}
Fei Wu, Jia Hu, Geyong Min, and Shiqiang Wang.
\newblock Efficient orthogonal fine-tuning with principal subspace adaptation.
\newblock In \emph{International Conference on Learning Representations (ICLR)}, 2026.
\newblock arXiv:2505.11235.

\bibitem[Wu et~al.(2024{\natexlab{a}})Wu, Liu, Wang, Li, Lv, Ling, JianHao, Zhang, Zheng, and Huang]{wu2024red}
Muling Wu, Wenhao Liu, Xiaohua Wang, Tianlong Li, Changze Lv, Zixuan Ling, Zhu JianHao, Cenyuan Zhang, Xiaoqing Zheng, and Xuanjing Huang.
\newblock Advancing parameter efficiency in fine-tuning via representation editing.
\newblock In \emph{Proceedings of the 62nd Annual Meeting of the Association for Computational Linguistics (Volume 1: Long Papers)}, pp.\  13445--13464, 2024{\natexlab{a}}.
\newblock arXiv:2402.15179.

\bibitem[Wu et~al.(2024{\natexlab{b}})Wu, Arora, Wang, Geiger, Jurafsky, Manning, and Potts]{wu2024reft}
Zhengxuan Wu, Aryaman Arora, Zheng Wang, Atticus Geiger, Dan Jurafsky, Christopher~D. Manning, and Christopher Potts.
\newblock {ReFT}: Representation finetuning for language models.
\newblock In \emph{Advances in Neural Information Processing Systems 37 (NeurIPS 2024)}, 2024{\natexlab{b}}.
\newblock arXiv:2404.03592.

\bibitem[Yuan et~al.(2024)Yuan, Liu, and Xu]{yuan2024hra}
Shen Yuan, Haotian Liu, and Hongteng Xu.
\newblock Bridging the gap between low-rank and orthogonal adaptation via householder reflection adaptation.
\newblock In \emph{Advances in Neural Information Processing Systems 37 (NeurIPS 2024)}, 2024.
\newblock arXiv:2405.17484.

\bibitem[Zadorozhnyy et~al.(2024)Zadorozhnyy, Mucllari, Pospisil, Nguyen, and Ye]{mucllari2024ncgru}
Vasily Zadorozhnyy, Edison Mucllari, Cole Pospisil, Duc Nguyen, and Qiang Ye.
\newblock Orthogonal gated recurrent unit with {N}eumann--{C}ayley transformation.
\newblock \emph{Neural Computation}, 36\penalty0 (12), 2024.
\newblock arXiv:2208.06496.

\bibitem[Zhang et~al.(2023{\natexlab{a}})Zhang, Rao, and Agrawala]{zhang2023controlnet}
Lvmin Zhang, Anyi Rao, and Maneesh Agrawala.
\newblock Adding conditional control to text-to-image diffusion models.
\newblock In \emph{Proceedings of the IEEE/CVF International Conference on Computer Vision (ICCV)}, pp.\  3836--3847, 2023{\natexlab{a}}.
\newblock arXiv:2302.05543.

\bibitem[Zhang et~al.(2023{\natexlab{b}})Zhang, Chen, Bukharin, He, Cheng, Chen, and Zhao]{zhang2023adalora}
Qingru Zhang, Minshuo Chen, Alexander Bukharin, Pengcheng He, Yu~Cheng, Weizhu Chen, and Tuo Zhao.
\newblock {AdaLoRA}: Adaptive budget allocation for parameter-efficient fine-tuning.
\newblock In \emph{International Conference on Learning Representations (ICLR)}, 2023{\natexlab{b}}.
\newblock arXiv:2303.10512.

\bibitem[Zhao et~al.(2026)Zhao, Mishra, Jawanpuria, Lin, Shi, Gao, and Han]{zhao2026loft}
Lanxin Zhao, Bamdev Mishra, Pratik Jawanpuria, Lequan Lin, Dai Shi, Junbin Gao, and Andi Han.
\newblock {LOFT}: Low-rank orthogonal fine-tuning via task-aware support selection.
\newblock In \emph{Advances in Neural Information Processing Systems 39 (NeurIPS 2026)}, 2026.

\end{thebibliography}
